%% file: main.tex
\documentclass[10pt,journal,compsoc]{IEEEtran}
\usepackage{amsmath}
\usepackage{amssymb}% http://ctan.org/pkg/amssymb
\usepackage{pifont}% http://ctan.org/pkg/pifont

\usepackage{subfig}
\usepackage{comment}
\usepackage{soul}

\usepackage{algorithmic}

\usepackage{hyperref}
\hypersetup{
	colorlinks=true,
	linkcolor=red,
	filecolor=magenta,      
	urlcolor=black,
	citecolor=black %
}
\usepackage{booktabs} % For formal tables
\usepackage{mathrsfs}
\usepackage{amsfonts}
\usepackage[ruled]{algorithm2e} % For algorithms

\usepackage{multirow}
\usepackage{graphicx}
\usepackage{color}
\usepackage{balance}

\usepackage{adjustbox}
\usepackage{soul}

\setstcolor{red}

\newtheorem{definition}{Definition}
\usepackage{makecell}
\newcommand{\hlt}{}

\ifCLASSOPTIONcompsoc

  \usepackage[nocompress]{cite}
\else
  \usepackage{cite}
\fi

\ifCLASSINFOpdf

\else

\fi

\begin{document}

\title{Graph Self-Supervised Learning: A Survey}

\author{Yixin Liu, 
Ming Jin, Shirui Pan,
Chuan Zhou, Yu Zheng, 
Feng Xia, %
Philip S. Yu,~\IEEEmembership{Life Fellow,~IEEE}%

\IEEEcompsocitemizethanks{
	\IEEEcompsocthanksitem Y. Liu,  M. Jin, and S. Pan are with Department of Data Science \& AI, Faculty of IT, Monash University, VIC 3800, Australia (E-mail: yixin.liu@monash.edu;  ming.jin@monash.edu; shirui.pan@monash.edu;).
	\IEEEcompsocthanksitem C. Zhou is with Academy of Mathematics and Systems Science, Chinese Academy of Sciences, China (Email: zhouchuan@amss.ac.cn). 
	\IEEEcompsocthanksitem Y. Zheng is with the Department of Computer Science and Information Technology, La Trobe University, Melbourne, Australia (E-mail: yu.zheng@latrobe.edu.au).
	\IEEEcompsocthanksitem F. Xia is with School of Engineering, Information Technology and Physical Sciences, Federation University, Australia (Email: f.xia@federation.edu.au). 
	\IEEEcompsocthanksitem P. S. Yu is with Department of Computer Science, University of Illinois at Chicago, Chicago, IL 60607-7053, USA (Email: psyu@uic.edu)
	\IEEEcompsocthanksitem Corresponding author: Shirui Pan.
    }

\thanks{%Manuscript received August 5, 2021; revised xx x, 202x. 
Y. Liu and M. Jin contributed equally to this work. This work was supported in part by an ARC Future Fellowship  (FT210100097), NSF under grants III-1763325, III-1909323,  III-2106758, and SaTC-1930941.
}

}

\markboth{Journal of \LaTeX\ Class Files,~Vol.~14, No.~8, August~2021}%
{Shell \MakeLowercase{\textit{et al.}}: Bare Demo of IEEEtran.cls for Computer Society Journals}

\IEEEtitleabstractindextext{%
\begin{abstract}
Deep learning on graphs has attracted significant interests recently. However, most of the works have focused on (semi-) supervised learning, resulting in shortcomings including heavy label reliance, poor generalization, and weak robustness. To address these issues, self-supervised learning (SSL), which extracts informative knowledge through well-designed pretext tasks without relying on manual labels, has become a promising and trending learning paradigm for graph data. Different from SSL on other domains like computer vision and natural language processing, SSL on graphs has an exclusive background, design ideas, and taxonomies. 
Under the umbrella of \textit{graph self-supervised learning},  
we present a timely and comprehensive review of the existing approaches which employ SSL techniques for graph data. 
We construct a unified framework that mathematically formalizes the paradigm of graph SSL.
According to the objectives of pretext tasks, we divide these approaches into four categories: generation-based, auxiliary property-based, contrast-based, and hybrid approaches.
We further {describe} the applications of graph SSL across various research fields and summarize the commonly used datasets, evaluation benchmark, performance comparison and open-source codes of graph SSL.
Finally, we discuss the remaining challenges and potential future directions in this research field. \end{abstract}

}

\maketitle

\IEEEdisplaynontitleabstractindextext

\IEEEpeerreviewmaketitle

\input{content.tex}

\appendices
\input{supplemental.tex}

\bibliographystyle{IEEEtran}
\bibliography{IEEEabrv,biblio_simp}

\input{bio}

\end{document}

%% file: content.tex
\newcommand{\methodHL}[1]{#1}
\newcommand{\EOL}{\hline}
\newcommand{\tabincell}[2]{\begin{tabular}{@{}#1@{}}#2\end{tabular}}

\section{Introduction}\label{sec:introduction}

\IEEEPARstart{I}{n} recent years, deep learning on graphs \cite{gcn_kipf2017semi,gat_ve2018graph,xu2018how,wu2020comprehensive} has become increasingly popular for the artificial intelligence research community since graph-structured data is ubiquitous in numerous domains, including e-commerce \cite{li2020hierarchical}, traffic \cite{wu2019graphwavenet}, chemistry \cite{liu2018constrained}, and knowledge base \cite{ji2021survey}. Most deep learning studies on graphs  focus on (semi-) supervised learning scenarios, where specific downstream tasks (\textit{e.g.,} node classification%
) are exploited to train models with well-annotated manual labels. Despite the success of these studies, the heavy reliance on labels brings several shortcomings. Firstly, the cost of the collection and annotation of manual labels is prohibitive, especially for the research areas which have large-scale datasets (\textit{e.g.}, citation and social networks \cite{hu2020gpt}) or demand on domain knowledge (\textit{e.g.}, chemistry and medicine \cite{rong2020self}). Secondly, a purely supervised learning scenario usually suffers from poor generalization owing to the over-fitting problem, particularly when training data is scarce \cite{Rong2020DropEdge}. Thirdly, supervised graph deep learning models are vulnerable to label-related adversarial attacks, causing the weak robustness of graph supervised learning \cite{zhang2020adversarial}. 

To address the shortcomings of (semi-) supervised learning, self-supervised learning (SSL) provides a promising learning paradigm that reduces the dependence on manual labels. 
In SSL,  models are learned by solving a series of handcrafted auxiliary tasks (so-called pretext tasks), in which the supervision signals are acquired from data itself automatically without the need for manual annotation. With the help of well-designed pretext tasks, SSL enables the model to learn more informative representations from unlabeled data to achieve better performance \cite{velivckovic2018deep,hassani2020contrastive}, generalization \cite{hu2020gpt,qiu2020gcc,hu2019strategies} and robustness \cite{you2020does,jovanovic2021towards} on various downstream tasks. 

Described as ``the key to human-level intelligence'' by Turing Award winners Yoshua Bengio and Yann LeCun, SSL has recently achieved great success in the domains of computer vision (CV) and natural language processing (NLP). 
Early SSL methods in CV domain design various semantics-related pretext tasks for visual representation learning \cite{jing2020self}, such as image inpainting \cite{pathak2016context_inpainting}, image colorizing \cite{zhang2016colorful}, and jigsaw puzzle \cite{noroozi2016unsupervised_jigsaw}, etc. 
Lately, self-supervised contrastive learning frameworks (\textit{e.g.}, MoCo \cite{he2020momentum}, SimCLR \cite{chen2020simple} and BYOL \cite{grill2020bootstrap}) leverage the invariance of semantics under image transformation to learn visual features. In the NLP domain, early word embedding methods \cite{mikolov2013befficient,mikolov2013distributed} share the same idea with SSL which learns from data itself. Pre-trained by linguistic pretext tasks, recent large-scale language models (\textit{e.g.}, BERT \cite{devlin2018bert} and XLNet \cite{yang2019xlnet}) achieve state-of-the-art performance on multiple NLP tasks.

\begin{figure}[tp]
\centering
\subfloat[SSL pretext tasks in CV: image colorizing (upper) and image contrastive learning (bottom).]{\includegraphics[height=1.22in]{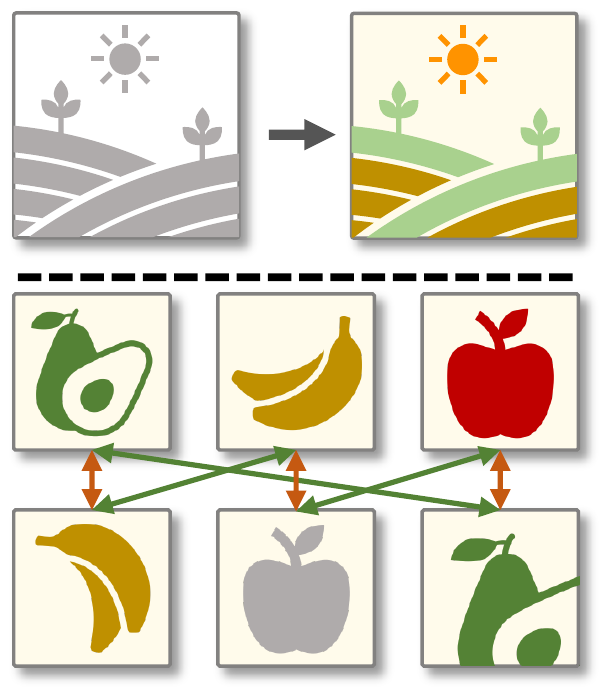}%
\label{subfig:toy_cv}}
\hfill
\subfloat[SSL pretext tasks in NLP: masked language modeling (upper) and next sentence prediction (bottom).]{\includegraphics[height=1.22in]{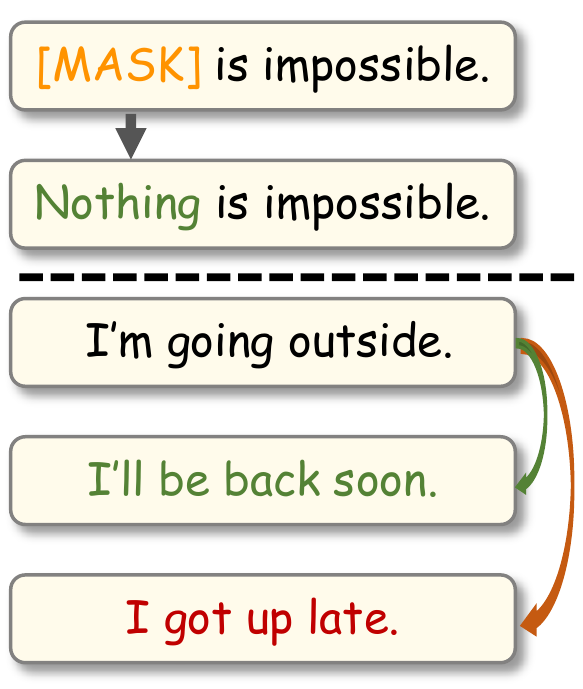}%
\label{subfig:toy_nlp}}
\hfill
\subfloat[SSL pretext tasks in graph {analytics}: masked graph generation (upper) and node contrastive learning (bottom).]{\includegraphics[height=1.22in]{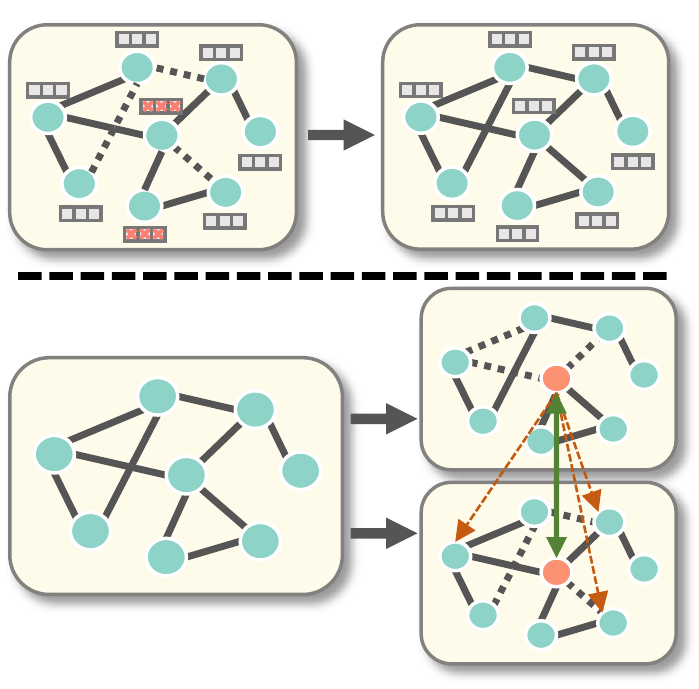}%
\label{subfig:toy_graph}}
\caption{Toy examples of different SSL pretext tasks in CV, NLP and graph {analytics}. In generative tasks, graph SSL should consider the topological structure in an irregular grid as well as node features, while SSL in CV/NLP just needs to recover the information in 2D/1D grid space. In contrastive tasks, the dependency between nodes is non-negligible in graph SSL, while the samples in CV/NLP are independent.}
\label{fig:toy_example}
\end{figure}

Following the immense success of SSL on CV and NLP, very recently, there has been increasing interest in applying SSL to graph-structured data. However, it is non-trivial to transfer the pretext tasks designed for CV/NLP for graph {data analytics}.
The main challenge is that graphs are in irregular non-Euclidean data space. 
Compared to the 2D/1D regular-grid Euclidean spaces where image/language data reside in, non-Euclidean spaces are more general but more complex. Therefore, some pretext tasks for grid-structure data cannot be mapped to graph data directly.
Furthermore, the data examples (nodes) in graph data are correlated {with} the topological structure naturally, while the examples in CV (image) and NLP (text) are often independent. Hence, how to deal with such dependency in graph SSL becomes a challenge for pretext task designs. Fig. \ref{fig:toy_example} illustrates such differences with some toy examples.
Considering the significant difference between SSL in graph {analytics} and other {research areas}, exclusive definitions and taxonomies are required for graph SSL.

The history of graph SSL goes back to at least the early studies on unsupervised graph embedding \cite{perozzi2014deepwalk,grover2016node2vec} \footnote{{A timeline of milestone works are summarized in Appendix \ref{appendix:timeline}.}}. These methods learn node representations by maximizing the agreement between contextual nodes within truncated random walks. A classical unsupervised learning model, graph autoencoder (GAE) \cite{kipf2016variational}, can also be regarded as a graph SSL method that learns to rebuild the graph structure. Since 2019, the recent wave of graph SSL has brought about various designs of pretext tasks, from contrastive learning \cite{velivckovic2018deep,zhu2020deep} to graph property mining \cite{rong2020self,you2020does}. Considering the increasing trend of graph SSL research and the diversity of related pretext tasks, there is an urgent need to construct a unified framework and systematic taxonomy to summarize the methodologies and applications of graph SSL.

To fill the gap, this paper conducts a comprehensive and up-to-date overview of the rapidly growing area of graph SSL, and also provides abundant resources and discussions of related applications.
The intended audiences for this article are general machine learning researchers who would like to know about self-supervised learning on graph data, graph learning researchers who want to keep track of the most recent advances on graph neural networks (GNNs), and domain experts who would like to generalize graph SSL approaches to new applications or other fields. The core contributions of this survey are summarized as follows:

\begin{itemize}
	\item \textbf{Unified framework and systematic taxonomy.} We propose a unified framework that mathematically formalizes graph SSL approaches. Based on our framework, we systematically categorize the existing works into four groups: generation-based, auxiliary property-based, contrast-based, and hybrid methods. We also build the taxonomies of downstream tasks and SSL learning schemes.
	\item \textbf{Comprehensive and up-to-date review.} We conduct a comprehensive and timely review for classical and latest graph SSL approaches. For each type of graph SSL approach, we provide fine-grained classification, mathematical description, detailed comparison, and high-level summary.
	\item \textbf{Abundant resources and applications.} We collect abundant resources on graph SSL, including datasets, evaluation benchmark, performance comparison, and open-source codes. We also summarize the practical applications of graph SSL in various research fields.
	\item \textbf{Outlook on future directions.} We point out the technical limitations of current research. We further suggest six promising directions for future works from different perspectives.
\end{itemize}

\textit{Comparison with related survey articles.} Some existing surveys mainly review from the perspectives of general SSL \cite{liu2020self}, SSL for CV \cite{jing2020self}, or self-supervised contrastive learning \cite{jaiswal2021survey}, while this paper purely focuses on SSL for {graph-structured data}. Compared to the recent surveys on graph self-supervised learning \cite{xie2021self,wu2021self}, our survey has a more comprehensive overview on this topic and provides the following differences: (1) {a unified encoder-decoder framework to define graph SSL}; (2) a systematical and more fine-grained taxonomy {from a mathematical perspective}; (3) more up-to-date review; (4) more detailed {summary of resources} including {performance comparison,} datasets, {implementations}, and practical applications; and (5) more forward-looking discussion for challenges and future directions. %

The remainder of this article is organized as follows. 
Section \ref{sec:definition} defines the related concepts and provides notations used in the remaining sections. 
Section \ref{sec:categorization} describes the framework of graph SSL and provides categorization from multiple perspectives. 
Section \ref{sec:generation}-\ref{sec:hybrid} review four categories of graph SSL approaches respectively.
Section \ref{sec:resource} summarizes {the useful resources for empirical study of graph SSL, including performance comparison, datasets, and open-source implementations.} %
Section \ref{sec:practical_application} surveys the real-world applications in various domains.
Section \ref{sec:future_direction} analyzes the remaining challenges and possible future directions.
Section \ref{sec:conclusion} concludes this article in the end.

\vspace{-3mm}
\section{Definition and Notation}\label{sec:definition}

In this section, we outline the related term definitions of graph SSL, list commonly used notations, and define graph-related concepts.

\subsection{Term Definitions}

In graph SSL, we provide the following definitions of related essential concepts. 

\noindent \textbf{Manual Labels Versus Pseudo Labels.}
\textit{Manual labels}, a.k.a. human-annotated labels in some papers \cite{jing2020self}, indicate the labels that human experts or workers manually annotate. \textit{Pseudo labels}, in contrast, denote the labels that can be acquired automatically from data by machines without any human knowledge. In general, pseudo labels require lower acquisition costs than manual labels so that they have advantages when manual labels are difficult to obtain or the amount of data is vast. In self-supervised learning settings, specific methods can be designed to generate pseudo labels, enhancing the representation learning. %

\noindent \textbf{Downstream Tasks Versus Pretext Tasks.}
\textit{Downstream tasks} are the graph analytic tasks used to evaluate the quality or performance of the feature representation learned by different models. %
Typical applications include node classification and graph classification. 
\textit{Pretext tasks} refer to the pre-designed tasks for models to solve (\textit{e.g.}, graph reconstruction), 
{which helps models to learn more generalized representations from unlabeled data, and thus benefits downstream tasks by providing a better initialization or more effective regularization.}
In general, solving downstream tasks needs manual labels, while pretext tasks are usually learned with pseudo labels.

\noindent \textbf{Supervised Learning, Unsupervised Learning and Self-Supervised Learning.} 
\textit{Supervised learning} refers to the learning paradigm that leverages well-defined manual labels to train machine learning models. Conversely, \textit{unsupervised learning} refers to the learning paradigm without using any manual labels. As a subset of unsupervised learning, \textit{self-supervised learning} indicates the learning paradigm where supervision signals are generated from data itself. In self-supervised learning methods, models are trained with pretext tasks to obtain better performance and generalization on downstream tasks.

\subsection{Notations}

We provide important notations used in this paper (which are summarized in Appendix \ref{appendix:notations}) and the definitions of different types of graphs and GNNs in this subsection.

\begin{definition}[Plain Graph]
	 A plain graph\footnote{A plain graph is an \textit{unattributed, static, and homogeneous} graph.} is represented as $\mathcal{G}=(\mathcal{V}, \mathcal{E})$, where $\mathcal{V}=\{v_1,\dots,v_n\}$ ($\lvert \mathcal{V} \rvert = n$) is the set of nodes and $\mathcal{E}$ ($\lvert \mathcal{E} \rvert = m$) is the set of edges, and naturally we have $\mathcal{E} \subseteq \mathcal{V} \times \mathcal{V}$. The neighborhood of a node $v_i$ is denoted as $\mathcal{N}(v_i)= \{v_j\in \mathcal{V}|e_{i,j}\in \mathcal{E}\}$. The topology of the graph is represented as an adjacency matrix $\mathbf{A} \in \mathbb{R}^{n \times n}$, where $\mathbf{A}_{i,j} = 1$ means $e_{i,j}\in \mathcal{E}$, and $\mathbf{A}_{i,j} = 0$ means $e_{i,j}\notin \mathcal{E}$.
\end{definition}

\vspace{1mm}

\begin{definition}[Attributed Graph]
	An attributed graph refers to a graph where nodes and/or edges are associated with their own features (a.k.a attributes). The feature matrices of nodes and edges are represented as $\mathbf{X_{node}} \in \mathbb{R}^{n \times d_{node}}$ and $\mathbf{X_{edge}} \in \mathbb{R}^{m \times d_{edge}}$ respectively. In a more common scenario where only nodes have features, we use $\mathbf{X} \in \mathbb{R}^{n \times d}$ to denote the node feature matrix for short, and denote the attributed graph as $\mathcal{G}=(\mathcal{V}, \mathcal{E}, \mathbf{X})$.
\end{definition}

There are also some dynamic graphs and heterogeneous graphs whose definitions are given in Appendix \ref{appendix:graphs}.

\vspace{1mm}

{Most of the reviewed methods leverage GNNs as backbone encoders to transform the input {raw} node features $\mathbf{X}$ into {compact} node representations $\mathbf{H}$ by leveraging the rich underlying node connectivity, i.e., adjacency matrix $\mathbf{A}$, with learnable parameters.}
Furthermore, readout functions $\mathcal{R}(\cdot)$ are often employed to generate a graph{-level} representation $\mathbf{h}_{\mathcal{G}}$ from node-level representations $\mathbf{H}$. The formulation of GNNs and readout functions are introduced in Appendix \ref{appendix:gnns}. {Besides, in Appendix \ref{appendix:loss}, we formulate the commonly used loss functions in this survey.}

\vspace{-3mm}
\section{Framework and Categorization}\label{sec:categorization}

In this section, we provide a unified framework of graph SSL, and further categorize it from different perspectives, including pretext tasks, downstream tasks, and the combination of both (i.e., self-supervised training schemes). 

\subsection{Unified Framework and Mathematical Formulation of Graph Self-Supervised Learning}

\begin{figure}[tbp]
	\centering
	\includegraphics[width=0.5\textwidth]{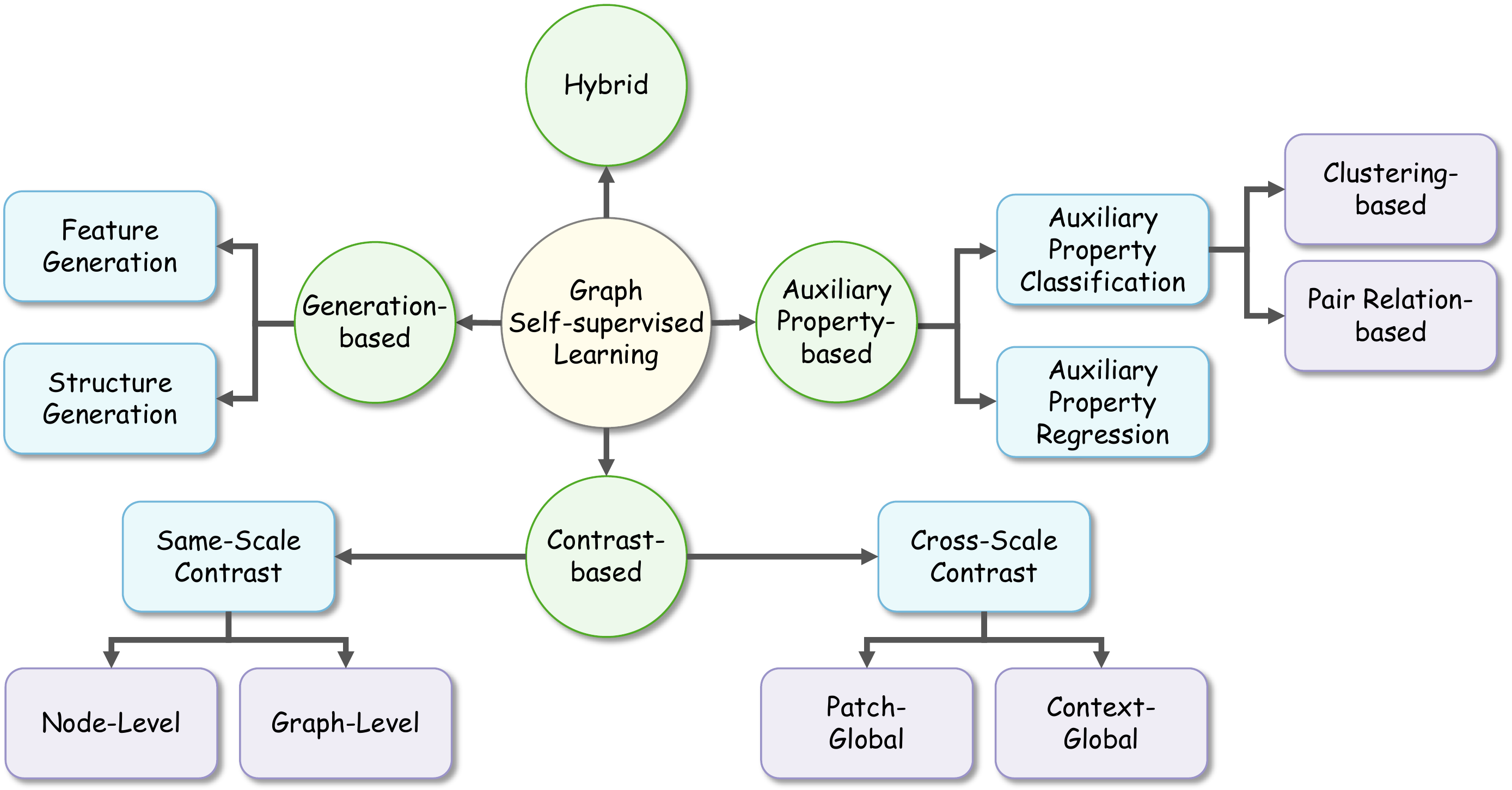}
	\caption{Categorization of graph SSL methods.}
	\label{fig:tree}
\end{figure}

We construct an encoder-decoder framework to formalize graph SSL. The encoder $f_{\theta}$ {(parameterized by $\theta$)} aims to learn a low-dimensional representation (a.k.a. embedding) $\mathbf{h_i} \in \mathbf{H}$ for each node $v_i$ from graph $\mathcal{G}$. In general, the encoder $f_{\theta}$ can be GNNs {\cite{velivckovic2018deep,zhu2020deep,you2020graph}} or other types of neural networks for graph learning {\cite{perozzi2014deepwalk,grover2016node2vec,zhang2020graph}}. The pretext decoder $p_{\phi}$ {(parameterized by $\phi$)} takes $\mathbf{H}$ as {its} input for the pretext tasks. The architecture of $p_{\phi}$ depends on specific downstream tasks. 

Under this framework, graph SSL can be formulated as:
\begin{equation}
{\theta^{*}, \phi^{*}} = \mathop{\arg\min}\limits_{{\theta, \phi}}\mathcal{L}_{ssl}\left( f_{\theta}, p_{\phi}, \mathcal{D} \right),
\label{eq: graph ssl}
\end{equation}
where $\mathcal{D}$ denotes the graph data distribution that satisfies $(\mathcal{V}, \mathcal{E}) \sim \mathcal{D}$ in an unlabeled graph $\mathcal{G}$, and $\mathcal{L}_{ssl}$ is the SSL loss function that regularizes the output of pretext decoder according to specific crafted pretext {tasks}. %

By leveraging the trained graph encoder $f_{\theta^{*}}$, the generated representations can then be used in various downstream tasks. Here we introduce a downstream decoder $q_{\psi}$ {(parameterized by $\psi$)}, %
and formulate the downstream task as a graph supervised learning task:
\begin{equation}
{\theta^{**}, \psi^{*}} = \mathop{\arg\min}\limits_{{{\theta^{*}}, {\psi}}}\mathcal{L}_{sup}\left( f_{\theta^{*}}, q_{\psi}, \mathcal{G}, y \right),
\label{eq: downstrem task}
\end{equation}
where $y$ denotes the downstream task labels, % 
and $\mathcal{L}_{sup}$ is the supervised loss that trains the model for downstream tasks.

\begin{figure}[htbp]
	\centering
    \subfloat[Generation-based graph SSL methods. The model input is generated by a (optional) graph perturbation. In the pretext task, a generative decoder tries to recover the original graph from representation $\mathbf{H}$, with a loss function aiming to minimize the difference between the reconstructed and original graphs.
    ]{\includegraphics[width=2.9in]{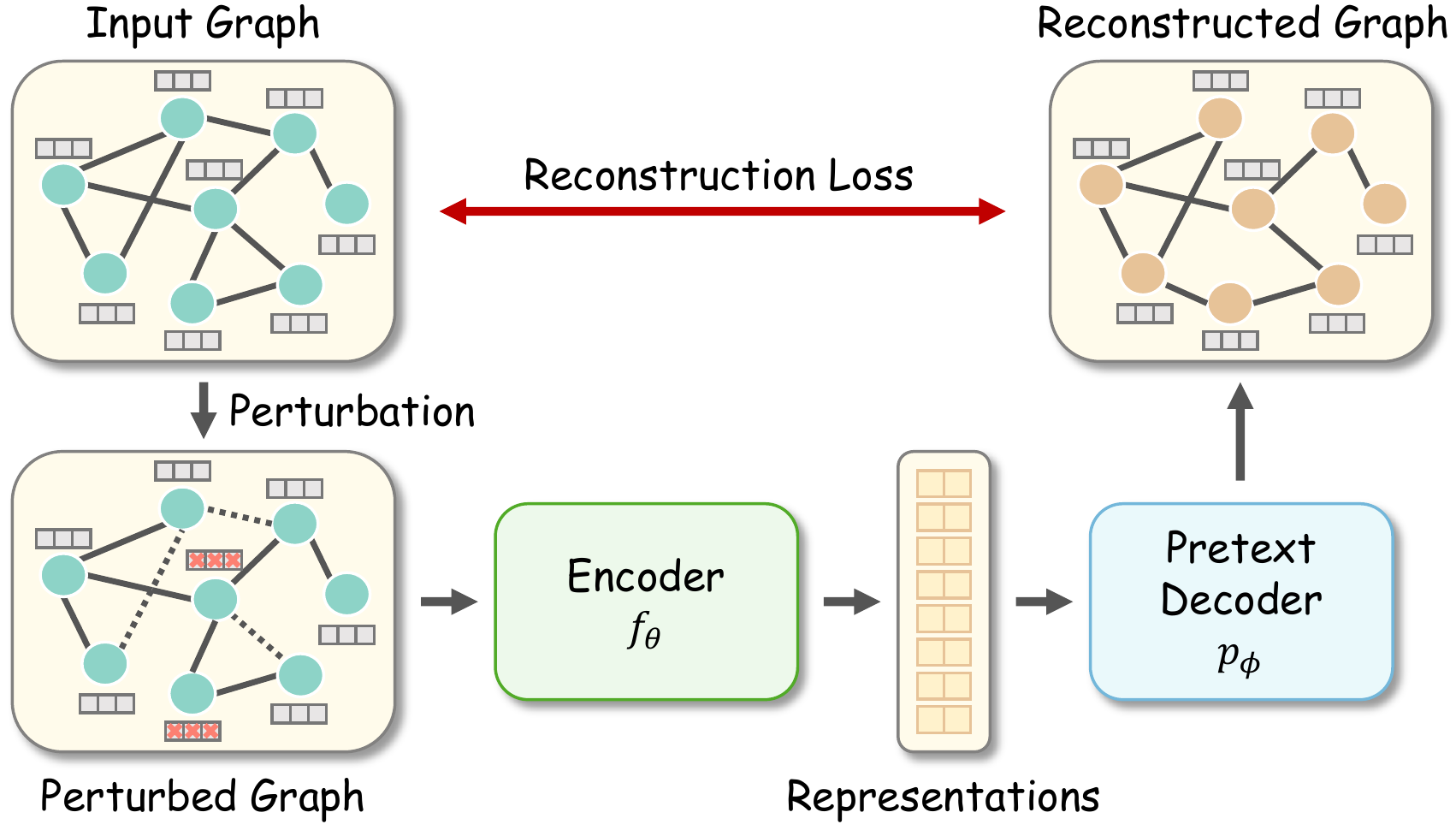}%
    \label{fig:ssl_gen}
    }
    \\
	\subfloat[Auxiliary property-based graph SSL methods. The auxiliary properties are extracted from graphs freely. A classification- or regression- based decoder aims to predict the extracted properties under the training of CE/MSE loss.
	]{\includegraphics[width=2.9in]{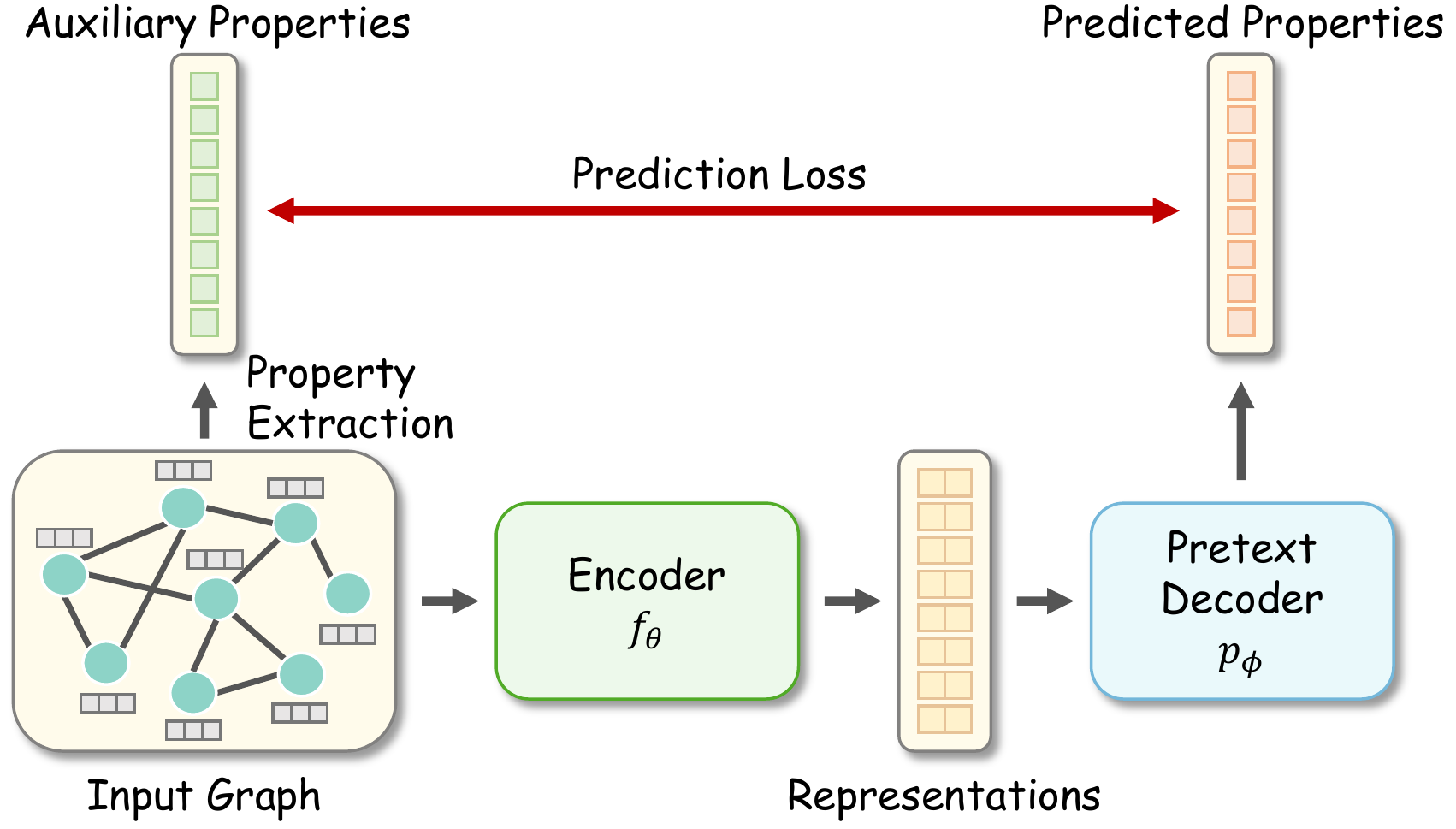}%
	\label{fig:ssl_app}}
    \\
	\subfloat[Contrast-based graph SSL methods. Two different augmented views are constructed from the original graph. Then, the model is trained by maximizing the MI between two views. The MI estimator with (optional) projection often serves as the pretext decoder and SSL loss.
	]{\includegraphics[width=2.9in]{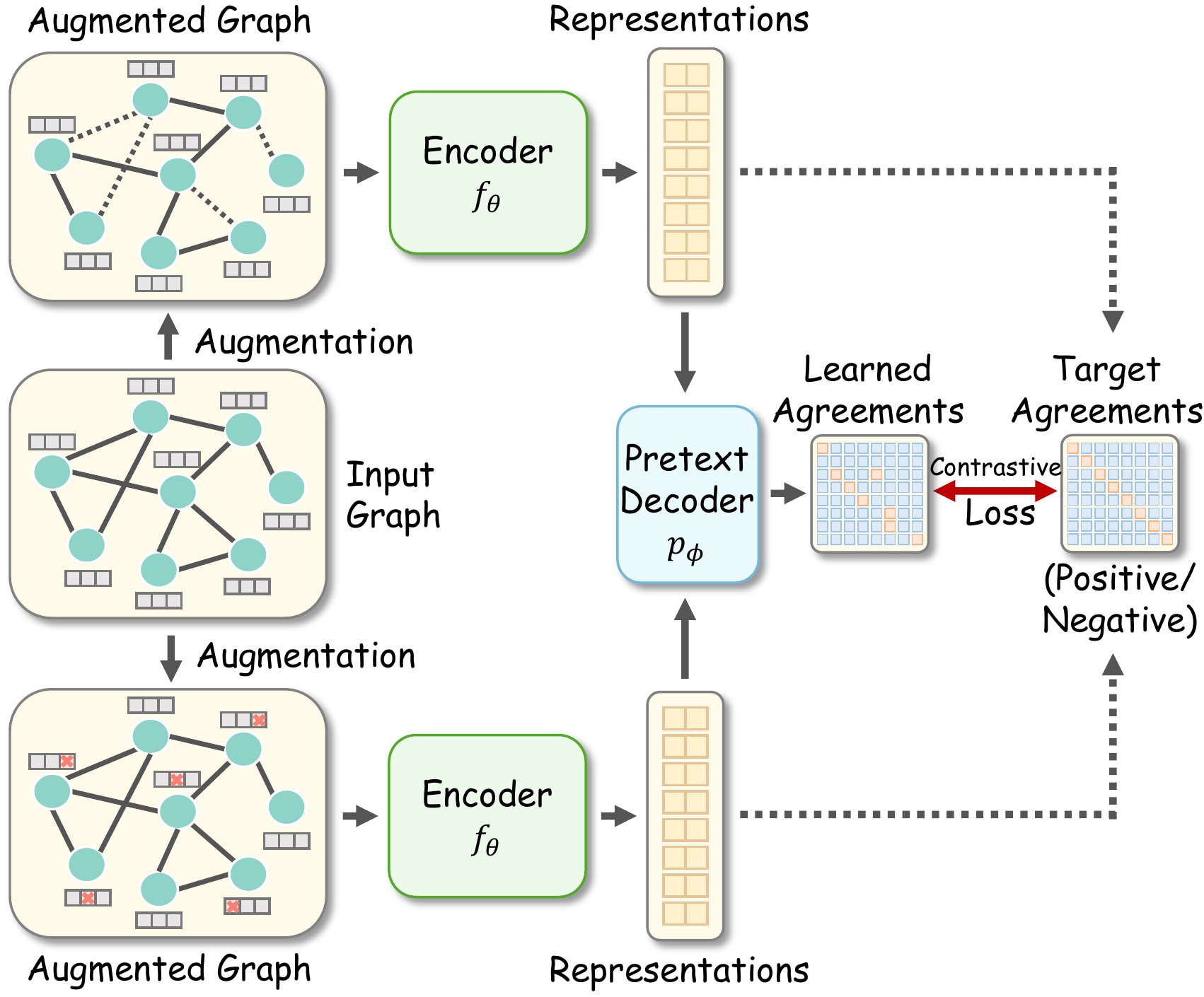}%
	\label{fig:ssl_con}}
    \\
	\subfloat[Hybrid graph SSL methods. Multiple pretext tasks are designed to train the model together in a multi-task learning manner.
	]{\includegraphics[width=2.9in]{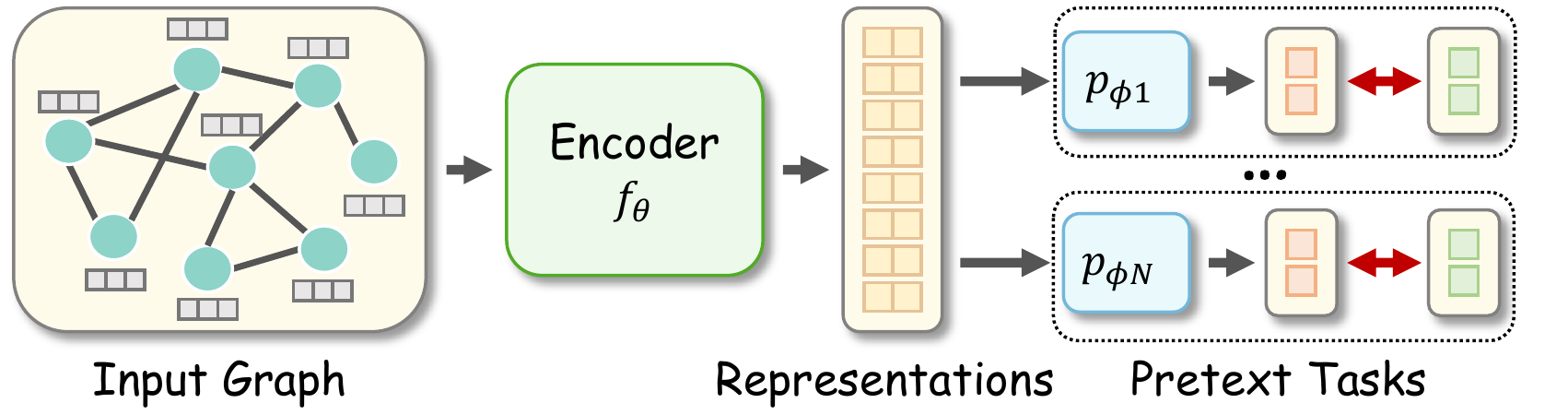}%
	\label{fig:ssl_hyb}}
	\caption{Four categories of graph SSL.}
	\label{fig:ssl_taxonomy}
\end{figure}

In the following subsections, we specify four graph SSL variants based on Equation (\ref{eq: graph ssl}) in Section \ref{subsec: gssl}, three graph self-supervised training schemes in Section \ref{subsec: training schemes} by combining Equations (\ref{eq: graph ssl}) and (\ref{eq: downstrem task}) differently, and three types of downstream tasks based on Equation (\ref{eq: downstrem task}) in Section \ref{subsec: downstream}.

\subsection{Taxonomy of Graph Self-supervised Learning}\label{subsec: gssl}
Graph SSL can be divided into four types conceptually, including generation-based, auxiliary property-based, contrastive-based and hybrid methods, by leveraging different designs of pretext decoders and objective functions. The categorizations of these methods are briefly discussed below and shown in Fig. \ref{fig:tree}, and the concept map of each type of methods is given in Fig. \ref{fig:ssl_taxonomy}.

\noindent \textbf{Generation-based Methods} 
form the pretext task as the graph data reconstruction from two perspectives: feature and structure. Specifically, they focus on the node/edge features or/and graph adjacency reconstructions. In such a case, Equation (\ref{eq: graph ssl}) can be further derived as:
\begin{equation}
{\theta^{*},\phi^{*}} = \mathop{\arg\min}\limits_{{\theta}, {\phi}}\mathcal{L}_{ssl}\Big( p_{\phi}\big( f_{\theta}(\tilde{\mathcal{G}})\big) , \mathcal{G} \Big),
\label{eq: generation}
\end{equation}
where $f_{\theta}(\cdot)$ and {$p_{\phi}(\cdot)$} are graph encoder and pretext decoder. $\tilde{\mathcal{G}}$ denotes the graph data with perturbed node/edge features or/and adjacency matrix. For most of the generation-based approaches, the self-supervised objective function $\mathcal{L}_{ssl}$ is typically defined {to measure} the difference between the reconstructed and the original graph data. {One of the representative approaches is GAE \cite{kipf2016variational} which learns embeddings by rebuilding the graph adjacency matrix.}

\noindent \textbf{Auxiliary Property-based Methods} 
enrich the supervision signals by capitalizing on a larger set of attributive and topological graph properties. In particular, for different crafted auxiliary properties, we further categorize these methods into two types: regression- and classification-based. Formally, they can be formulated as:
\begin{equation}
{\theta^{*},\phi^{*}} = \mathop{\arg\min}\limits_{{\theta}, {\phi}}\mathcal{L}_{ssl}\Big( p_{\phi}\big( f_{\theta}(\mathcal{G})\big),c \Big),
\label{eq: auxiliary}
\end{equation}
where $c$ denotes the specific crafted auxiliary properties. For regression-based approaches, $c$ can be localized or global graph properties, such as the node degree or distance to clusters within $\mathcal{G}$. For classification-based methods, on the other hand, the auxiliary properties are typically constructed as pseudo labels, such as the graph partition or cluster indices. Regarding to the objective function, $\mathcal{L}_{ssl}$ can be mean squared error (MSE) for regression-based and cross-entropy (CE) loss for classification-based methods. {As a pioneering work, M3S \cite{sun2020multi} uses node clustering to construct pseudo labels that provide supervision signals.}

\noindent \textbf{Contrast-based Methods} {are} usually {developed based} on the concept of mutual information (MI) maximization, where the estimated MI between augmented instances of the same object (\textit{e.g.}, node, subgraph, and graph) is maximized. For contrastive-based graph SSL, Equation (\ref{eq: graph ssl}) is reformulated as:
\begin{equation}
{\theta^{*},\phi^{*}} = \mathop{\arg\min}\limits_{{\theta}, {\phi}}\mathcal{L}_{ssl}\Bigg( p_{\phi}\Big(f_{\theta}(\tilde{\mathcal{G}}^{(1)}), f_{\theta}(\tilde{\mathcal{G}}^{(2)})\Big)\Bigg),
\label{eq: contrastive}
\end{equation}
{where $\tilde{\mathcal{G}}^{(1)}$ and $\tilde{\mathcal{G}}^{(2)}$ are two differently augmented instances of $\mathcal{G}$. In these methods, the pretext decoder $p_{\phi}$ indicates the discriminator that estimates the agreement between two instances (\textit{e.g.}, the bilinear function or the dot product), and $\mathcal{L}_{ssl}$ denotes the contrastive loss. By combining them and optimizing $\mathcal{L}_{ssl}$, the pretext tasks aim to estimate and maximize the MI between positive pairs (e.g., augmented instances of the same object) and minimize the MI between negative samples (e.g., instances derived from different objects), which is implicitly included in $\mathcal{L}_{ssl}$. Representative works include cross-scale methods (e.g., DGI \cite{velivckovic2018deep}) and same-scale methods (e.g., GraphCL \cite{you2020graph} and GCC \cite{qiu2020gcc}). }

\begin{figure}[tp]
	\centering
    \subfloat[Pre-training and Fine-tuning (PF). First, the encoder is pre-trained with pretext tasks in an unsupervised manner. Then, the pre-trained parameters are leveraged as the initial parameters in the fine-tuning phase where encoder is trained by downstream tasks independently.
    ]{\includegraphics[width=3.1in]{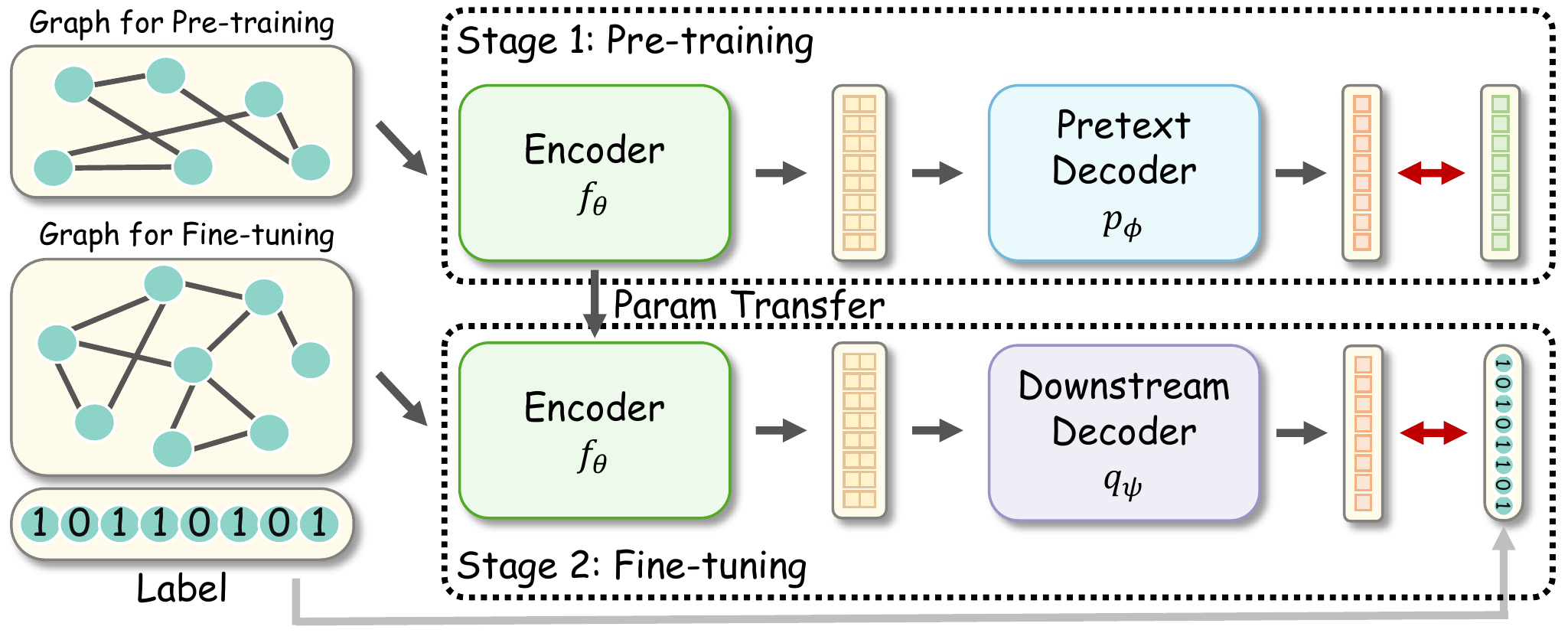}
    \label{fig:scheme_pf}
    }
    \\
	\subfloat[Joint Learning (JL). The model is trained with pretext and downstream tasks in a multi-task learning manner.
	]{\includegraphics[width=3.1in]{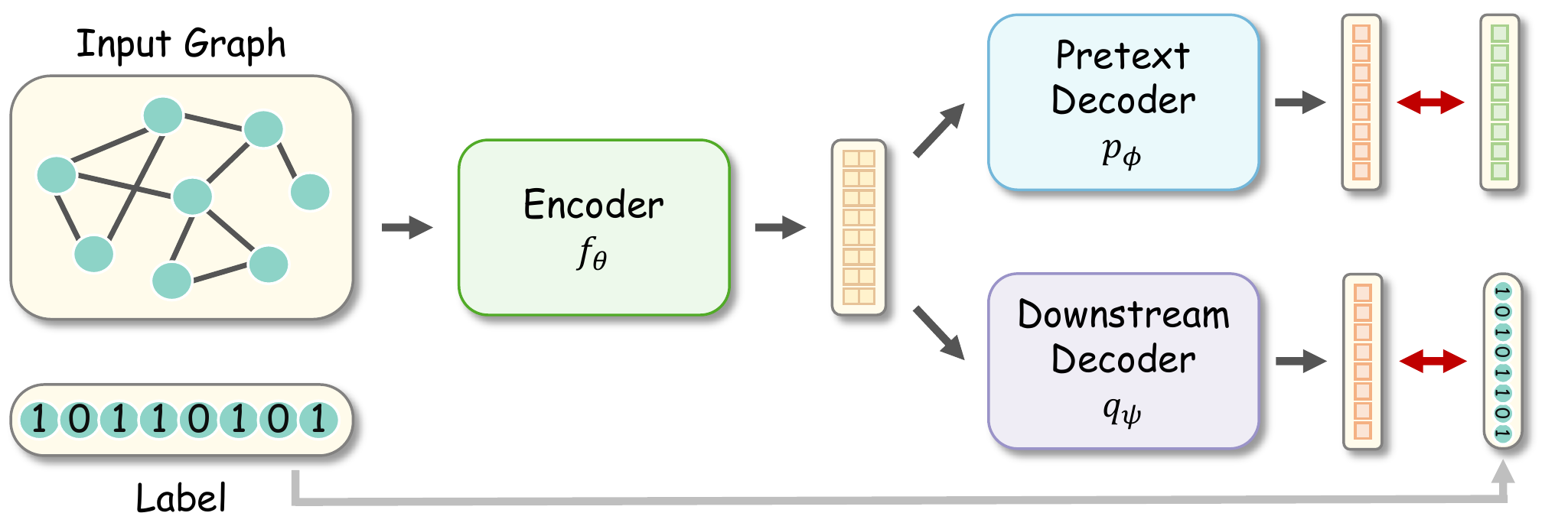}
	\label{fig:scheme_jl}}
	\\
	\subfloat[Unsupervised Representation Learning (URL). It first trains encoder with pretext tasks, and uses the fixed representations to learn the downstream decoder in Stage 2.
	]{\includegraphics[width=3.1in]{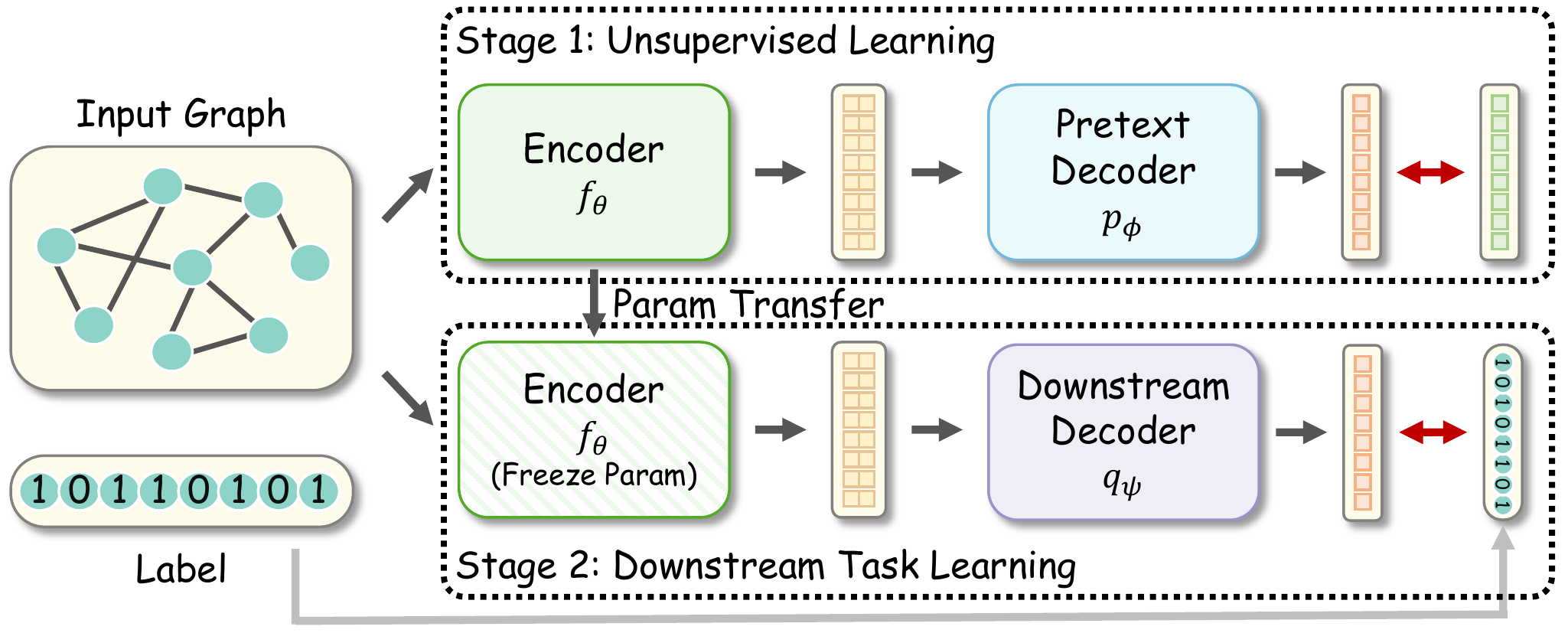}
	\label{fig:scheme_url}}
	\caption{Three types of learning schemes for SSL.} 
	\label{fig:learning_scheme}
	\vspace{-3mm}
\end{figure}

\noindent \textbf{Hybrid Methods}  
take advantage of previous categories and consist of more than one pretext decoder and/or training objective. We formulate this branch of methods as the weighted or unweighted combination of two or more graph SSL schemes based on formulas from Equation (\ref{eq: generation}) to (\ref{eq: contrastive}). {GMI \cite{peng2020graph}, which jointly considers edge-level reconstruction and node-level contrast, is a typical hybrid method.}

\noindent {\textbf{Discussion.} {Different} graph SSL methods have {different} properties. Generation-based methods are simple to implement since the reconstruction task is easy to build, but sometimes recovering input data is memory-consuming for large-scale graphs. Auxiliary property-based methods enjoy the uncomplicated design of decoders and loss functions; however, the selection of helpful auxiliary properties often needs domain knowledge. Compared to other categories, contrast-based methods have more flexible designs and boarder applications. Nevertheless, the designs of contrastive frameworks, augmentation strategies, and loss functions usually {rely} on time-consuming empirical experiments. Hybrid methods benefit from multiple pretext tasks, but a main challenge is how to design a joint learning framework to balance each component. 
}

\subsection{Taxonomy of Self-Supervised Training Schemes}\label{subsec: training schemes}
According to the relationship among graph encoders, self-supervised pretext tasks, and downstream tasks, we investigate three types of graph self-supervised training schemes: Pre-training and Fine-tuning (PF), Joint Learning (JL), and Unsupervised Representation Learning (URL). Brief pipelines of them are given in Fig. \ref{fig:learning_scheme}.

\noindent \textbf{Pre-training and Fine-tuning (PF).} 
In PF scheme, the encoder $f_{\theta}$ is first pre-trained with pretext tasks on pre-training datasets, which can be viewed as an initialization for the encoder's parameters. After that, the pre-trained encoder $f_{\theta_{init}}$ is fine-tuned together on fine-tuning datasets (with labels) with a downstream decoder $q_{\psi}$ under the supervision of specific downstream tasks. Note that the datasets for pre-training and fine-tuning could be the same or different. The formulation of PF scheme is defined as follows:% 
\begin{equation}
\begin{aligned}
{\theta^{*},\phi^{*}} &= \mathop{\arg\min}\limits_{{\theta}, {\phi}}\mathcal{L}_{ssl}\left( f_{\theta}, p_{\phi}, \mathcal{D} \right), \\
{\theta^{**},\phi^{*}} &= \mathop{\arg\min}\limits_{{\theta^{*}}, {\psi}}\mathcal{L}_{sup}\left( p_{\theta^{*}}, q_{\psi}, \mathcal{G}, y \right).
\end{aligned}
\label{eq: PT&FT}
\end{equation}

\begin{figure*}[tp]
	\centering
    \subfloat[Graph Completion is a representative approach of feature generation-based graph SSL. The features of certain nodes are masked and then fed into the model, and the learning target is to reconstruct the masked features. {An} MSE loss is used to recover the features.
    ]{\includegraphics[width=3.2in]{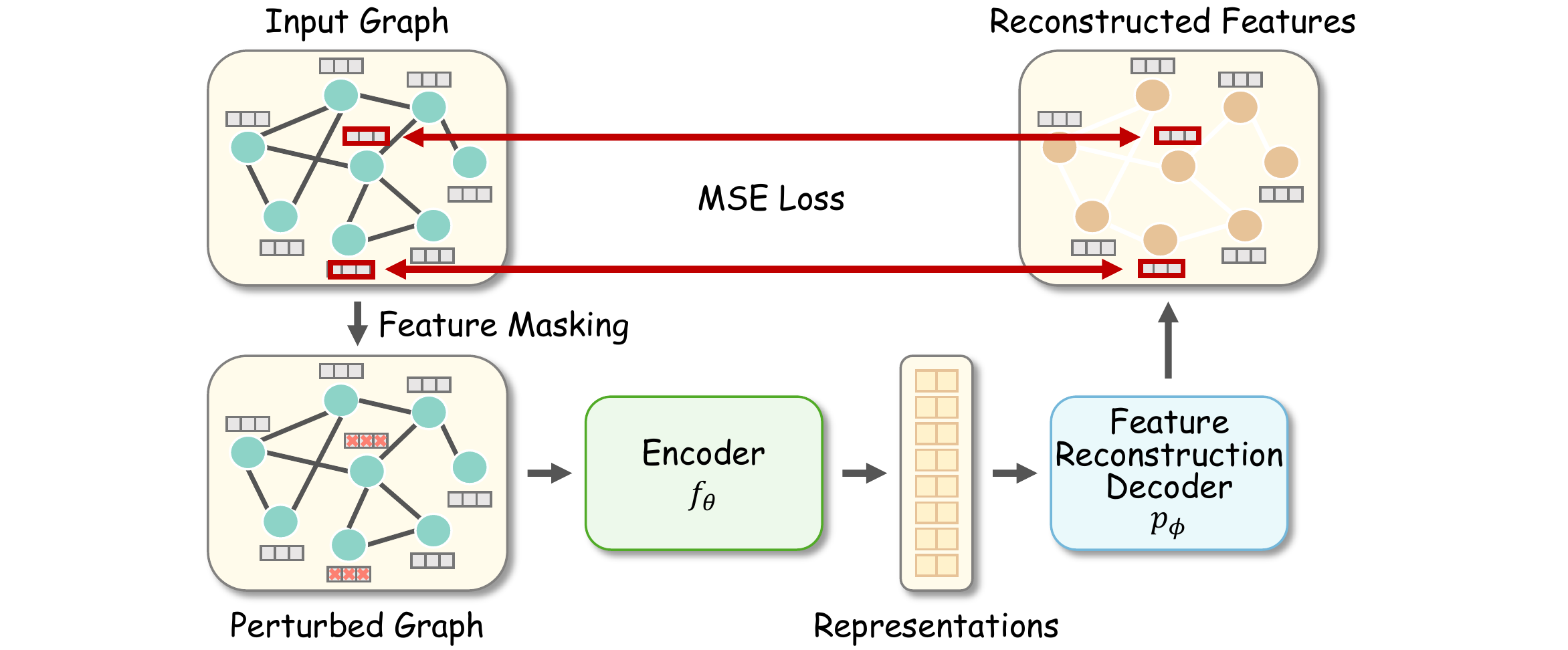}%
    \label{fig:gen_feat}
    }
    \hfill
	\subfloat[The objective of Denoising Link Reconstruction is to rebuild the masked edges. A binary cross-entropy (BCE) loss is employed to train the model where existing edges are the positive samples and the unrelated node pairs are the negative samples. A negative sampling (Neg. Samp.) strategy is used to balance the classes.%
	]{\includegraphics[width=3.2in]{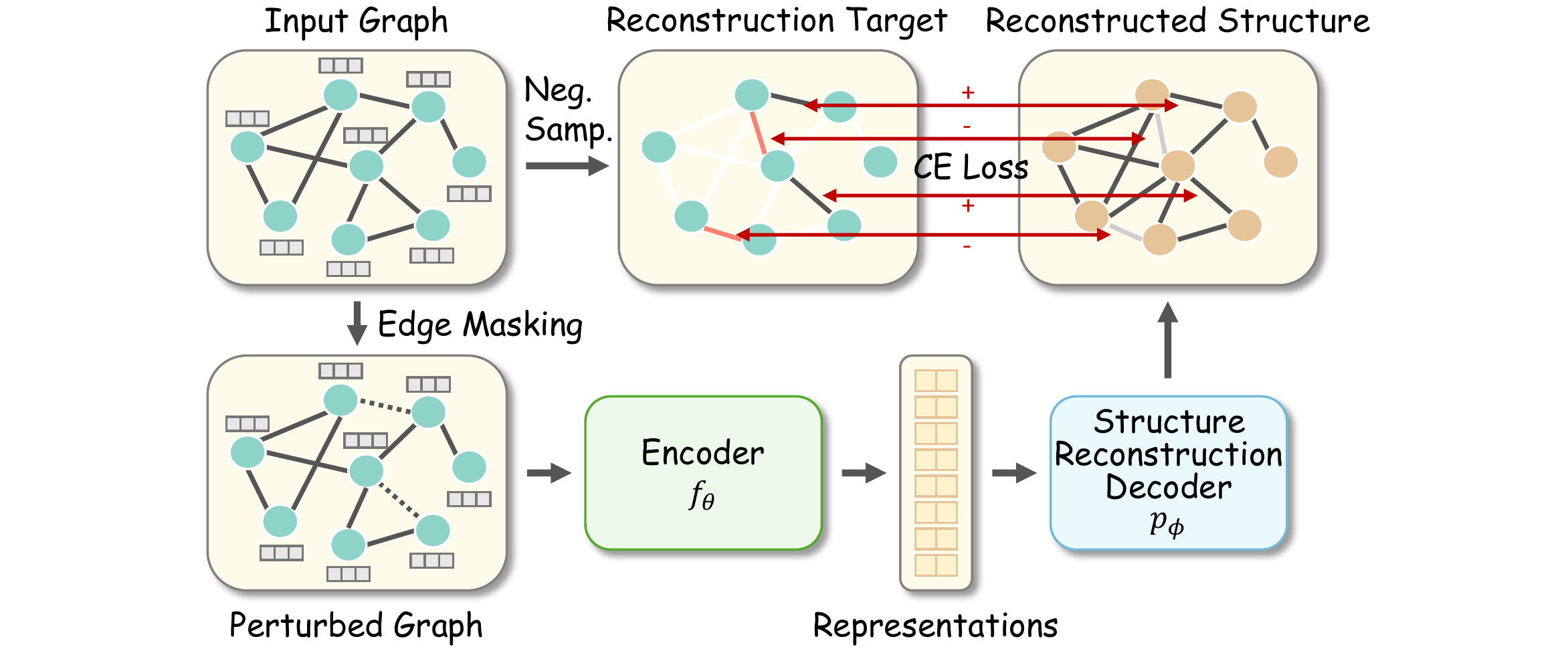}%
	\label{fig:gen_str}}
	\caption{Examples of two categories of generation-based methods: Graph Completion and Denoising Link Reconstruction.}
	\label{fig:generation}
	\vspace{-3mm}
\end{figure*}

\noindent \textbf{Joint Learning (JL).} 
In JL scheme, the encoder is jointly trained with the pretext and downstream tasks. The loss function consists of both the self-supervised and downstream task loss functions, where a trade-off hyper-parameter $\alpha$ controls the contribution of self-supervision term. This can be considered as a kind of multi-task learning where the pretext task is served as a regularization of the downstream task:
\begin{equation}
{\theta^{*}}, {\phi^{*}}, {\psi^{*}} = \mathop{\arg\min}\limits_{{\theta}, {\phi}, {\psi}}\Big[\alpha\mathcal{L}_{ssl}\left( f_{\theta}, p_{\phi}, \mathcal{D} \right) + \mathcal{L}_{sup}\left( f_{\theta}, q_{\psi}, \mathcal{G}, y \right)\Big].
\label{eq: JL}
\end{equation}

\noindent \textbf{Unsupervised Representation Learning (URL).} 
The first stage of the URL scheme is similar to that of PF. The differences are: (1) In the second stage, the encoder's parameters are frozen (i.e., $\theta^{*}$) when the model is trained with the downstream task; (2) The training of two stages is performed on the same dataset. The formulation of URL is defined as: %
\begin{equation}
\begin{aligned}
{\theta^{*},\phi^{*}} &= \mathop{\arg\min}\limits_{{\theta}, {\phi}}\mathcal{L}_{ssl}\left( f_{\theta}, p_{\phi}, \mathcal{D} \right), \\
{\psi^{*}} &= \mathop{\arg\min}\limits_{{\psi}}\mathcal{L}_{sup}\left( f_{\theta^{*}}, q_{\psi}, \mathcal{G}, y \right).
\end{aligned}
\label{eq: URL}
\end{equation}

Compared with other schemes, URL is more challenging since there is no supervision during the encoder training.

\subsection{Taxonomy of Downstream Tasks}\label{subsec: downstream}

According to the scale of prediction target, we divide downstream tasks into node-, link-, and graph-level tasks. 
Specifically, node-level tasks aim to predict the property of nodes in graph(s) according to node representations. Link-level tasks infer the property of edges or pairs of nodes, where downstream decoders map the embeddings of two nodes into link-level predictions. Besides, graph-level tasks learn from a dataset with multiple graphs and forecast the property of each graph. Based on Equation (\ref{eq: downstrem task}), we provide the specific definitions of downstream decoders $q_{\psi}$, downstream objectives $\mathcal{L}_{sup}$, and downstream task labels $y$ of three types of tasks, which are detailed in Appendix \ref{appendix:downstream_tasks}.

\vspace{-3mm}
\section{Generation-based Methods} \label{sec:generation}

The generation-based methods aim to reconstruct the input data and use the input data as their supervision signals. The origin of this category of methods can be traced back to Autoencoder \cite{hinton2006reducing} which learns to compress data vectors into low-dimensional representations with the encoder network and then try to rebuild the input vectors with the decoder network. % 
Different from generic input data represented in vector formats, graph data are interconnected. As a result, generation-based graph SSL approaches often take the full graph or a subgraph as the model input, and reconstruct one of the components, i.e. feature or structure, individually.
According to the objects of reconstruction, we divide these works into two sub-categories: (1) \textit{feature generation} that learns to reconstruct the feature information of graphs, and (2) \textit{structure generation} that learns to reconstruct the topological structure information of graphs. The pipelines of two example methods are given in Fig. \ref{fig:generation}, and a summary of the generation-based works is illustrated in Table \ref{tab:summary_generation}. 

\begin{table*}[thp]
\caption{Main characteristics of generation-based graph SSL approaches. ``FG'' and ``SG'' mean ``Feature Generation'' and ``Structure Generation'', respectively. Missing values (``-'') in Input Data Perturbation indicate that the method takes the original graph data as input.}
\label{tab:summary_generation}
% \vspace{-3mm}
\centering
\begin{adjustbox}{width=2\columnwidth,center}
\begin{tabular}{lcccccc}
\toprule
\thead{Approach} & \thead{Pretext Task\\ Category} & \thead{Downstream \\Task Level} & \thead{Training \\Scheme} & \thead{Data Type\\ of Graph}   & \thead{Input Data  \\ Perturbation} & \thead{Generation \\ Target} \\ \bottomrule
Graph Completion \cite{you2020does}   &  FG   &  Node       &   PF/JL   &  Attributed  & Feature Masking   &  Node Feature   \\ \EOL
AttributeMask \cite{jin2020self}   &  FG   &  Node       &   PF/JL   &  Attributed  & Feature Masking   & PCA Node Feature   \\ \EOL
AttrMasking \cite{hu2019strategies}   &  FG   &  Node       &   PF   &  Attributed  & Feature Masking   &  Node/Edge Feature   \\ \EOL
MGAE \cite{wang2017mgae}   &  FG   &  Node       &   JL   &  Attributed  & Feature Noising   &  Node Feature   \\ \EOL
Corrupted Features Reconstruction \cite{manessi2020graph}   &  FG   &  Node       &   JL   &  Attributed  & Feature Noising   &  Node Feature   \\ \EOL
Corrupted Embeddings Reconstruction \cite{manessi2020graph}   &  FG   &  Node       &   JL   &  Attributed  & Embedding Noising   & Node Embedding   \\ \EOL
GALA \cite{park2019symmetric}   &  FG   &  Node/Link       &   JL   &  Attributed  & -   &  Node Feature   \\ \EOL
Autoencoding \cite{manessi2020graph}   &  FG   &  Node       &   JL   &  Attributed  & -   &  Node Feature   \\ \EOL
GAE/VGAE \cite{kipf2016variational}   &  SG   &  Link       &   URL   &  Attributed  & -   & Adjacency Matrix   \\ \EOL
SIG-VAE \cite{hasanzadeh2019semi}   &  SG   &  Node/Link       &   URL   &  \tabincell{c}{Plain/Attributed}  & -   & Adjacency Matrix   \\ \EOL
ARGA/ARVGA \cite{pan2018adversarially}  &  SG   &  Node/Link       &   URL   &  Attributed  & -   & Adjacency Matrix   \\ \EOL
SuperGAT \cite{kim2021how}  &  SG   &  Node       &   JL   &  Attributed  & -   & Partial Edge   \\ \EOL
Denoising Link Reconstruction \cite{hu2019pre}  &  SG   &  \tabincell{c}{Node/Link/Graph}       &   PF   &  Attributed  & Edge Masking   & Masked Edge   \\ \EOL
EdgeMask \cite{jin2020self}  &  SG   &  Node       &   PF/JL   &  Attributed  & Edge Masking   & Masked Edge   \\ \EOL
Zhu et al. \cite{zhu2020self}  &  SG   &  Node       &   PF   &  Attributed  & \tabincell{c}{Feature Masking/Edge Masking}   & Partial Edge   \\ \toprule
\end{tabular}
\end{adjustbox}
\vspace{-0.3cm}
\end{table*}

\subsection{Feature Generation}

Feature generation approaches learn by recovering feature information from the perturbed or original graphs.
Based on Equation (\ref{eq: generation}), the feature generation approaches can be further formalized as:

\begin{equation}
{\theta^{*},\phi^{*}} = \mathop{\arg\min}\limits_{{\theta}, {\phi}} {{\mathcal{L}_{mse}} \left( p_{\phi}\left( f_{\theta}\left( \tilde{\mathcal{G}} \right)\right), \hat{\mathbf{X}}\right)},
\label{eq: feat_generation}
\end{equation}

\noindent where $p_{\phi}(\cdot)$ is {the} decoder {for feature regression (e.g., a fully connected network that maps the representations to reconstructed features)}, ${\mathcal{L}_{mse}}$ is the Mean Squared Error (MSE) loss function, and $\hat{\mathbf{X}}$ is a general expression of various kinds of feature matrices, \textit{e.g.}, node feature matrix, edge feature matrix, or low-dimensional feature matrix.

To leverage the dependency between nodes, a representative branch of feature generation approaches follows the \textit{masked feature regression} strategy, which is motivated by image inpainting in CV domain \cite{pathak2016context_inpainting}. Specifically, the features of certain nodes/edges are masked with zero or specific tokens in the pre-processing phase. Then, the model tries to recover the masked features according to the unmasked information.
\methodHL{Graph Completion} \cite{you2020does} is a representative method. It first masks certain nodes of the input graph by removing their features. Then, the learning objective is to predict the masked node features from the features of neighboring nodes with a GCN {\cite{gcn_kipf2017semi}} encoder. We can consider Graph Completion as an implement of Equation (\ref{eq: feat_generation}) where $\hat{\mathbf{X}}=\mathbf{X}$ and $\tilde{\mathcal{G}}=({\mathbf{A}},\tilde{\mathbf{X}})$. 
Similarly, \methodHL{AttributeMask} \cite{jin2020self} aims to reconstruct the dense feature matrix processed by Principle Component Analysis (PCA) \cite{wold1987principal} ($\hat{\mathbf{X}}=PCA(\mathbf{X})$) instead of the raw features due to the difficulty of rebuilding high-dimensional and sparse features. %
\methodHL{AttrMasking} \cite{hu2019strategies} rebuilds not only node attributes but also the edge one, which can be written as $\hat{\mathbf{X}} = [\mathbf{X},\mathbf{X_{edge}}]$.%

Another branch of methods aims to generate features from noisy features. Inspired by denoising autoencoder \cite{vincent2010stacked}, \methodHL{MGAE} \cite{wang2017mgae} recovers raw features from noisy input features with each GNN layer. Here we also denote $\tilde{\mathcal{G}}=({\mathbf{A}},\tilde{\mathbf{X}})$ but here $\tilde{\mathbf{X}}$ is corrupted with random noise. Proposed in \cite{manessi2020graph}, \methodHL{Corrupted Features Reconstruction} and \methodHL{Corrupted Embeddings Reconstruction} aim to reconstruct raw features and hidden embeddings from corrupted features. 

Besides, directly rebuilding features from the clean data is also an available solution. \methodHL{GALA} \cite{park2019symmetric} trains a Laplacian smoothing-sharpening graph autoencoder model with the objective that rebuilds the raw feature matrix according to the clean input graph. Similarly, \methodHL{autoencoding} \cite{manessi2020graph} reconstructs the raw features from clean inputs. For these two methods, we can formalize that $\tilde{\mathcal{G}}=({\mathbf{A}},\mathbf{X})$ and $\hat{\mathbf{X}}=\mathbf{X}$.

\subsection{Structure Generation}

Different from the feature generation approaches that rebuild the feature information,  structure generation approaches learn by recovering the structural information. In most cases, the objective is to reconstruct the adjacency matrix, since the adjacency matrix can briefly represent the topological structure of graphs. 
Based on Equation (\ref{eq: generation}), the structure generation methods can be formalized as follows:

\begin{equation}
{\theta^{*},\phi^{*}} = \mathop{\arg\min}\limits_{{\theta}, {\phi}} \mathcal{L}_{ssl} \Big( p_{\phi}\left( f_{\theta}\left( \tilde{\mathcal{G}} \right)\right) , {\mathbf{A}} \Big),
\label{eq: stru_generation}
\end{equation}

\noindent where $p_{\phi}(\cdot)$ is a decoder for structure reconstruction, and ${\mathbf{A}}$ is the (full or partial) adjacency matrix.

\methodHL{GAE} \cite{kipf2016variational} is the simplest instance of the structure generation method. In GAE, a GCN-based encoder first generates node embeddings $\mathbf{H}$ from the original graph ($\tilde{\mathcal{G}} = \mathcal{G}$). Then, an inner production function with sigmoid activation serves as its decoder to recover the adjacency matrix from $\mathbf{H}$. Since adjacency matrix $\mathbf{A}$ is usually binary and sparse, a BCE loss function is employed to maximize the similarity between the recovered adjacency matrix and the original one, where positive and negative samples are the existing edges ($\mathbf{A}_{i, j}=1$) and unconnected node pairs ($\mathbf{A}_{i, j}=0$), respectively. To avoid the imbalanced training sample problem caused by extremely sparse adjacency, two strategies can be used to prevent trivial solution: (1) re-weighting the terms with $\mathbf{A}_{i, j}=1$; or (2) sub-sampling terms with $\mathbf{A}_{i, j}=0$.%

As a classic learning paradigm, GAE has a series of derivative works. \methodHL{VGAE} \cite{kipf2016variational} further integrates the idea of variational autoencoder \cite{kingma2013auto} into GAE. It employs an inference model-based encoder that estimates the mean and deviation with two parallel output layers and uses Kullback-Leibler divergence between the prior distribution and the estimated distribution. Following VGAE, \methodHL{SIG-VAE} \cite{hasanzadeh2019semi} considers hierarchical variational inference to learn more generative representations for graph data. \methodHL{ARGA/ARVGA} \cite{pan2018adversarially} regularizes the GAE/VGAE model with generative adversarial networks (GANs) \cite{goodfellow2014generative}. Specifically, a discriminator is trained to distinguish the fake and real data, which forces the distribution of latent embeddings closer to the Gaussian prior. \methodHL{SuperGAT} \cite{kim2021how} further extends this idea to every layers in the encoder. Concretely, it rebuilds the adjacency matrix from the latent representations of every layer in the encoder. \looseness-1

Instead of rebuilding the full graph, another solution is to reconstruct the masked edges. 
\methodHL{Denoising Link Reconstruction} \cite{hu2019pre} randomly drops existing edges to obtain the perturbed graph $\tilde{\mathcal{G}}$. Then, the model aims to recover the discarded connections with a pairwise similarity-based decoder trained by a BCE loss. 
\methodHL{EdgeMask} \cite{jin2020self} also has a similar perturbation strategy, where a non-parametric MAE function minimizes the difference between the embeddings of two connected nodes. 
\methodHL{Zhu et al.} \cite{zhu2020self} apply two perturbing strategies, i.e. Randomly Removing Links and Randomly Covering Features, to the input graph ($\tilde{\mathcal{G}}=(\tilde{\mathbf{A}},\tilde{\mathbf{X}})$), while its target is to recover the masked link by a decoder. 

\noindent {\textbf{Discussion.} Due to the different learning targets, two branches of generation-based methods have distinct designs of {the} decoder and loss {functions}. The learned representations by structure generation usually contain more node pair-level information since structure generation focuses on edge reconstruction; by contrary, feature generation methods often capture node-level knowledge.
}

\begin{figure*}[tp]
	\centering
    \subfloat[Auxiliary property classification methods extract discrete properties as pseudo labels, and the pretext decoder is used to predict the classification results. A CE loss function is used to train the models. Two types of properties (clustering-based and pair relation-based) can be used to define the pseudo labels.
    ]{\includegraphics[height=1.6in]{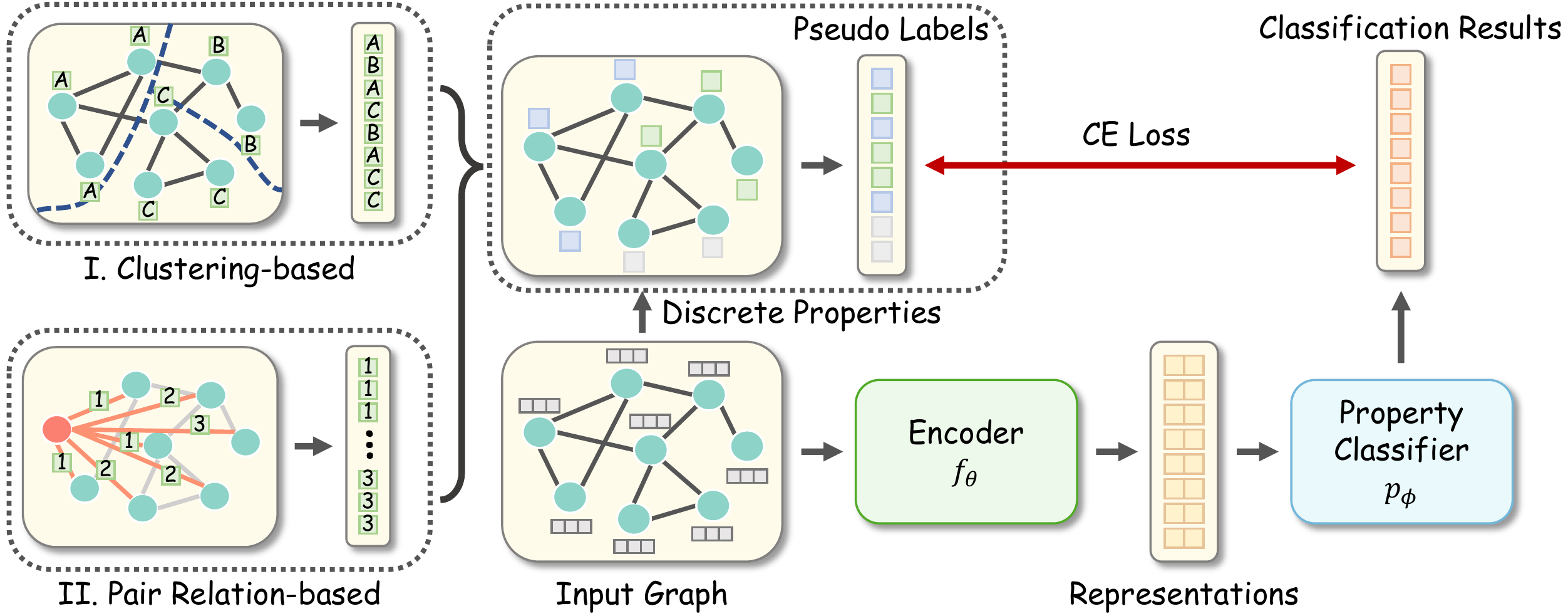}%
    \label{fig:apc}
    }
    \hfill
	\subfloat[Auxiliary property regression methods aim to predict the continuous auxiliary properties with the decoder, where models are trained by the MSE loss.
	]{\includegraphics[height=1.6in]{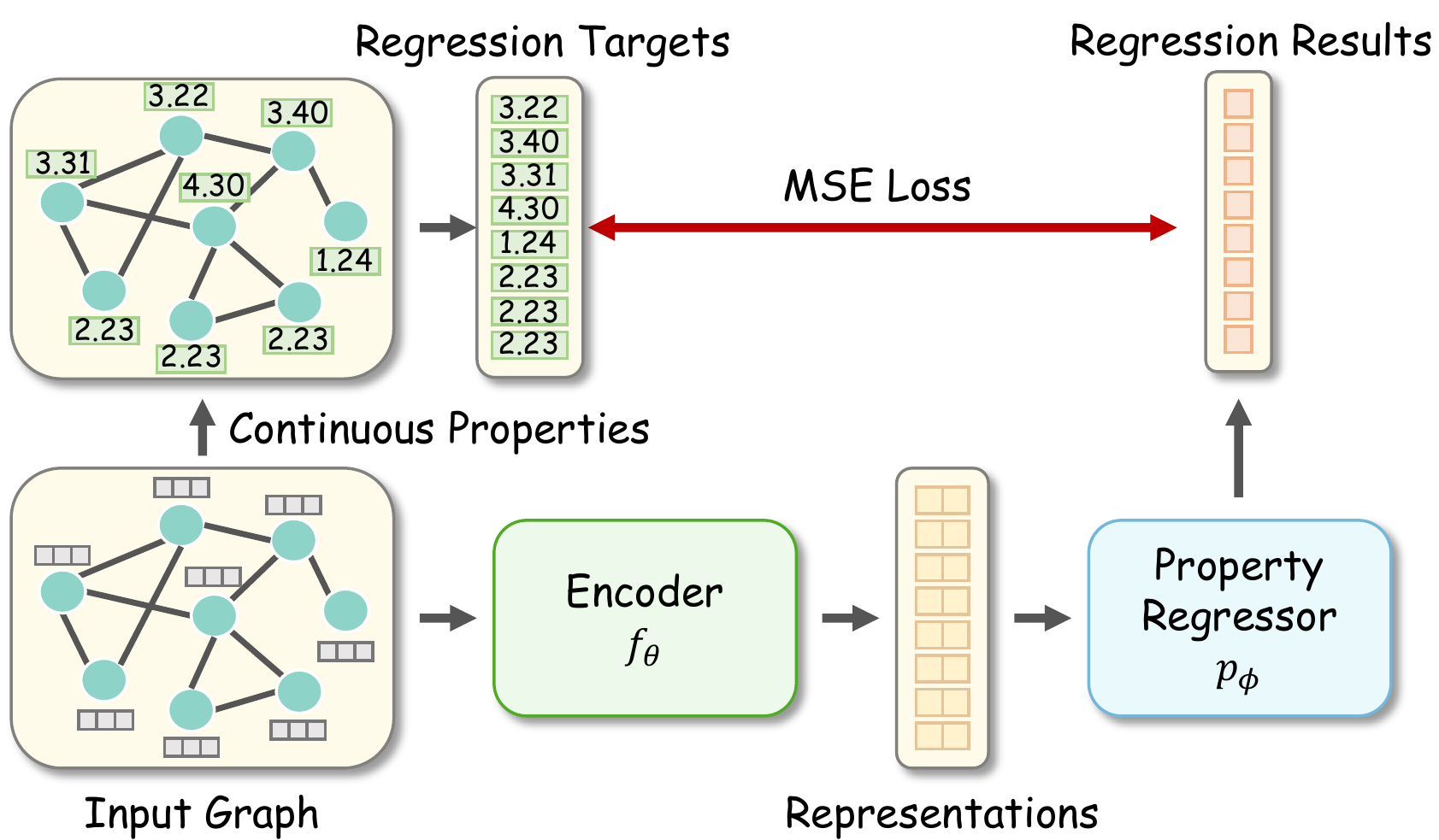}%
	\label{fig:apr}}
	\caption{Two categories of auxiliary property-based graph SSL.}
	\label{fig:ap}
\end{figure*}

\vspace{-3mm}
\section{Auxiliary Property-based Methods} \label{sec:app}

The auxiliary property-based methods acquire supervision signals from the node-, link- and graph- level properties which can be obtained from the graph data freely. These methods have a similar training paradigm with supervised learning since both of them learn with ``sample-label'' pairs. Their difference lies in how the label is obtained: In supervised learning, the manual label is human-annotated which often needs expensive costs; in auxiliary property-based SSL, the pseudo label is self-generated automatically without any cost. 

Following the general taxonomy of supervised learning, we divide auxiliary property-based methods into two sub-categories: (1) \textit{auxiliary property classification} which leverages classification-based pretext tasks to train the encoder and (2) \textit{auxiliary property regression} which performs SSL via regression-based pretext tasks. Fig. \ref{fig:ap} provides the pipelines of them, and Table \ref{tab:summary_property} summarizes the auxiliary property-based methods.

\begin{table*}[thp]
\caption{Main characteristics of auxiliary property-based graph SSL approaches. ``CAPC'', ``PAPC'' and ``APR'' mean clustering-based auxiliary property classification, pair relation-based auxiliary property classification and auxiliary property regression, respectively.}
\label{tab:summary_property}
% \vspace{-2mm}
\centering
\begin{adjustbox}{width=1.95\columnwidth,center}
\begin{tabular}{lcccccc}
\toprule
\thead{Approach} & \thead{Pretext Task\\ Category} & \thead{Downstream \\Task Level} & \thead{Training \\Scheme} & \thead{Data Type\\ of Graph}   & \thead{Property  \\ Level} & \thead{Mapping Function} \\ \bottomrule
Node Clustering \cite{you2020does}   &  CAPC   &  Node       &   PF/JL   &  Attributed  & Node   &  Feature-based Clustering   \\ \EOL
M3S \cite{sun2020multi}   &  CAPC   &  Node       &   JL   &  Attributed  & Node   &  Feature-based Clustering   \\ \EOL
Graph Partitioning \cite{you2020does}   &  CAPC   &  Node       &   PF/JL   &  Attributed  & Node   &  Structure-based Clustering   \\ \EOL
Cluster Preserving \cite{hu2019pre}   &  CAPC   &  Node/Link/Graph       &   PF   &  Attributed  & Node   &  Structure-based Clustering   \\ \EOL
CAGNN \cite{zhu2020cagnn}   &  CAPC   &  Node       &   URL   &  Attributed  & Node   &  \tabincell{c}{Feature-based Clustering \\ with Structural Refinement}   \\ \EOL
S$^{2}$GRL \cite{peng2020self}   &  PAPC   &  Node/Link       &   URL   &  Attributed  & Node Pair   &  Shortest Distance Function   \\ \EOL
PairwiseDistance \cite{jin2020self}   &  PAPC   &  Node       &   PF/JL   &  Attributed  & Node Pair   &  Shortest Distance Function   \\ \EOL
Centrality Score Ranking \cite{hu2019pre}   &  PAPC   &    Node/Link/Graph       &   PF    &  Attributed  & Node Pair   &  Centrality Scores Comparison   \\ \EOL
NodeProperty \cite{jin2020self}   &  APR   &    Node       &   PF/JL   &  Attributed   & Node   &  Degree Calculation  \\\EOL
Distance2Cluster \cite{jin2020self}   &  APR   &    Node       &   PF/JL   &  Attributed   & Node Pair   &  Distance to Cluster Center  \\\EOL
PairwiseAttrSim \cite{jin2020self}   &  APR   &    Node       &   PF/JL   &  Attributed   & Node Pair   &  Cosine Similarity of Feature  \\\EOL
SimP-GCN \cite{jin2021node}   &  APR   &    Node       &   JL   &  Attributed   & Node Pair   &  Cosine Similarity of Feature  \\\toprule
\end{tabular}
\end{adjustbox}
\vspace{-0.3cm}
\end{table*}

\subsection{Auxiliary Property Classification}

Borrowing the training paradigm from supervised classification tasks, the methods of auxiliary property classification create discrete pseudo labels automatically, build a classifier as the pretext decoder, and use a cross entropy (CE) loss $\mathcal{L}_{ce}$ to train the model. Originated from Equation (\ref{eq: auxiliary}), we provide the formalization of this branch of methods as:

\begin{equation}
{\theta^{*},\phi^{*}} = \mathop{\arg\min}\limits_{{\theta}, {\phi}}\mathcal{L}_{ce}\Big( p_{\phi}\big( f_{\theta}(\mathcal{G})\big),c\Big),
\label{eq: auxiliary_classification}
\end{equation}

\noindent where $p_{\phi}$ is the neural network classifier-based decoder which outputs a $k$-dimensional probability vector ($k$ is the number of classes), and $c \in \mathcal{C} = \{ c_1, \cdots, c_k \}$ is the corresponding pseudo label which belongs to a discrete and finite label set $\mathcal{C}$.
According to the definition of pseudo label set $\mathcal{C}$, we further construct two sub-categories under auxiliary property classification, i.e., clustering-based and pair relation-based methods. 

\subsubsection{Clustering-based Methods}

A promising way to construct pseudo label is to divide nodes into different clusters according to their attributive or structural characteristics. To achieve that, a mapping function $\Omega: \mathcal{V} \to \mathcal{C}$ is introduced to acquire the pseudo label for each node, which is built on specific unsupervised clustering/partitioning algorithms \cite{karypis1995multilevel,kang2021multi,caron2018deep,kang2021structured}. Then, the learning objective is to classify each node into its corresponding cluster. Following Equation (\ref{eq: auxiliary_classification}), the learning objective is refined as:

\begin{equation}
{\theta^{*},\phi^{*}} = \mathop{\arg\min}\limits_{{\theta}, {\phi}} \frac{1}{\lvert \mathcal{V} \rvert}  \sum_{v_i \in \mathcal{V}}  \mathcal{L}_{ce}\Big( p_{\phi}\big([ f_{\theta}(\mathcal{G})]_{v_i}\big),\Omega (v_i) \Big),
\label{eq: auxiliary_classification_cluster}
\end{equation}

\noindent where $[\cdot]_{v_i}$ is the picking function that extracts the representation of $v_i$. 

\methodHL{Node Clustering} \cite{you2020does} is a representative approach that utilizes attributive information to generate pseudo labels. Specifically, it leverages a feature-based clustering algorithm (which is an instance of $\Omega$) taking $\mathbf{X}$ as input to divide node set into $k$ clusters, and each cluster indicates a pseudo label for classification. {The intuition behind Node Clustering is that nodes with similar features tend to have consistent semantic properties.} \methodHL{M3S} \cite{sun2020multi} introduces a multi-stage self-training mechanism for SSL using DeepCluster \cite{caron2018deep} algorithm. In each stage, it first runs K-means clustering on node embedding $\mathbf{H}$. After that, an alignment is executed to map each cluster to a class label. Finally, the unlabeled nodes with high confidence are given the corresponding (pseudo) labels and used to train the model. In M3S, $\mathcal{C}$ is borrowed from the manual label set $\mathcal{Y}$, and $\Omega$ is composed of the K-means and alignment algorithms. 

In addition to feature-based clustering, \methodHL{Graph Partitioning} \cite{you2020does} divides the nodes according to the structural characteristics of nodes. Concretely, it groups nodes into multiple subsets by minimizing the connections across subsets \cite{karypis1995multilevel}, defining $\Omega$ as the graph partitioning algorithm. 
\methodHL{Cluster Preserving} \cite{hu2019pre} first leverages graph clustering algorithm \cite{schaeffer2007graph} to acquire non-overlapping clusters, and then calculates the representation of each cluster via an attention-based aggregator. After that, a vector representing the similarities between each node and the cluster representations is assigned as the soft pseudo label for each node. %
Besides, \methodHL{CAGNN} \cite{zhu2020cagnn} first runs feature-based clusters to generate pseudo labels and then refines the clusters by minimizing inter-cluster edges, which absorbs the advantages of both attributive and structural clustering algorithms.

\subsubsection{Pair Relation-based Methods}

Apart from the clustering and graph properties, an alternative supervision signal is the relationship between each pair of nodes within a graph. In these methods, the input of the decoder is not a single node or graph but a pair of nodes. A mapping function $\Omega: \mathcal{V} \times \mathcal{V} \to \mathcal{C}$ is utilized to define the pseudo label according to pair-wise contextual relationship. We write the objective function as:

\begin{equation}
{\theta^{*},\phi^{*}} = \mathop{\arg\min}\limits_{{\theta}, {\phi}} \frac{1}{\lvert \mathcal{P} \rvert}  \sum_{v_i,v_j \in \mathcal{P}}  \mathcal{L}_{ce}\Big( p_{\phi}\big([ f_{\theta}(\mathcal{G})]_{v_i,v_j}\big),\Omega (v_i,v_j) \Big),
\label{eq: auxiliary_classification_pair}
\end{equation}

\noindent where $\mathcal{P} \subseteq \mathcal{V} \times \mathcal{V}$ is the node pair set defined by specific pretext tasks, and $[\cdot]_{v_i,v_j}$ is the picking function that extracts {and concatenates} the node representations of $v_i$ and $v_j$.

Some approaches regard the distance between two nodes as the auxiliary property. For instance, \methodHL{S$^{2}$GRL} \cite{peng2020self} learns by predicting the shortest path between two nodes. 
Specifically, the label for a pair of nodes is defined as the shortest distance between them. Formally, we can write the mapping function as $\Omega (v_i,v_j) = \operatorname{dist}(v_i,v_j)$.
The decoder is built to measure the interaction between pairs of nodes, which is defined as an element-wise distance between two embedding vectors. The node pair set $\mathcal{P}$ collects all possible node pairs including the combination of all nodes with their $1$ to $K$ hops neighborhoods. \methodHL{PairwiseDistance} \cite{jin2020self} has a very similar learning target and decoder with S$^{2}$GRL, but introduces an upper bound of distance, which can be represented as $\Omega (v_i,v_j) = \operatorname{max} (\operatorname{dist}(v_i,v_j),4)$. %

\methodHL{Centrality Score Ranking} \cite{hu2019pre} presents a pretext task that predicts the relative order of centrality scores between a pair of nodes. For each node pair $(v_i,v_j)$, it first calculates four types of centrality scores $s_i,s_j$ (eigencentrality, betweenness, closeness, and subgraph centrality), and then creates its pseudo label by comparing the value of $s_i$ and $s_j$. We formalize the mapping function as: $\Omega (v_i,v_j) = \mathbb{I}(s_i > s_j)${, where $\mathbb{I}(\cdot)$ is the identity function.}  %

\subsection{Auxiliary Property Regression}

Auxiliary property regression approaches construct the pretext tasks on {predicting} extensive numerical properties of graphs. Compared to auxiliary property classification, the most significant difference is that the auxiliary properties are continuous values within a certain range instead of discrete pseudo labels in a limited set. We refine Equation (\ref{eq: auxiliary}) into a {regression} version:

\begin{equation}
{\theta^{*},\phi^{*}} = \mathop{\arg\min}\limits_{{\theta}, {\phi}} {{\mathcal{L}_{mse}} \left( p_{\phi}\big( f_{\theta}(\mathcal{G})\big), c \right)},
\label{eq: auxiliary_regression}
\end{equation}

\noindent where ${\mathcal{L}_{mse}}$ is the MSE loss function for regression, and $c \in \mathbb{R}$ is a continuous property value.

\methodHL{NodeProperty} \cite{jin2020self} is a node-level pretext task that predicts the property for each node. The available choices of node properties include their degree, local node importance, and local clustering coefficient. Taking node degree as an example, the objective function is illustrated as follows: 

\begin{equation}
{\theta^{*},\phi^{*}} = \mathop{\arg\min}\limits_{{\theta}, {\phi}} \frac{1}{\lvert \mathcal{V} \rvert}  \sum_{v_i \in \mathcal{V}}  {{\mathcal{L}_{mse}} \left( p_{\phi}\big( [f_{\theta}(\mathcal{G})\big)]_{v_i}, \Omega(v_i) \right)},
\label{eq: auxiliary_regression_node}
\end{equation}

\noindent where $\Omega(v_i)=\sum_{j=1}^{n} \mathbf{A}_{i j}$ is the mapping function that calculates the degree of node $v_i$. %
\methodHL{Distance2Cluster} \cite{jin2020self} aims to regress the distances from each node to predefined graph clusters. 
Specifically, it first partitions the graph into several clusters with the METIS algorithm \cite{karypis1998fast} and defines the node with the highest degree within each cluster as its cluster center. Then, the target is to predict the distances between each node and all cluster centers.

Another type of methods take the pair-wise property as their regression targets. For instance, the target of \methodHL{PairwiseAttrSim} \cite{jin2020self} is to predict the feature similarity of two nodes according to their embeddings. We formalize its objective function as follows:  

\begin{equation}
{\theta^{*},\phi^{*}} = \mathop{\arg\min}\limits_{{\theta}, {\phi}} \frac{1}{\lvert \mathcal{P} \rvert}  \sum_{v_i,v_j \in \mathcal{P}}  {{\mathcal{L}_{mse}} \left(  p_{\phi}\big([ f_{\theta}(\mathcal{G})]_{v_i,v_j}\big), \Omega (v_i,v_j)  \right)},
\label{eq: auxiliary_regression_pair}
\end{equation}

\noindent where mapping function $\Omega (v_i,v_j) = \operatorname{cosine} (\mathbf{x}_i, \mathbf{x}_j)$ is the cosine similarity of raw features. In PairwiseAttrSim, the node pairs with the highest similarity and dissimilarity are selected to form the node pair set $\mathcal{P}$. Similar to PairwiseAttrSim, \methodHL{SimP-GCN} \cite{jin2021node} also considers predicting the cosine similarity of raw features as a self-supervised regularization for downstream tasks.% 

\noindent {\textbf{Discussion.} As we can observe, the auxiliary property classification methods are more diverse than the regression methods, since the discrete pseudo labels can be acquired by various algorithms. In future works, more continuous properties are expected to be leveraged for regression methods. 
}

\vspace{-3mm}
\section{Contrast-based Methods} \label{sec:contrast}

The contrast-based methods are built on the idea of mutual information (MI) maximization \cite{hjelm2018learning}, which learns by predicting the agreement between two augmented instances. Specifically, the MI between graph instances with the similar semantic information (i.e., positive samples) is maximized, while the MI between those with unrelated information (i.e., negative samples) is minimized.
Similar to the visual domain \cite{chen2020simple, Tian2020ContrastiveMC}, there exist various graph augmentations and contrastive pretext tasks on multiple {granularities} to enrich the supervision signals.

Following the taxonomy of contrast-based graph SSL {defined in Section 3.2.3}, we survey this branch of methods from three perspectives: (1) \textit{Graph augmentations} that generate various graph instances; (2) \textit{Graph contrastive learning} which forms various contrastive pretext tasks on the non-Euclidean space; (3) \textit{Mutual information estimation} that measures the MI between instances and forms the contrastive learning objective together with specific pretext tasks.

\subsection{Graph Augmentations} \label{subsec: augmentation}
\begin{figure}[tbp]
	\centering
	\includegraphics[width=0.4\textwidth]{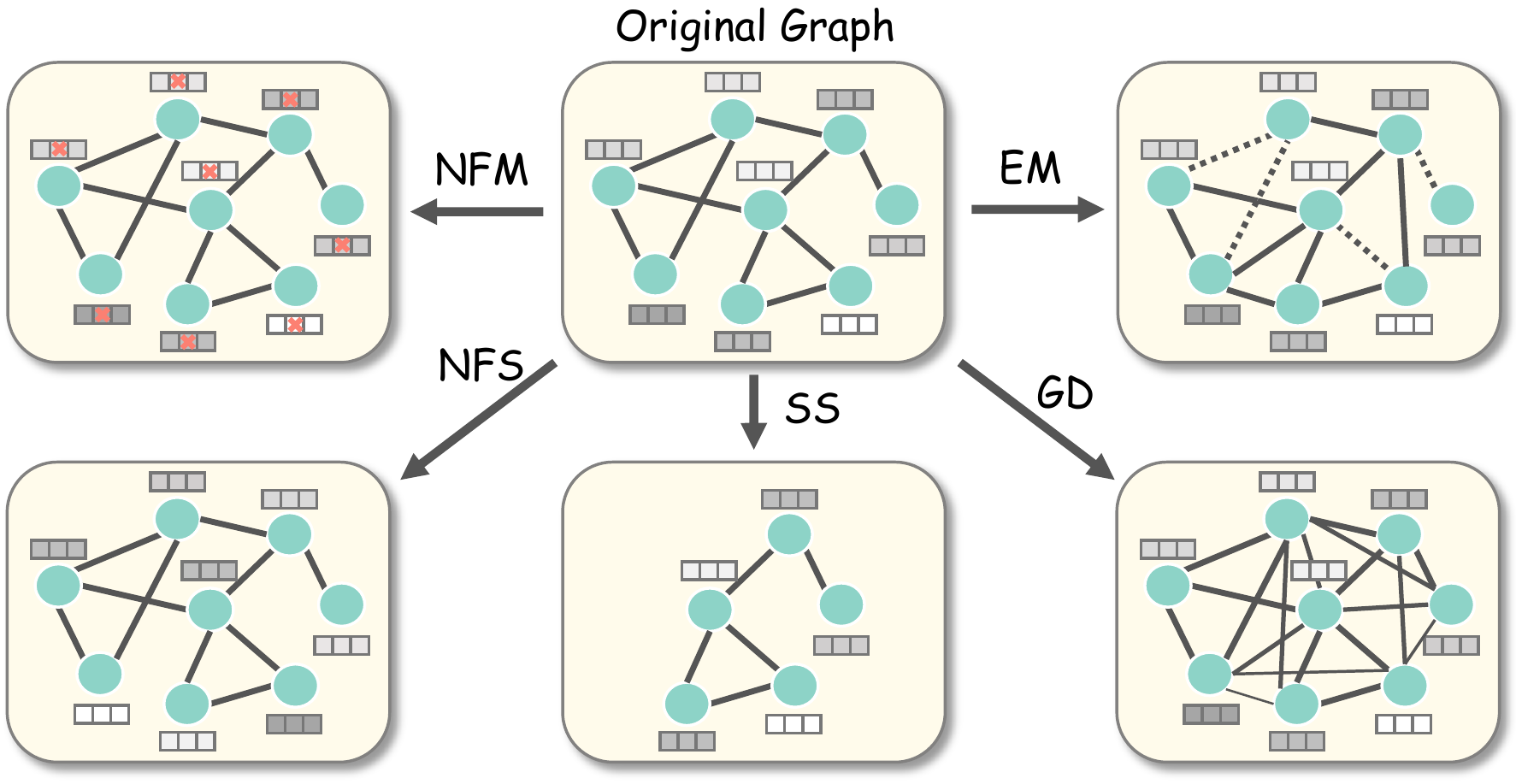}
	\caption{Brief examples of five types of common graph augmentations, including Node Feature Masking (NFM), Node Feature Shuffle (NFS), Edge Modification (EM), Graph Diffusion (GD), and Subgraph Sampling (SS).}
	\label{fig:augmentation}
	\vspace{-3mm}
\end{figure}

Recent success of contrastive learning on the visual domain relies heavily on well-crafted image augmentations, which reveals that data augmentations benefit the model to explore richer underlying semantic information by making pretext tasks more challenging to solve \cite{wang2021contrastive}. However, due to the nature of graph-structured data, it is difficult to apply the augmentations from the Euclidean to the non-Euclidean space directly. Motivated by image augmentations (\textit{e.g.}, image cutout and cropping \cite{chen2020simple}), existing graph augmentations can be categorized into three types: \textit{attributive-based}, \textit{topological-based}, and the combination of both (i.e., \textit{hybrid augmentations}). The examples of five representative augmentation strategies are demonstrated in Fig. \ref{fig:augmentation}.
Formally, given a graph $\mathcal{G}$, we define the $i$-th augmented graph instance as $\tilde{\mathcal{G}}^{(i)}=t_i(\mathcal{G})$, where $t_i \sim \tau$ is a selected graph augmentation and $\tau$ is a set of available augmentations.

\subsubsection{Attributive augmentations}
This category of augmentations is typically placed on node attributes. Given $\mathcal{G}=(\mathbf{A}, \mathbf{X})$, the augmented graph is represented as:
\begin{equation}
\tilde{\mathcal{G}}^{(i)}=(\mathbf{A}, \tilde{\mathbf{X}}^{(i)})=\big(\mathbf{A}, t_i(\mathbf{X})\big),
\label{eq: attributive augmentation}
\end{equation}
where $t_i(\cdot)$ is placed on the node feature matrix only, and $\tilde{\mathbf{X}}^{(i)}$ denotes the augmented node features. Specifically, attributive augmentations have two variants. The first type is \textbf{Node feature masking (NFM)} \cite{jin2020self, hu2019strategies, you2020graph, zhu2020deep, Jin2021MultiScaleCS}, which randomly masks the features of a portion of nodes within the given graph. In particular, we can completely (i.e., row-wisely) mask selected feature vectors with zeros \cite{jin2020self, hu2019strategies}, or partially (i.e., column-wisely) mask a number of selected feature channels with zeros \cite{zhu2020deep, Jin2021MultiScaleCS}. We formulate the node feature masking operation as:
\begin{equation}
t_i(\mathbf{X})=\mathbf{M} \circ \mathbf{X},
\label{eq: node feature masking}
\end{equation}
where $\mathbf{M}$ is the masking matrix with the same shape of $\mathbf{X}$, and $\circ$ denotes the Hadamard product. For a given masking matrix, its elements have been initialized to one and masking entries are assigned to zero. In addition to randomly sampling a masking matrix $\mathbf{M}$, we can also calculate it adaptively \cite{zhu2020graph, you2021graph}. For example, \methodHL{GCA} \cite{zhu2020graph} keeps important node features unmasked while assigning a higher masking probability for those unimportant nodes, where the importance is measured by node centrality.% 

On the other hand, instead of masking a part of the feature matrix, \textbf{node feature shuffle (NFS)} \cite{velivckovic2018deep, opolka2019spatio, ren2019heterogeneous} partially and row-wisely perturbs the node feature matrix. In other words, several nodes in the augmented graph are placed to other positions when compared with the input graph, as formulated below:
\begin{equation}
t_i(\mathbf{X})=[\mathbf{X}]_{\widetilde{\mathcal{V}}},
\label{eq: node feature shuffle}
\end{equation}
where $[\cdot]_{v_i}$ is a picking function that indexes the feature vector of $v_i$ from the node feature matrix, and $\widetilde{\mathcal{V}}$ denotes the partially shuffled node set.

\subsubsection{Topological augmentations}
Graph augmentations from the structural perspectives mainly work on the graph adjacency matrix, which is formulated as follows:
\begin{equation}
\tilde{\mathcal{G}}^{(i)}=( \tilde{\mathbf{A}}^{(i)}, \mathbf{X})=\big(t_i(\mathbf{A}), \mathbf{X}\big),
\label{eq: topological augmentation}
\end{equation}
where $t_i(\cdot)$ is typically placed on the graph adjacency matrix. For this branch of methods, \textbf{edge modification (EM)} \cite{zhu2020self, you2020graph, hu2020gpt, zhang2020iterative, Jin2021MultiScaleCS, zeng2020contrastive} is one of the most common approaches, which partially perturbs the given graph adjacency by randomly dropping and inserting a portion of edges. We define this process as follows:
\begin{equation}
t_i(\mathbf{A})= \mathbf{M_1} \circ \mathbf{A} + \mathbf{M_2} \circ (1-\mathbf{A}),
\label{eq: edge modification}
\end{equation}
where $\mathbf{M}_1$ and $\mathbf{M}_2$ are edge dropping and insertion matrices. Specifically, $\mathbf{M}_1$ and $\mathbf{M}_2$ are generated by randomly masking a portion of elements with the value equal to one in $\mathbf{A}$ and $(1-\mathbf{A})$. Similar to node feature masking, $\mathbf{M}_1$ and $\mathbf{M}_2$ can also be calculated adaptively \cite{zhu2020graph}. %
Furthermore, edge modification matrices can be generated based on adversarial learning \cite{jovanovic2021towards, suresh2021adversarial}, which increases the robustness of learned representations.

Different from the edge modification, \textbf{graph diffusion (GD)} \cite{klicpera2019diffusion} is another type of structural augmentations \cite{hassani2020contrastive, Jin2021MultiScaleCS}, which injects the global topological information to the given graph adjacency by connecting nodes with their indirectly connected neighbors with calculated weights:
\begin{equation}
t_i(\mathbf{A})= \sum^{\infty}_{k=0}{\Theta}_{k}\mathbf{T}^{k},
\label{eq: graph diffusion}
\end{equation}
where ${\Theta}$ and $\mathbf{T}$ are weighting coefficient and transition matrix, respectively. Specifically, the above diffusion formula has two instantiations \cite{hassani2020contrastive}. Let $\Theta_{k}=\frac{e^{- \iota }t^k}{k!}$ and $\mathbf{T}=\mathbf{A}\mathbf{D}^{-1}$, we have the heat kernel-based graph diffusion:
\begin{equation}
t_i(\mathbf{A})= \exp{(\iota\mathbf{A}\mathbf{D}^{-1}-\iota)},
\label{eq: heat kernel-based diffusion}
\end{equation}
where $\iota$ denotes the diffusion time. Similarly, the Personalized PageRank-based graph diffusion is defined below by letting $\Theta_{k}=\beta(1-\beta)^{k}$ and $\mathbf{T}=\mathbf{D}^{-\frac{1}{2}}\mathbf{A}\mathbf{D}^{-\frac{1}{2}}$:
\begin{equation}
t_i(\mathbf{A})= \beta\big(\mathbf{I}-(1-\beta)\mathbf{D}^{-\frac{1}{2}}\mathbf{A}\mathbf{D}^{-\frac{1}{2}}\big),
\label{eq: PPR-based diffusion}
\end{equation}
where $\beta$ denotes the tunable teleport probability.

\subsubsection{Hybrid augmentations}
It is worth noting that a given graph augmentation may involve not only the attributive but also the topological augmentations simultaneously, where we define it as the hybrid augmentation and formulate as:
\begin{equation}
\tilde{\mathcal{G}}^{(i)}=( \tilde{\mathbf{A}}^{(i)}, \tilde{\mathbf{X}}^{(i)})=\big(t_i(\mathbf{A}, \mathbf{X})\big).
\label{eq: hybrid augmentation}
\end{equation}

In such a case, the augmentation $t_i(\cdot)$ is placed on both the node feature and graph adjacency matrices. \textbf{Subgraph sampling (SS)} \cite{zeng2020contrastive, hu2019strategies, hassani2020contrastive, jiao2020sub} is a typical hybrid graph augmentation which is similar to image cropping. Specifically, it samples a portion of nodes and their underlying linkages as augmented graph instances:

\begin{equation}
t_i(\mathbf{A}, \mathbf{X})= [(\mathbf{A}, \mathbf{X})]_{\mathcal{V}' \in \mathcal{V}},
\label{eq: subgraph sampling}
\end{equation}
where $\mathcal{V}'$ denotes a subset of $\mathcal{V}$, and $[\cdot]_{\mathcal{V}'}$ is a picking function that indexes the node feature and adjacency matrices of the subgraph with node set $\mathcal{V}'$.
Regarding the generation of $\mathcal{V}'$, several approaches have been proposed, such as uniform sampling \cite{zeng2020contrastive}, random walk-based sampling \cite{qiu2020gcc}, and top-k importance-based sampling \cite{jiao2020sub}.

Apart from the subgraph sampling, most of graph contrastive methods heavily rely on hybrid augmentations by combining the aforementioned strategies. For example, \methodHL{GRACE} \cite{zhu2020deep} applies the edge dropping and node feature masking, while \methodHL{MVGRL} \cite{hassani2020contrastive} adopts the graph diffusion and subgraph sampling to generate different contrastive views.

\begin{figure*}[tp]
	\centering
    \subfloat[In same-scale contrast, different views are generated with various graph augmentations based on the input graph at first. Then, the contrast can be placed on node or graph scales.
    ]{\includegraphics[width=3.2in]{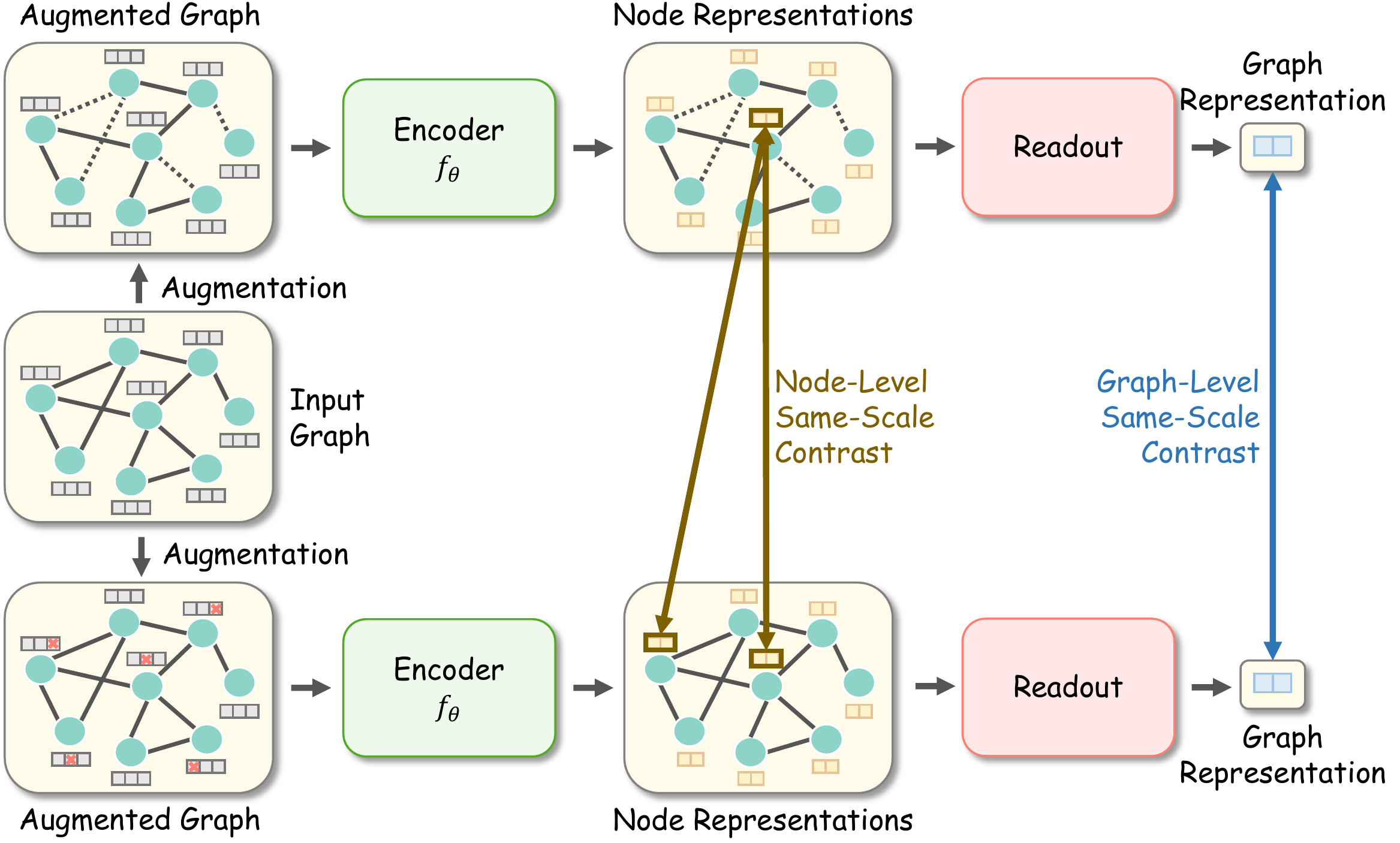}%
    \label{fig:con_ssc}
    }
    \hfill
	\subfloat[In cross-scale contrast, the augmented graph views are generated firstly. Then, the cross-scale contrast aims to discriminate patch- or context- level embeddings with global representations.
	]{\includegraphics[width=3.2in]{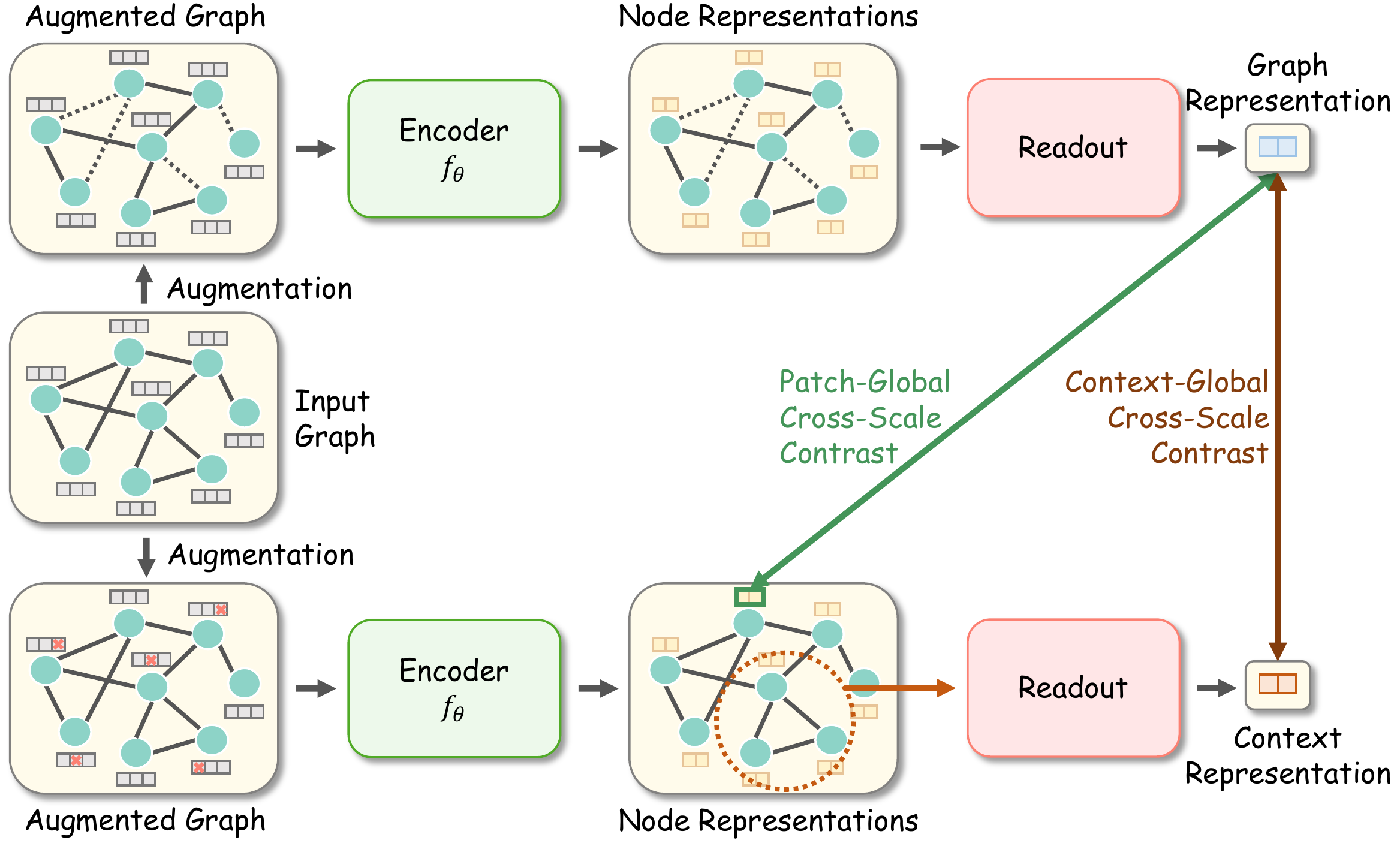}%
	\label{fig:con_csc}}
	\caption{Two categories of contrast-based graph SSL. %
	}
	\label{fig:contrast}
\vspace{-3mm}
\end{figure*}

\subsection{Graph Contrastive Learning}
As contrastive learning aims to maximize the MI between instances with similar semantic information, various pretext tasks can be constructed to enrich the supervision signals from such information.
Regarding the formulation of pretext decoder $p_{\phi}(\cdot)$ in Equation (\ref{eq: contrastive}), we classify existing works into two mainstreams: \textit{same-scale} and \textit{cross-scale} contrastive learning. The former branch of methods discriminates graph instances in an equal scale (\textit{e.g.}, node versus node), while the second type of methods places the contrasting across multiple granularities (\textit{e.g.}, node versus graph). Fig. \ref{fig:contrast} and Table \ref{tab:summary_contrast} provide the pipelines and summaries of contrast-based methods, respectively.

\subsubsection{Same-Scale Contrast}

According to the scale for contrast, we further divide the same-scale contrastive learning approaches into two sub-types: node-level and graph-level.

\paragraph{\textit{Node-Level Same-Scale Contrast}}
Early methods \cite{perozzi2014deepwalk, grover2016node2vec, hamilton2017inductive, tang2015line} under this category are mainly to learn node-level representations and built on the idea that nodes with similar contextual information should share the similar representations. In other word, these methods are trying to pull the representation of a node closer to its contextual neighborhood without relying on complex graph augmentations. %
We formulate them as below: % 
\begin{equation}
{\theta^{*}} = \mathop{\arg\min}\limits_{{\theta}} \frac{1}{\lvert \mathcal{V} \rvert} \sum_{v_i \in \mathcal{V}} {\mathcal{L}_{con}}\big(p\big([f_{\theta}(\mathbf{A}, \mathbf{X})]_{v_i}, [f_{\theta}(\mathbf{A}, \mathbf{X})]_{v_c}\big)\big),
\label{eq: C-SSC}
\end{equation}
where $v_c$ denotes the contextual node of $v_i$, for example, a neighboring node in a random walk starting from $v_i$. In those methods, the pretext discriminator (i.e., decoder) is typically the dot product and thus we omit its parameter in equation. Specifically, \methodHL{DeepWalk} \cite{perozzi2014deepwalk} introduces a random walk (RW)-based approach to extract the contextual information around a selected node in an unattributed graph. It maximizes the co-occurrence (i.e., MI measured by the binary classifier) of nodes within the same walk as in the Skip-Gram model \cite{mikolov2013befficient, mikolov2013distributed}. Similarly, \methodHL{node2vec} \cite{grover2016node2vec} adopts biased RWs to explore richer node contextual information and yields a better performance. \methodHL{GraphSAGE} \cite{hamilton2017inductive}, on the other hand, extends aforementioned two methods to attributed graphs, and proposes a novel GNN to calculate node embedding in an inductive manner, which applies RW as its internal sampling strategy as well. On heterogeneous graphs, \methodHL{SELAR} \cite{hwang2020self} samples meta-paths to capture the contextual information. It consists of a primary link prediction task and several meta-paths prediction auxiliary tasks to enforce nodes within the same meta-path to share closer semantic information. \looseness-1 %

Different from the aforementioned approaches, modern node-level same-scale contrastive methods are exploring richer underlying semantic information via various graph augmentations, instead of limiting on subgraph sampling:
\begin{equation}
{\theta^{*},\phi^{*}} = \mathop{\arg\min}\limits_{{\theta}, {\phi}}{\mathcal{L}_{con}}\Big( p_{\phi}\big( f_{\theta}( \tilde{\mathbf{A}}^{(1)},\tilde{\mathbf{X}}^{(1)}), f_{\theta}( \tilde{\mathbf{A}}^{(2)},\tilde{\mathbf{X}}^{(2)}) \big)\Big),
\label{eq: A-SSC node-level GRL}
\end{equation}
where $\tilde{\mathbf{A}}^{(1)}$ and $\tilde{\mathbf{A}}^{(2)}$ are two augmented graph adjacency matrices. Similarly, $\tilde{\mathbf{X}}^{(1)}$ and $\tilde{\mathbf{X}}^{(2)}$ are two node feature matrices under different augmentations. The discriminator $p_{\phi}(\cdot)$ in above equation can be parametric with $\Phi$ (\textit{e.g.}, bilinear transformation) or not (\textit{e.g.}, cosine similarity where $\Phi=\emptyset$). In those methods, most of them {deal with} attributed graphs: \methodHL{GRACE} \cite{zhu2020deep} adopts two graph augmentation strategies, namely node feature masking and edge dropping, to generate two contrastive views, which then pulls the representations of the same nodes closer between two graph views while pushing the rest of nodes away (i.e., intra- and inter-view negatives). Based on this framework, \methodHL{GCA} \cite{zhu2020graph} further introduces an adaptive augmentation for graph-structured data based on underlying graph properties, which results in a more competitive performance. Differently, \methodHL{GROC} \cite{jovanovic2021towards} proposes an adversarial augmentation on graph linkages to increase the robustness of learned node representations. %
Because of the success of SimCLR \cite{chen2020simple} in the visual domain, \methodHL{GraphCL(N)} \cite{hafidi2020graphcl} \footnote{The approaches proposed in \cite{hafidi2020graphcl} and \cite{you2020graph} have the same name ``GraphCL''. For distinction, we denote the node-level approach \cite{hafidi2020graphcl} as GraphCL(N) and the graph-level approach \cite{you2020graph} as GraphCL(G).} further extends this idea to graph-structured data, which relies on the node feature masking and edge modification to generate two contrastive views, and then the MI between two target nodes within different views is maximized. 
CGPN \cite{wan2021contrastive} introduces Poisson learning to node-level contrastive learning, which benefits node classification task under extremely limited labeled data.
On plain graphs, \methodHL{GCC} \cite{qiu2020gcc} utilizes RW as augmentations to extract the contextual information of a node, which then contrasts the representation of it with its counterparts by leveraging the contrastive framework of MoCo \cite{he2020momentum}. %
On the other hand, \methodHL{HeCo} \cite{wang2021self} is contrasting on heterogeneous graphs, where two contrastive views are generated from two perspectives, i.e., network schema and meta-path, while the encoder is trained by maximizing the MI between the embeddings of the same node in two views. %

Apart from those methods relying on carefully-crafted negative samples, approaches like \methodHL{BGRL} \cite{thakoor2021bootstrapped} propose to contrast on graph instances themselves and thus alleviate the reliance on deliberately designed negative sampling strategies. BGRL takes the advantage of knowledge distillation in BYOL \cite{grill2020bootstrap}, where a momentum-driven Siamese architecture has been introduced to guide the extraction of supervision signals. Specifically, it uses node feature masking and edge modification as augmentations, and the objective of BGRL is the same as in BYOL where the MI between node representations from online and target networks is maximized. \methodHL{SelfGNN} \cite{kefato2021self} adopts the same technique while the difference is that SelfGNN uses other graph augmentations, such as graph diffusion \cite{klicpera2019diffusion}, node feature split, standardization, and pasting. Apart from BYOL, Barlow Twins \cite{zbontar2021barlow} is another similar yet powerful method without using negative samples to prevent {the model from collapsing}. \methodHL{G-BT} \cite{bielak2021graph} extends the redundancy-reduction principle {for graph data analytics}, where the optimization objective is to minimize the dissimilarity between the identity and cross-correlation metrics generated via node embeddings of two augmented graph views. \methodHL{MERIT} \cite{Jin2021MultiScaleCS}, on the other hand, proposes to combine the advantages of Siamese knowledge distillation and conventional graph contrastive learning. It leverages a self-distillation framework in SimSiam \cite{chen2021exploring} while introducing extra node-level negatives to further exploit the underlying semantic information and enrich the supervision signals. %

\begin{table*}[thp]
% \scriptsize
\caption{Main characteristics of contrast-based graph SSL approaches. ``NSC'', ``GSC'', ``PGCC'' and ``CGCC'' mean node-level same-scale, graph-level same-scale, patch-global cross-scale, and context-global cross-scale contrast, respectively.}
\label{tab:summary_contrast}
% \vspace{-3mm}
\centering
\begin{adjustbox}{width=1.97\columnwidth,center}
\begin{tabular}{lcccccc}
\toprule
\thead{Approach} & \thead{Pretext Task \\ Category} & \thead{Downstream \\Task Level} & \thead{Training \\Scheme} & \thead{Data Type\\ of Graph} & \thead{Graph\\ Augmentation} & \thead{Objective\\ Function}\\ \bottomrule

DeepWalk \cite{perozzi2014deepwalk}   &  NSC   &  Node   &   URL   &  Plain & SS & SkipGram  \\ \EOL

node2vec \cite{grover2016node2vec}   &  NSC   &  Node   &   URL   &  Plain & SS & SkipGram  \\ \EOL

GraphSAGE \cite{hamilton2017inductive}   &  NSC   &  Node   &   URL   &  Attributed & SS & JSD  \\ \EOL

SELAR \cite{hwang2020self}   &  NSC   &  Node   &   JL   &  Heterogeneous & Meta-path sampling & JSD  \\ \EOL % 

LINE \cite{tang2015line}   &  NSC   &  Node   &   URL   &  Plain & SS & JSD  \\ \EOL

GRACE \cite{zhu2020deep}   &  NSC   &  Node   &   URL   &  Attributed & NFM+EM & InfoNCE  \\ \EOL

GROC \cite{jovanovic2021towards}   &  NSC   &  Node   &   URL   &  Attributed & NFM+Adversarial EM & InfoNCE  \\ \EOL

GCA \cite{zhu2020graph}   &  NSC   &  Node   &   URL   &  Attributed & Adaptive NFM+Adaptive EM & InfoNCE  \\ \EOL

GraphCL(N) \cite{hafidi2020graphcl}   &  NSC   &  Node   &   URL   &  Attributed & SS+NFS+EM & InfoNCE  \\ \EOL

CGPN \cite{wan2021contrastive} &  NSC   &  Node   &   JL   &  Attributed & None & InfoNCE  \\ \EOL

GCC \cite{qiu2020gcc}   &  NSC   &  Node/Graph   &   PF/URL   &  Plain & SS & InfoNCE  \\ \EOL

HeCo \cite{wang2021self}   &  NSC   &  Node   &   URL   &  Heterogeneous & NFM & InfoNCE  \\ \EOL

BGRL \cite{thakoor2021bootstrapped}   &  NSC   &  Node   &   URL   &  Attributed & NFM+EM & BYOL  \\ \EOL

SelfGNN \cite{kefato2021self}   &  NSC   &  Node   &   URL   &  Attributed & GD+Node attributive transformation & BYOL  \\ \EOL

G-BT \cite{bielak2021graph}   &  NSC   &  Node   &   URL   &  Attributed & NFM+EM & Barlow Twins \\ \EOL

MERIT \cite{Jin2021MultiScaleCS}   &  NSC   &  Node   &   URL   &  Attributed & SS+GD+NFM+EM & BYOL+InfoNCE  \\ \EOL

GraphCL(G) \cite{you2020graph}   &  GSC   &  Graph   &   PF/URL   &  Attributed & SS+NFM+EM & InfoNCE \\ \EOL

DACL \cite{verma2020towards}   &  GSC   &  Graph   &   URL   &  Attributed & Noise Mixing & InfoNCE \\ \EOL % ? -> URL by Yixin

AD-GCL \cite{suresh2021adversarial}   &  GSC   &  Graph  &   PF/URL   &  Attributed & Adversarail EM & InfoNCE \\ \EOL

JOAO \cite{you2021graph}   &  GSC   &  Graph   &   PF/URL   &  Attributed & Automated & InfoNCE \\ \EOL

CSSL \cite{zeng2020contrastive}   &  GSC   &  Graph   &   PF/JL/URL   &  Attributed & SS+Node insertion/deletion+EM & InfoNCE \\ \EOL

LCGNN \cite{ren2021label}   &  GSC   &  Graph   &   JL   &  Attributed & Arbitrary & InfoNCE \\ \EOL

IGSD \cite{zhang2020iterative}   &  GSC   &  Graph   &   JL/URL   &  Attributed & GD+EM & BYOL+InfoNCE \\ \EOL

%%%%%%%%%%%%%%%%%%%%%%%%%%%%

DGI \cite{velivckovic2018deep}   &  PGCC   &  Node  &   URL   &  Attributed & None & JSD \\ \EOL

GIC \cite{mavromatis2020graph}   &  PGCC   &  Node  &   URL   &  Attributed & Arbitrary & JSD \\ \EOL

HDGI \cite{ren2020heterogeneous}   &  PGCC   &  Node  &   URL   &  Heterogeneous & None & JSD \\ \EOL

ConCH \cite{li2020leveraging}   &  PGCC   &  Node  &   JL   &  Attributed & None & JSD \\ \EOL

DMGI \cite{park2020unsupervised}   &  PGCC   &  Node  &   JL/URL   &  Heterogeneous & None & JSD \\ \EOL

EGI \cite{zhu2020transfer}   &  PGCC   &  Node  &   PF/JL   &  Attributed & SS & JSD \\ \EOL %

STDGI \cite{opolka2019spatio}   &  PGCC   &  Node  &   URL   &  Spatial-temporal & Node feature shuffling & JSD \\ \EOL

MVGRL \cite{hassani2020contrastive}   &  PGCC   &  Node/Graph  &   URL   &  Attributed & GD+SS & JSD \\ \EOL

SUBG-CON \cite{jiao2020sub}   &  PGCC   &  Node  &   URL   &  Attributed & SS+Node representation shuffling & Triplet \\ \EOL

SLiCE \cite{wang2020self}   &  PGCC   &  Edge  &   JL   &  Heterogeneous & None & JSD \\ \EOL

InfoGraph \cite{sun2019infograph}   &  PGCC   &  Graph  &   JL/URL   &  Attributed & None & JSD \\ \EOL

Robinson et al. \cite{robinson2020contrastive}   &  PGCC   &  Graph  &   URL   &  Attributed & Arbitrary & JSD \\ \EOL

BiGI \cite{cao2020bipartite}   &  CGCC   &  Graph  &   URL   &  Heterogeneous & SS & JSD \\ \EOL

HTC \cite{wang2021learning}   &  CGCC   &  Graph  &   JL   &  Attributed & NFS & JSD \\ \EOL

MICRO-Graph \cite{zhang2020motif}   &  CGCC   &  Graph  &   URL   &  Attributed & SS & InfoNCE \\ \EOL %JL?

SUGAR \cite{sun2021sugar}   &  CGCC   &  Graph  &   JL   &  Attributed & SS & JSD \\ \toprule

\end{tabular}
\end{adjustbox}
\vspace{-3mm}
\end{table*}

\paragraph{\textit{Graph-Level Same-Scale Contrast}}
For graph-level representation learning under same-scale contrasting, the discrimination is typically placed on graph representations:

\begin{equation}
{\theta^{*},\phi^{*}} = \mathop{\arg\min}\limits_{{\theta}, {\phi}}{\mathcal{L}_{con}}\Big( p_{\phi}\big( \tilde{\mathbf{g}}^{(1)}, \tilde{\mathbf{g}}^{(2)} \big)\Big),
\label{eq: A-SSC graph-level GRL}
\end{equation}
where $\tilde{\mathbf{g}}^{(i)}=\mathcal{R}\big( f_{\theta}( \tilde{\mathbf{A}}^{(i)}, \tilde{\mathbf{X}}^{(i)} ) \big)$ denotes the representation of augmented graph $\tilde{\mathcal{G}}^{(i)}$, and $\mathcal{R}(\cdot)$ is a readout function to generate the graph-level embedding based on node representations.
Methods under Equation (\ref{eq: A-SSC graph-level GRL}) may share similar augmentations and backbone contrastive frameworks with {the} aforementioned node-level approaches. 
For example, \methodHL{GraphCL(G)} \cite{you2020graph} adopts SimCLR \cite{chen2020simple} to form its contrastive pipeline which pulls the graph-level representations of two views closer. %
Similarly, \methodHL{DACL} \cite{verma2020towards} is also built on SimCLR but it designs a general yet effective augmentation strategy, namely mixup-based data interpolation.
\methodHL{AD-GCL} \cite{suresh2021adversarial} proposes an adversarial edge dropping mechanism as augmentations to reduce the amount of redundant information taken by encoders.
\methodHL{JOAO} \cite{you2021graph} % 
proposes the concept of joint augmentation optimization, where a bi-level optimization problem is formulated by jointly optimizing the augmentation selection together with the contrastive objectives. %
Similar to GCC \cite{qiu2020gcc}, \methodHL{CSSL} \cite{zeng2020contrastive} is built on MoCo \cite{he2020momentum} but it contrasts graph-level embeddings. % 
A similar design can also be found in \methodHL{LCGNN} \cite{ren2021label}. On the other hand, regarding to the knowledge-distillation, \methodHL{IGSD} \cite{zhang2020iterative} is leveraging the concept of BYOL \cite{grill2020bootstrap} and similar to MERIT \cite{Jin2021MultiScaleCS}. %

\subsubsection{Cross-Scale Contrast}
Different from contrasting graph instances in an equivalent scale, this branch of methods places the discrimination across various graph topologies (\textit{e.g.}, node versus graph). We further build two sub-classes under this category, namely patch-global and context-global contrast.

\paragraph{\textit{Patch-Global Cross-Scale Contrast}}
For node-level representation learning, we define this contrast as below:

\begin{equation}
\begin{aligned}
&{\theta^{*},\phi^{*}} = \\ &\mathop{\arg\min}\limits_{{\theta}, {\phi}} \frac{1}{\lvert \mathcal{V} \rvert} \sum_{v_i \in \mathcal{V}} {\mathcal{L}_{con}}\bigg(p_{\phi}\Big([f_{\theta}(\mathbf{A}, \mathbf{X})]_{v_i}, \mathcal{R}\big(f_{\theta}(\mathbf{A}, \mathbf{X})\big)\Big)\bigg),
\label{eq: C-C patch-global}
\end{aligned}
\end{equation}
where $\mathcal{R}(\cdot)$ denotes the readout function as we mentioned in previous subsection. 
Under this category, \methodHL{DGI} \cite{velivckovic2018deep} is the first method that proposes to contrast node-level embeddings with the graph-level representation, which aims to maximize the MI between such two representations from different scales to assist the graph encoder to learn both localized and global semantic information. 
Based on this idea, \methodHL{GIC} \cite{mavromatis2020graph} first clusters nodes within a graph based on their embeddings, and then pulls nodes closer to their corresponding cluster summaries, which is optimized with a DGI objective simultaneously. 
Apart from attributed graphs, some works on heterogeneous graphs are based on the similar schema: 
\methodHL{HDGI} \cite{ren2020heterogeneous} can be regarded as a version of DGI on heterogeneous graphs, where the difference is that the final node embeddings of a graph are calculated by aggregating node representations under different meta-paths. % 
Similarly, \methodHL{ConCH} \cite{li2020leveraging} shares the same objective with DGI and aggregates meta-path-based node representations to calculate node embeddings of a heterogeneous graph.
Differently, \methodHL{DMGI} \cite{park2020unsupervised} considers a multiplex graph as the combination of several attributed graphs. For each of them, given a selected target node and its associated relation type, the relation-specific node embedding is firstly calculated. The MI between the graph-level representation and such an node embedding is maximized as in DGI. %
\methodHL{EGI} \cite{zhu2020transfer} extracts high-level transferable graph knowledge by enforcing node features to be structure-respecting and then maximizing the MI between the embedding of a node and its surrounding ego-graphs. 
On spatial-temporal graphs, \methodHL{STDGI} \cite{opolka2019spatio} maximizes the agreement between the node representations at timestep $t$ with the raw node features at $t+1$ to guide the graph encoder to capture rich semantic information to predict future node features. %

Note that aforementioned methods are not explicitly using any graph augmentations. %
For patch-global contrastive approaches based on augmentations, we reformulate Equation (\ref{eq: C-C patch-global}) as follows:

\begin{equation}
{\theta^{*},\phi^{*}} = \mathop{\arg\min}\limits_{{\theta}, {\phi}} \frac{1}{\lvert \mathcal{V} \rvert} \sum_{v_i \in \mathcal{V}} {\mathcal{L}_{con}}\Big( p_{\phi}\big(\tilde{\mathbf{h}}_{i}^{(1)}, \tilde{\mathbf{g}}^{(2)}\big)\Big),
\label{eq: C-C patch-global augmentation-based}
\end{equation}

\noindent where $\tilde{\mathbf{h}}_{i}^{(1)}=[f_{\theta}(\tilde{\mathbf{A}}^{(1)}, \tilde{\mathbf{X}}^{(1)})]_{v_i}$ is the representation of node $v_i$ in augmented view 1, and $\tilde{\mathbf{g}}^{(2)}=\mathcal{R}\big(f_{\theta}(\tilde{\mathbf{A}}^{(2)}, \tilde{\mathbf{X}}^{(2)})\big)$ denotes the representation of differently augmented view 2. Under the umbrella of this definition, \methodHL{MVGRL} \cite{hassani2020contrastive} first generates two graph views via graph diffusion \cite{klicpera2019diffusion} and subgraph sampling. Then, it enriches the localized and global supervision signals by maximizing the MI between the node embeddings in a view and the graph-level representation of another view. \methodHL{SUBG-CON} \cite{jiao2020sub}, on the other hand, inherits the objective of MVGRL while it adopts different graph augmentations. Specifically, it first extracts the top-$k$ most informative neighbors of a central node from a large-scale input graph. Then, the encoded node representations are further shuffled to increase the difficulty of pretext task. On heterogeneous graphs, \methodHL{SLiCE} \cite{wang2020self} pulls nodes closer to their closest contextual graphs, instead of explicitly contrasting nodes with the entire graph. In addition, SLiCE enriches the localized information of node embeddings via a contextual translation mechanism.%

For graph-level representation learning based on patch-global contrast, we can formulate it by using Equation (\ref{eq: C-C patch-global}). \methodHL{InfoGraph} \cite{sun2019infograph} shares a similar schema with DGI \cite{velivckovic2018deep}. %
It contrasts the graph representation directly with node embeddings to discriminate whether a node belongs to the given graph. %
To further boost contrastive methods like InfoGraph, \methodHL{Robinson et al.} \cite{robinson2020contrastive} propose a general yet effective hard negative sampling strategy to make the underlying pretext task more challenging to solved.

\paragraph{\textit{Context-Global Cross-Scale Contrast}}

Another popular design under the category of cross-scale graph contrastive learning is context-global contrast, which is defined below:

\begin{equation}
{\theta^{*},\phi^{*}} = \mathop{\arg\min}\limits_{{\theta}, {\phi}} \frac{1}{\lvert \mathcal{S} \rvert} \sum_{ {s} \in \mathcal{S} } {\mathcal{L}_{con}}\Big( p_{\phi}\big(\tilde{\mathbf{h}}_{{s}}, \tilde{\mathbf{g}}\big)\Big),
\label{eq: C-C context-global}
\end{equation}
where $\mathcal{S}$ denotes a set of contextual subgraphs in an augmented input graph $\tilde{\mathcal{G}}$, where augmentations are typically based on graph sampling under this category. In above formula, ${\tilde{\mathbf{h}}_{s}}$ is the representation of augmented contextual subgraph $s$, and $\tilde{\mathbf{g}}$ represents the graph-level representation over all subgraphs in $\mathcal{S}$. Specifically, we let ${\tilde{\mathbf{h}}_{s}=\mathcal{R}\big( [ f_{\theta}(\tilde{\mathbf{A}}, \tilde{\mathbf{X}}) ]_{v_i \in s} \big)}$, and $\tilde{\mathbf{g}}=\mathcal{R}\big( f_{\theta}(\tilde{\mathbf{A}},\tilde{\mathbf{X}}) \big)$. However, for some methods, such as \cite{cao2020bipartite} and \cite{wang2021learning}, the graph-level representation is calculated on the original input graph, where $\tilde{\mathbf{g}}=\mathcal{R}\big( f_{\theta}(\mathbf{A},\mathbf{X}) \big)$. 
Among them, \methodHL{BiGI} \cite{cao2020bipartite} is a node-level representation learning approach on bipartite graphs, inheriting the contrasting schema of DGI \cite{velivckovic2018deep}. Specifically, it first calculates graph-level representation of the input graph by aggregating two types of node embeddings. Then, it samples the original graph, and then calculates the local contextual representation of a target edge between two nodes. The optimization objective of BiGI is to maximize the MI between such a local contextual and global representations, where the trained graph encoder can then be used in various edge-level downstream tasks. 
Aiming to learn graph-level embedding, \methodHL{HTC} \cite{wang2021learning} maximizes the MI between full-graph representation and the contextual embedding which is the aggregation of sampled subgraphs. % 
Similar to but different from HTC, \methodHL{MICRO-Graph} \cite{zhang2020motif} proposes a different yet novel motif learning-based sampling as the implicit augmentation to generate several semantically-informative subgraphs, where the embedding of each subgraph is pulled closer to the representation of entire graph. %
Considering a scenario that the graph-level representation is based on the augmented input graph, as the default setting shown in Equation (\ref{eq: C-C context-global}), \methodHL{SUGAR} \cite{sun2021sugar} first samples $n$ subgraphs from the given graph, and then proposes a reinforcement learning-based top-$k$ sampling strategy to select the $n'$ informative subgraphs among the candidate set with size $n$. Finally, the contrast of SUGAR is established between the subgraph embedding and the representation of sketched graph, i.e., the generated graph by combining these $n'$ subgraphs.

\vspace{-3mm}
\subsection{Mutual Information Estimation}
Most of contrast-based methods rely on the MI estimation between two or more instances. Specifically, the representations of a pair of instances sampled from the positive pool are being pulled closer while the counterparts from negative sets are pushed away. Given a pair of instances $(x_i, x_j)$, we let $(\mathbf{h}_i, \mathbf{h}_j)$ to denote their representations. Thus, the MI between $(i, j)$ is given by \cite{tschannen2019mutual}:

\begin{align}
\mathcal{MI}(\mathbf{h}_i, \mathbf{h}_j)&=KL\big( P(\mathbf{h}_i, \mathbf{h}_j) || P(\mathbf{h}_i)P(\mathbf{h}_j) \big) \notag \\
&=\mathbb{E}_{P(\mathbf{h}_i, \mathbf{h}_j)}\Big[ \operatorname{log} \frac{P(\mathbf{h}_i, \mathbf{h}_j)}{P(\mathbf{h}_i)P(\mathbf{h}_j)} \Big],
\label{eq: MI general form}
\end{align}
where $KL(\cdot)$ denotes the Kullback-Leibler divergence, and the end goal is to train the encoder to be discriminative between a pair of instances from the joint density ${P}(\mathbf{h}_i, \mathbf{h}_j)$ and negatives from marginal densities ${P}(\mathbf{h}_i)$ and ${P}(\mathbf{h}_j)$. In this subsection, we define two common forms of lower bound and three specific forms of non-bound MI estimators derived from Equation (\ref{eq: MI general form}).

\subsubsection{Jensen-Shannon Estimator}
Although Donsker-Varadhan representation provides a tight lower bound of KL divergence \cite{xie2021self}, Jensen-Shannon divergence (JSD) is more common on graph contrastive learning, which provides a lower bound and more efficient estimation on MI.
{We define the contrastive loss based on it as follows:}

\begin{equation}
\begin{aligned}
&{\mathcal{L}_{con}\Big( p_{\phi}\big(\mathbf{h}_{i}, \mathbf{h}_{j}\big)\Big)}={-}\mathcal{MI}_{JSD}(\mathbf{h}_i, \mathbf{h}_j) \\
&={\mathbb{E}_{\mathcal{P} \times \widetilde{\mathcal{P}}}\Big[\operatorname{log}\big( 1-p_{\phi}(\mathbf{h}_i, \mathbf{h'}_j)\big)\Big] - \mathbb{E}_{\mathcal{P}}\Big[ \operatorname{log}\big(p_{\phi}(\mathbf{h}_i, \mathbf{h}_j)\big)\Big]}.
\end{aligned}
\label{eq: JS estimator}
\end{equation}
{In above equation, $\mathbf{h}_{i}$ and $\mathbf{h}_{j}$ are sampled from the same distribution $\mathcal{P}$, and $\mathbf{h'}_{j}$ is sampled from a different distribution $\widetilde{\mathcal{P}}$. For the discriminator $p_{\phi}(\cdot)$, it can be taken from various forms,} where a bilinear transformation {\cite{oord2018representation}} is typically adopted, i.e., ${p_{\phi}(\mathbf{h}_i, \mathbf{h}_j)=\mathbf{h}_i^{T}\mathbf{\Phi}\mathbf{h}_j}$, such as in \cite{velivckovic2018deep, hassani2020contrastive, sun2021sugar}. Specifically, by letting ${p_{\phi}(\mathbf{h}_i, \mathbf{h}_j)=\operatorname{sigmoid}\big( p'_{\phi}(\mathbf{h}_i, \mathbf{h}_j) \big)}$, Equation (\ref{eq: JS estimator}) can be presented in another form as in InfoGraph \cite{sun2019infograph}.

\subsubsection{Noise-Contrastive Estimator}
Similar to JSD, noise-contrastive estimator (a.k.a. InfoNCE) provides a lower bound MI estimation that naturally consists of a positive and $N$ negative pairs \cite{xie2021self}. {An InfoNCE-based contrasitve loss is defined as follows:}
\begin{equation}
\begin{aligned}
&{\mathcal{L}_{con}\Big(p_{\phi}\big(\mathbf{h}_{i}, \mathbf{h}_{j}\big)\Big)}={-}\mathcal{MI}_{NCE}(\mathbf{h}_i, \mathbf{h}_j) \\
&={-}\mathbb{E}_{\mathcal{P}\times \widetilde{\mathcal{P}}^N}\Big[\operatorname{log}\frac{e^{p_{\phi}(\mathbf{h}_i,\mathbf{h}_j)}}{e^{p_{\phi}(\mathbf{h}_i,\mathbf{h}_j)} + \sum_{n \in N} e^{p_{\phi}(\mathbf{h}_i,\mathbf{h}'_n)}} \Big],
\end{aligned}
\label{eq: InfoNCE}
\end{equation}
where the discriminator ${p_{\phi}(\cdot)}$ can be the dot product with a temperature parameter $\tau$, i.e., ${p(\mathbf{h}_i,\mathbf{h}_j)=\mathbf{h}_i^{T}\mathbf{h}_j / \tau}$, such as in GRACE \cite{zhu2020deep} and GCC \cite{qiu2020gcc}.

\subsubsection{Triplet Loss}
Apart from aforementioned two lower bound MI estimators, a triplet margin loss can also be adopted to estimate the MI between data instances. However, minimizing this loss can not guarantee that the MI is being maximized because it cannot represent the lower bound of MI. Formally, Jiao et al. \cite{jiao2020sub} define this loss function as follows:

\begin{equation}
{\mathcal{L}_{con}\Big(p\big(\mathbf{h}_{i}, \mathbf{h}_{j}\big)\Big)}=\mathbb{E}_{\mathcal{P}\times\widetilde{\mathcal{P}}}\bigg[ \operatorname{max}\Big[p_{\phi}(\mathbf{h}_i,\mathbf{h}_j) - p_{\phi}(\mathbf{h}_i,\mathbf{h'}_j)+\epsilon, 0\Big]\bigg],
\label{eq: triplet loss}
\end{equation}
where $\epsilon$ is a margin value, and the discriminator ${p(\mathbf{h}_i, \mathbf{h}_j)=1/1+e^{(-\mathbf{h}^{T}_i\mathbf{h}_j)}}$. 

\subsubsection{BYOL Loss}
For the methods inspired by BYOL \cite{grill2020bootstrap} and not relying on negative samples, such as BGRL \cite{grill2020bootstrap}, their objective functions can also be interpreted as a non-bound MI estimator. 

{Given $\mathbf{h}_{i}, \mathbf{h}_{j} \sim \mathcal{P}$, we define this loss as in below:}

\begin{equation}
{\mathcal{L}_{con}\Big(p\big(\mathbf{h}_{i}, \mathbf{h}_{j}\big)\Big)}{=\mathbb{E}_{\mathcal{P} \times \mathcal{P}}\Big[2 - 2 \cdot \frac{\big[p_{\psi}(\mathbf{h}_i)\big]^T\mathbf{h}_j}{ \left \| p_{\psi}(\mathbf{h}_i) \right\| \left \| \mathbf{h}_j \right\|}\Big]},
\label{eq: byol}
\end{equation}
where $p_{\psi}$ denotes an online predictor parameterized by $\mathbf{\psi}$ in Siamese networks, which prevents {the model from collapsing} with other mechanisms such as momentum encoders, stop gradient, etc. {In particular, the pretext decoder in this case denotes the mean square error between two instances, which has been expanded in the above equation.}

\subsubsection{Barlow Twins Loss}
Similar to BYOL, this objective alleviates the reliance on negative samples but much simpler in implementation, which is motivated by the redundancy-reduction principle. Specifically, given the representations of two views $\mathbf{H^{(1)}}$ and $\mathbf{H^{(2)}}$ for a batch of data instances sampled from a distribution $\mathcal{P}$, we define this loss function as below \cite{zbontar2021barlow}:

\begin{equation}
\begin{aligned}
{\mathcal{L}_{con}\big(\mathbf{H^{(1)}}, \mathbf{H^{(2)}}\big)}=&\mathbb{E}_{{\mathcal{B}} \sim \mathcal{P}^{{|\mathcal{B}|}}}\bigg[\sum_{a}(1-\frac{\sum_{i \in {\mathcal{B}}}\mathbf{H}_{ia}^{(1)}\mathbf{H}_{ia}^{(2)}}{\left \| \mathbf{H}_{ia}^{(1)} \right\| \left \|\mathbf{H}_{ia}^{(2)}\right\|})^2 \\ &+ \lambda \sum_{a}\sum_{b \neq a}\Big(\frac{\sum_{i \in {\mathcal{B}}}\mathbf{H}_{ia}^{(1)}\mathbf{H}_{ib}^{(2)}}{\left \| \mathbf{H}_{ia}^{(1)} \right\| \left \|\mathbf{H}_{ib}^{(2)}\right\|}\Big)^2\bigg],
\end{aligned}
\label{eq: barlow twins}
\end{equation}
where $a$ and $b$ index the dimension of a representation vector, and $i$ indexes the samples within a batch ${\mathcal{B}}$.

\noindent {\textbf{Discussion.} In general, same-scale methods usually follow the two-stream contrastive learning frameworks (e.g., SimCLR \cite{chen2020simple} and BYOL \cite{grill2020bootstrap}), where InfoNCE and BYOL losses are widely used; In contrast, cross-scale methods often derive from DGI \cite{velivckovic2018deep}, hence they prefer the JSD loss.
}

\vspace{-3mm}
\section{Hybrid Methods} \label{sec:hybrid}

Compared to the aforementioned methods that only utilize a single pretext task to train models, hybrid methods adopt multiple pretext tasks to better leverage the advantages of various types of supervision signals. %
The hybrid methods integrate various pretext tasks together in a \textit{multi-task learning} fashion, where the objective function is the weighted sum of two or more self-supervised objectives. The formulation of hybrid graph SSL methods is:

\begin{equation}
\begin{aligned}
{\theta^{*},\phi^{*}} = \mathop{\arg\min}\limits_{{\theta}, {\phi}} \sum\limits_{i=1}^N \alpha_i\mathcal{L}_{ssl_i}\left( f_{\theta}, p_{\phi_i}, \mathcal{D}_i \right),
\end{aligned}
\label{eq: hybrid}
\end{equation}

\noindent where $N$ is the number of pretext tasks, $\alpha_i$, $\mathcal{L}_{ssl_i}$, $p_{\phi_1}$ and $\mathcal{D}_i$ are the trade-off weight, loss function, pretext decoder and data distribution of the $i$-th pretext task, respectively. 
 
\begin{table}[t]
\caption{Main characteristics of hybrid graph SSL approaches. ``M. et al.'' and ``Hetero.''  are the abbreviations for ``Manessi  et  al.'' and ``Heterogeneous'', respectively. The abbreviations for pretext tasks categories please refer to Table \ref{tab:summary_generation}, \ref{tab:summary_property} and \ref{tab:summary_contrast}.}
\label{tab:summary_hybrid}
\vspace{-3mm}
\centering
\begin{adjustbox}{width=1\columnwidth,center}
\begin{tabular}{lcccc}
\toprule
\thead{Approach} & \thead{Pretext Task \\ Categories} & \thead{Downstream \\Task Level} & \thead{Training \\Scheme} & \thead{Data Type\\ of Graph} \\ \bottomrule
GPT-GNN \cite{hu2020gpt}   &  FG/SG   &  Node/Link       &   PF   &  Hetero.  \\ \EOL
Graph-Bert \cite{zhang2020graph}   &  FG/SG   &  Node       &   PF   &  Attributed  \\ \EOL
PT-DGNN \cite{zhang2021pre}   &  FG/SG   &  Link       &   PF   &  Dynamic  \\
\EOL
M. et al. \cite{manessi2020graph}   &  \tabincell{c}{FG/FG/FG}   &  Node       &   JL   &  Attributed  \\\EOL
GMI \cite{peng2020graph}   &  SG/NSC   &  Node/Link       &   URL   &  Attributed  \\\EOL
CG$^3$ \cite{wan2020contrastive}   &  SG/NSC   &  Node       &   JL   &  Attributed  \\\EOL
MVMI-FT \cite{fan2021maximizing} & SG/PGCC & Node & URL & Attributed \\ \EOL
GraphLoG \cite{xu2021self} & \tabincell{c}{NSC/GSC/ \\ CGCC} & Graph & PF & Attributed \\\EOL
HDMI \cite{jing2021hdmi} & NSC/PGCC & Node & URL & Multiplex \\ \EOL
G-Zoom \cite{zheng2021gzoom} & \tabincell{c}{NSC/NSC/ \\ GSC} & Node & URL & Attributed \\ \EOL
LnL-GNN \cite{roy2021node} & NSC/NSC & Node & JL & Attributed \\\EOL
Hu et al. \cite{hu2019pre}   &  \tabincell{c}{SG/APC/ \\ APC}   & \tabincell{c}{Node/Link/ \\ Graph}       &   PF    &  Attributed   \\\EOL
GROVER \cite{rong2020self}   & \tabincell{c}{APC/APC}   & \tabincell{c}{Node/Link/ \\ Graph}       &   PF    &  Attributed   \\\EOL
Kou et al. \cite{kou2021self} &  \tabincell{c}{FG/SG/ \\ APC}   & Node       &   JL    &  Attributed   \\\toprule
\end{tabular}
\end{adjustbox}
\vspace{-5mm}
\end{table}

A common idea of hybrid graph SSL is to combine different generation-based tasks together. \methodHL{GPT-GNN} \cite{hu2020gpt} integrates feature and structure generation into a pre-training framework for GNNs. Specifically, for each sampled input graph, it first randomly masks a certain amount of edges and nodes. Then, two generation tasks are used to train the encoder simultaneously: \methodHL{Attribute Generation} that rebuilds the masked features with MSE loss, and \methodHL{Edge Generation} that predicts the masked edges with a contrastive loss. 
\methodHL{Graph-Bert} \cite{zhang2020graph} combines attributive and structural pretext tasks %
to pre-train a graph transformer model. Concretely, \methodHL{Node Raw Attribute Reconstruction} reconstructs the raw features from the node's embedding, while \methodHL{Graph Structure Recovery} aims to recover the graph diffusion value between two nodes with a cosine similarity decoder. %
\methodHL{PT-DGNN} \cite{zhang2021pre} extends the idea of combining attributive and structural generation to pre-train GNNs for dynamic graphs. %
Besides, \methodHL{Manessi et al.} \cite{manessi2020graph} propose to train GNNs with three types of feature generation tasks.%,

Another idea is to integrate generative and contrastive pretext tasks together. 
{\methodHL{GMI} \cite{peng2020graph} adopts a joint learning objective for graph representation learning. In GMI, the contrastive learning target (i.e., feature MI) is to maximize the agreement between node embeddings and neighbors' features with a JSD estimator, and the generative target (i.e., edge MI) is to minimize the reconstruction error of the adjacency matrix with a BCE loss. }
\methodHL{CG$^3$} \cite{wan2020contrastive} considers contrastive and generative SSL jointly for semi-supervised node classification problem. In CG$^3$, two parallel encoders (GCN and HGCN) are established to provide local and global views for graphs. In \methodHL{contrastive learning}, an Info-NCE contrastive loss is used to maximize the MI between the node embeddings from two views. In \methodHL{generative learning}, a generative decoder is used to rebuild the topological structure from the concatenation of two {views'} embeddings. %
\methodHL{MVMI-FT} \cite{fan2021maximizing} presents a cross-scale contrastive learning framework that learns node representation from different views, and also uses a graph reconstruction module to learn the cross-view sharing information. 

Since different types of contrasts can provide supervision signals from different views, some approaches integrate multiple contrast-based tasks together. 
\methodHL{GraphLoG} \cite{xu2021self} consists of three contrastive objectives: the subgraph versus subgraph, graph versus graph, and graph versus contextual. The InfoNCE loss serves as the MI estimator for three types of contrasts. %
\methodHL{HDMI} \cite{jing2021hdmi} mixes both same-scale and cross-scale contrastive learning, which dissects a given multiplex network into multiple attributed graphs. For each of them, HDMI proposes three different objectives to maximize the MI between raw node features, node embeddings, and graph-level representations. 
\methodHL{G-Zoom} \cite{zheng2021gzoom} uses same-scale contrasts in three scales to learns representations, which extracts valuable clues from multiple perspectives.
\methodHL{LnL-GNN} \cite{roy2021node} leverages a bi-level MI maximization to learn from local and non-local neighborhoods obtained by community detection and feature-based clustering respectively.% 

Different auxiliary property-based tasks can also be integrated into a hybrid method. \methodHL{Hu et al.} \cite{hu2019pre} present to pre-train GNNs with multiple tasks simultaneously to capture transferable generic graph structures, including \methodHL{Denoising Link Reconstruction}, \methodHL{Centrality Score Ranking}, and \methodHL{Cluster Preserving}. 
In \methodHL{GROVER} \cite{rong2020self}, the authors pre-train the GNN Transformer model with auxiliary property classification tasks in node level (\methodHL{Contextual Property Prediction}) and graph level (\methodHL{Motif Prediction}) simultaneously. 
\methodHL{Kou et al.} \cite{kou2021self} mix structure generation, feature generation, and auxiliary property classification tasks into a clustering model. %

\vspace{-3mm}
\section{{Empirical Study}} \label{sec:resource}

{In this section, we summarize essential resources for empirical study of graph SSL. Specifically, we conduct an experimental comparison of the representative methods on two commonly used downstream tasks on graph learning, i.e., node classification and graph classification. We also collect useful resources for empirical research, including benchmark datasets and open-source implementations.}

\noindent {\textbf{Performance Comparison of Node Classification.} 
We consider two learning settings for node classification, i.e., semi-supervised transductive learning and supervised inductive learning. For transductive learning, we consider three citation network datasets, including Cora, Citeseer and Pubmed \cite{sen2008collective}, for performance evaluation. The standard split of train/valid/test often follows \cite{gcn_kipf2017semi}, where 20 nodes per class are used for training, 500/1000 nodes are used for validation/testing. For inductive learning, we use PPI dataset \cite{hamilton2017inductive} to evaluate the performance. Following \cite{hamilton2017inductive}, 20 graphs are employed to train the model, while 2 graphs are used to validate and 2 graphs are used to test. In both setting, the performance is measured by classification accuracy.}

{We compare the performance of two groups of graph SSL methods. In URL group, the encoder is purely trained by SSL pretext tasks, and the learned representations are directly fed into classification decoders. In PF/JL group, the training labels are accessible for encoders' learning. We consider two conventional (semi-) supervised classification methods (i.e., GCN \cite{gcn_kipf2017semi} and GAT \cite{gat_ve2018graph}) as baselines. }

The results of performance comparison are illustrated in Table \ref{tab:result_node_classification}. According to the results, we have the following observations \hlt{and analysis}: 
(1) Early random walk-based contrastive methods (e.g., DeepWalk and GraphSAGE) and autoencoder-based generative methods (e.g., GAE and SIG-VAE) perform worse than \hlt{the majority of graph SSL methods. The possible reason is that they train encoders with simple unsupervised learning targets instead of well-designed self-supervised pretext tasks, hence failing to fully leverage the original data to acquire supervision signals. For example, DeepWalk only maximizes the MI among nodes within a random walk, ignoring the global structural information of graphs. 
(2) The methods employing advanced contrastive objectives from visual contrastive learning (e.g., BGRL which uses BYOL loss \cite{grill2020bootstrap} and G-BT which uses Barlow Twins loss \cite{zbontar2021barlow}) do not show a superior performance like their prototypes performing on visual data. Such an observation indicates that directly borrowing self-supervised objectives from other domains does not always bring enhancement. 
(3) Some representative contrast-based methods (e.g., MVGRL, MERIT, and SubG-Con) perform better than the generalization-based and auxiliary property-based methods, which reflects the effectiveness of contrastive pretext tasks and the potential room for improvement of other methods.} 
(4) The hybrid methods have competitive performance and some of them even outperform the supervised baselines. \hlt{The outperformance suggests that integrating multiple pretext tasks can provide supervision signals from diverse perspectives, which brings significant performance gain. For instance, G-Zoom \cite{zheng2021gzoom} achieves excellent results by combining contrastive pretext tasks in three different levels.} 
(5) The performance of methods in PF/JL groups is generally better than that in URL groups, \hlt{which demonstrates that the accessibility of label information leads to further improvement for graph SSL.}

\noindent {\textbf{More resources.} For the evaluation and performance comparison of \textit{graph classification}, we please readers refer to Appendix \ref{appendix:evaluation}. We also collect widely applied benchmark datasets and divide them into four groups. The description and statistics of the selected \textit{benchmark datasets} are detailed in Appendix \ref{appendix:datasets}. Besides, we provide a collection of the \textit{open-source implementations} of the surveyed works in Appendix \ref{appendix:implementations}, which can facilitate the reproduction, improvement, and baseline experiments in further research.}

\begin{table}[t!]
 	\caption{A summary of experimental results for node classification in four benchmark datasets.}
	\label{tab:result_node_classification}
	\centering
	\vspace{-3mm}
	\begin{adjustbox}{width=1\columnwidth,center}
	\begin{tabular}{l l l c c c c}
		\toprule
		
		\multicolumn{1}{l}{Group} & Approach & Category & Cora & Citeseer & Pubmed & PPI \\ 
		\bottomrule
% 		\hline
		\multicolumn{1}{l|}{\multirow{2}{*}
		{\tabincell{l}{Base- \\ lines}}} & GCN \cite{gcn_kipf2017semi} & - & 81.5 & 70.3 & 79.0 & - \\ \cline{2-7}  
		
		\multicolumn{1}{l|}{} & GAT \cite{gat_ve2018graph} & - & 83.0 & 72.5 & 79.0  & 97.3 \\  \hline

		\multicolumn{1}{l|}{\multirow{17}{*}
		{URL}} &  GAE  \cite{kipf2016variational} & SG & 80.9 & 66.7 & 77.1 & - \\  \cline{2-7}  % 
		
		\multicolumn{1}{l|}{} & SIG-VAE  \cite{hasanzadeh2019semi} & SG  & 79.7 & 70.4 & 79.3 & - \\  \cline{2-7}  % from original paper
		
		\multicolumn{1}{l|}{} & S$^2$GRL  \cite{peng2020self} & PAPC & 83.7 & 72.1 & 82.4 & 66.0 \\  \cline{2-7} % from original paper
		
% 		\multicolumn{1}{l|}{} & TopoTER \cite{gao2021topoter} & PAPC & 83.7 & 71.7 & 79.1 & - \\   \cline{2-7} % from original paper
		
		\multicolumn{1}{l|}{} & DeepWalk  \cite{perozzi2014deepwalk} & NSC & 67.2 & 43.2 & 65.3 & - \\  \cline{2-7} % from SIG-VAE
		
		\multicolumn{1}{l|}{} & GraphSAGE  \cite{hamilton2017inductive} & NSC & 78.7 & 69.4 & 78.1 & 50.2 \\  \cline{2-7} 
		
		\multicolumn{1}{l|}{} & GRACE  \cite{zhu2020deep} & NSC & 80.0 & 71.7 & 79.5 & - \\  \cline{2-7} % from CG3
		
		\multicolumn{1}{l|}{} & GCA  \cite{zhu2020graph} & NSC & 81.2 & 71.8 & 82.8 & - \\  \cline{2-7} 
		
        \multicolumn{1}{l|}{} & GraphCL(N) \cite{hafidi2020graphcl} & NSC & 83.6 & 72.5 & 79.8 & 65.9 \\  \cline{2-7} % from original paper
		
		\multicolumn{1}{l|}{} & BGRL  \cite{thakoor2021bootstrapped} & NSC & 80.5 & 71.0 & 79.5 & - \\  \cline{2-7} 
		
		\multicolumn{1}{l|}{} & G-BT  \cite{bielak2021graph} & NSC & 81.0 & 70.8 & 79.0 & - \\  \cline{2-7} 
		
		\multicolumn{1}{l|}{} & MERIT \cite{Jin2021MultiScaleCS} & NSC & 83.1 & 74.0 & 80.1 & - \\  \cline{2-7} % from original paper
		
		\multicolumn{1}{l|}{} & DGI \cite{velivckovic2018deep} & PGCC & 82.3 & 71.8 & 76.8 & 63.8 \\ \cline{2-7} % from original paper
		
		\multicolumn{1}{l|}{} & MVGRL  \cite{hassani2020contrastive} & PGCC & 82.9 & 72.6 & 79.4 & - \\  \cline{2-7} % from CG3
		
		\multicolumn{1}{l|}{} & SubG-Con \cite{jiao2020sub} & PGCC & 83.5 & 73.2 & 81.0 & 66.9 \\  \cline{2-7} % from original paper
		
		\multicolumn{1}{l|}{} & GMI \cite{peng2020graph} & Hybrid & 82.7 & 73.0 & 80.1 & 65.0 \\  \cline{2-7} 	% from original paper	
		
        \multicolumn{1}{l|}{} & MVMI-FT \cite{fan2021maximizing} & Hybrid & 83.1 & 72.7 & 81.0 & - \\  \cline{2-7} % from original paper
		
		\multicolumn{1}{l|}{} & G-Zoom \cite{zheng2021gzoom} & Hybrid & 84.7 & 74.2 & 81.2 & - \\  \hline % from original paper
		
		\multicolumn{1}{l|}{\multirow{9}{*}
		{PF/JL}} & G. Comp.  \cite{you2020does} & FG & 81.3 & 71.7 & 79.2 \\  \cline{2-7} % from original paper
		
		\multicolumn{1}{l|}{} & SuperGAT \cite{kim2021how} & SG & 84.3 & 72.6 & 81.7 & 74.4\\  \cline{2-7} 
		
		\multicolumn{1}{l|}{} & N. Clu.  \cite{you2020does} & CAPC & 81.8 & 71.7 & 79.2 & -  \\  \cline{2-7}  % from original paper
		
		\multicolumn{1}{l|}{} & M3S  \cite{sun2020multi} & CAPC & 81.6 & 71.9 & 79.3 & - \\  \cline{2-7} % from you2020
		
		\multicolumn{1}{l|}{} & G. Part.  \cite{you2020does} & CAPC & 81.8 & 71.3 & 80.0 & - \\  \cline{2-7} % from original paper
		
		\multicolumn{1}{l|}{} & SimP-GCN \cite{jin2021node} & APR & 82.8 & 72.6 & 81.1 & -  \\  \cline{2-7} % from original paper
		
		\multicolumn{1}{l|}{} & Graph-Bert \cite{zhang2020graph} & Hybrid & 84.3 & 71.2 & 79.3 & -  \\  \cline{2-7} % from original paper
		
		\multicolumn{1}{l|}{} & M. et al. \cite{manessi2020graph} & Hybrid & 82.2 & 71.1 & 79.3 & - \\  \cline{2-7} % from original paper
		
		\multicolumn{1}{l|}{} & CG$^3$ \cite{wan2020contrastive} & Hybrid & 83.4 & 73.6 & 80.2 & - \\  \hline % from original paper
		\toprule
		
	\end{tabular}
	\end{adjustbox}
	\vspace{-4mm}
\end{table}

\vspace{-3mm}
\section{Practical Applications} \label{sec:practical_application}

Graph SSL has also been applied to a wide range of disciplines. We summarize the applications of graph SSL in three research fields. More can be found in Appendix \ref{subsec:other_application}.

\newcommand{\stopic}[1]{\noindent \textbf{#1}.}

\stopic{Recommender Systems} 
Graph-based recommender system has drawn great research attention since it can model items and users with networks and leverage their underlying linkages to produce high-quality recommendations \cite{wu2020comprehensive}. Recently, researchers introduce graph SSL in recommender systems to deal with several issues, including the cold-start problem, pre-training for recommendation model, selection bias, etc. 
For instance, \methodHL{Hao et al.} \cite{hao2021pre} present a reconstruction-based pretext task to pre-train GNNs on the cold-start users and items. 
\methodHL{S$^2$-MHCN} \cite{yu2021self} and \methodHL{DHCN} \cite{xia2021self} employ contrastive tasks for hypergraph representation learning for social- and session- based recommendation, respectively. 
\methodHL{Liu et al.} \cite{liu2021contrastive} overcome the message dropout problem and reduce the selection bias in GNN-based recommender system by introducing a graph contrastive learning module with a debiased loss.
\methodHL{PMGT} \cite{liu2020pre} utilizes two generation-based tasks to capture multimodal side information for recommendation. 

\stopic{Anomaly Detection}
Graph anomaly detection is often performed under an unsupervised scenario due to the lack of annotated anomalies, which is naturally consistent with the setting of SSL \cite{liu2021anomaly}. Hence, various works apply SSL to graph anomaly detection problem. 
To be concrete, \methodHL{DOMINANT} \cite{ding2019deep}, \methodHL{SpecAE} \cite{li2019specae} and \methodHL{AEGIS} \cite{ding2020inductive} employ hybrid SSL frameworks that combine structure and feature generation to capture the patterns of anomalies.
\methodHL{CoLA} \cite{liu2021anomaly} and ANEMONE \cite{jin2021anemone} utilize contrastive learning to detect anomalies on graphs. 
SL-GAD \cite{zheng2021generative} applies hybrid graph SSL to anomaly detection. 
\methodHL{HCM} \cite{huang2021hop} introduces an auxiliary property classification task that predicts the hop-count of each node pair for graph anomaly detection.

\stopic{Chemistry}
In the domain of chemistry, researchers usually model molecules or compounds as graphs where atoms and chemical bonds are denoted as nodes and edges respectively. Note that \methodHL{GROVER} \cite{rong2020self} and \methodHL{Hu et al.} \cite{hu2019strategies} also focus on graph SSL for molecule data, which have been reviewed before.
Additionally, \methodHL{MolCLR} \cite{wang2021molclr} and \methodHL{CKGNN} \cite{fang2021knowledge} learn molecular representations with graph-level contrast-based pretext tasks. Besides, \methodHL{GraphMS} \cite{cheng2021graphms} and \methodHL{MIRACLE} \cite{wang2021multi} employ contrastive learning to solve the drug–target and drug-drug interaction prediction problems. 

\vspace{-3mm}
\section{Future Directions} \label{sec:future_direction}

In this section, we analyze existing challenges in graph SSL and pinpoint a few future research directions aiming to address the shortcomings.

\stopic{Theoretical Foundation}
Despite its great success in various tasks and datasets, graph SSL still lacks a theoretical foundation to prove its usefulness. Most existing methods are mainly designed with intuition and evaluated by empirical experiments. Although MI estimation theory \cite{hjelm2018learning} supports some of the works on contrastive learning, the choice of the MI estimator still relies on empirical studies \cite{hassani2020contrastive}. 
Setting up a solid theoretical foundation for graph SSL is urgently needed. It is desirable to bridge the gap between empirical SSL and fundamental graph theories, including the graph signal processing and spectral graph theory.

\stopic{Interpretability and Robustness}
Graph SSL applications may be risk-sensitive and privacy-related (\textit{e.g.}, fraud detection), an explainable and robust SSL framework is of great significance to adapt to such learning scenarios. However, most existing graph SSL methods only aim to reach a higher performance on downstream tasks with black-box models, ignoring the explainability of learned representations and predicted results. Moreover, except for a few pioneering works \cite{you2020does,jovanovic2021towards} that consider the robustness problem, most graph SSL methods assume input data is perfect, despite the fact that real-world data is often noisy and GNNs are vulnerable to adversarial attacks \cite{jovanovic2021towards}. It would be an interesting and practical direction to explore explainable and robust graph SSL methods in future.

\stopic{Pretext Tasks for Complex Types of Graphs} 
Most current works concentrate on SSL for attributed graphs, and only a few focus on complex graph types, \textit{e.g.}, heterogeneous or spatial-temporal graphs. For complex graphs, the main challenge is how to design pretext tasks to capture unique data characteristics of these complex graphs. Some existing methods use MI maximization \cite{velivckovic2018deep} for complex graph learning, which is limited in its ability to leverage rich information from data, \textit{e.g.}, the temporal dynamics in spatial-temporal/dynamic graphs. A future opportunity is to produce various SSL tasks for complex graph data, where specific data characteristics are the main focus. Furthermore, extending SSL to more ubiquitous graph types (\textit{e.g.}, hypergraphs) would be a feasible direction for further exploration.

\stopic{Augmentation for Graph Contrastive Learning}
In contrastive learning for CV \cite{chen2020simple}, a large amount of augmentation strategies (including rotation, color distort, crop, etc.) provide diverse views of image data, maintaining the representation invariance during contrastive learning. However, due to the nature of graph-structured data (\textit{e.g.}, complex and non-Euclidean structure), data augmentation schemes on graphs are not well explored and thus compromise the effectiveness of graph augmentation-based approaches as discussed in Section \ref{subsec: augmentation}. Most of the existing graph augmentations consider uniformly masking/shuffling node features, modifying edges, or other alternative ways like subgraph sampling and graph diffusion \cite{klicpera2019diffusion}, which provides limited diversity and uncertain invariance when generating multiple graph views. To bridge the gaps, adaptively performing graph augmentations \cite{zhu2020graph}, {automatically} selecting augmentations \cite{you2021graph} or jointly considering stronger augmented samples \cite{wang2021contrastive} by mining the rich underlying structural and attributive information would be interesting directions for further investigation. 

\stopic{Learning with Multiple Pretext Tasks}
Most existing graph SSL approaches learn representations by solving one pretext task, while only a few hybrid methods explore the combination of multiple pretext tasks. As shown in previous NLP pre-training models \cite{devlin2018bert} and the reviewed hybrid methods, the integration of diverse pretext tasks can provide different supervision signals from various perspectives, which facilitates graph SSL methods to produce more informative representations. Therefore, more advanced hybrid approaches that consider a diverse and adaptive combination of multiple pretext tasks deserve further {studies}. 

\stopic{Broader Scope of Applications}
Graphs are ubiquitous data structures in numerous domains; nevertheless, acquiring manual labels is often costly in most application fields. In that case, graph SSL has a promising prospect on a wide range of applications, especially those that highly depend on domain knowledge to annotate data. However, most current practical applications merely concentrate on a few areas (\textit{e.g.}, recommender systems, anomaly detection, and chemistry), indicating that graph SSL holds untapped potential for most application fields.  It is promising to extend graph SSL to more expansive fields of applications, for instance, financial networks, cybersecurity, community detection \cite{jin2021survey}, and federated learning.

\vspace{-3mm}
\section{Conclusion}\label{sec:conclusion}

This paper conducts a comprehensive overview of self-supervised learning on graphs. We present a unified framework and further provide a systematic taxonomy that groups graph SSL into four categories: generation-based, auxiliary property-based, contrast-based, and hybrid approaches. For each category, we provide mathematical summary, up-to-date review, and comparison between methods. More importantly, we collect abundant resources including datasets, evaluation methods, performance comparison and open-source codes for graph SSL approaches. A wide range of practical applications of graph SSL are also introduced in our paper. Finally, we suggest open challenges and promising research directions of graph SSL in the future. 

%% file: supplemental.tex
\begin{figure*}[t!]
	\centering
	\includegraphics[width=1\textwidth]{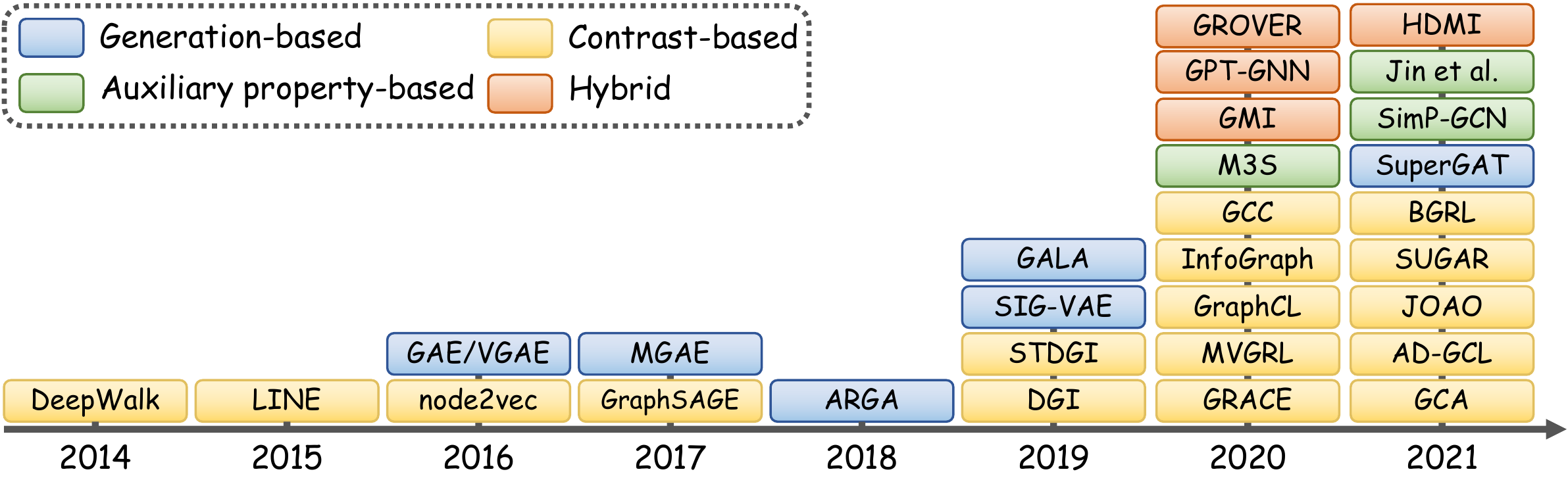}
	\caption{A timeline of graph self-supervised learning (SSL) development.}
	\label{fig:timeline}
\vspace{-3mm}
\end{figure*}

\section{Timeline of Graph Self-Supervised Learning} \label{appendix:timeline}

{Fig. \ref{fig:timeline} provides a timeline of some of the most representative graph SSL methods since 2014. A group of early works is the random walk-based node embedding methods including DeepWalk \cite{perozzi2014deepwalk}, LINE \cite{tang2015line}, and node2vec \cite{grover2016node2vec}. Another line of early works is the graph autoencoder-based generation methods, i.e. GAE/VGAE \cite{kipf2016variational}, MGAE \cite{wang2017mgae}, ARGA \cite{pan2018adversarially}, and SIG-VAE \cite{hasanzadeh2019semi}. In 2019, a representation contrast-based work, DGI \cite{velivckovic2018deep}, was proposed, bringing the flourishing development of graph contrastive learning. In 2020, more types graph SSL methods were introduced, containing the first auxiliary property-based method (M3S \cite{sun2020multi}) and hybrid method (GMI \cite{peng2020graph}). In 2021, more advanced techniques are integrated with graph SSL, such as adaptive augmentation (GCA \cite{zhu2020graph}), automatic machine learning (JOAO \cite{you2021graph}), and adversarial augmentation (AD-GCL \cite{suresh2021adversarial}).}

\section{Notations and Definitions} \label{appendix:notations_definitions}

\subsection{Notations} \label{appendix:notations}

Throughout this paper, we use bold uppercase letters (e.g. $\mathbf{X}$), bold lowercase letters (e.g. $\mathbf{x}$), and calligraphic fonts (e.g. $\mathcal{V}$) to denote matrices, vectors and sets, respectively. Unless specified otherwise, the notations used in this paper are summarized in Table \ref{table:notation}.

\subsection{Dynamic and Heterogeneous Graphs} \label{appendix:graphs}

\begin{definition}[Dynamic Graph]
	A dynamic graph indicates a graph where nodes and edges change with respect to time, i.e., $\mathcal{G}^{(t)}=(\mathcal{V}^{(t)}, \mathcal{E}^{(t)})$, which denotes a dynamic graph at time $t \in \mathbb{R}^{+}$ that consists of a node set $\mathcal{V}^{(t)}$ and an edge set $\mathcal{E}^{(t)}$ indexed by time. In practice, a dynamic graph is also known as temporal dependency interaction graph, where an interaction at time $t$ is represented as $e_{i,j,t} \in \mathcal{E}^{(t)}$, where node $v_i \in \mathcal{V}^{(t)}$ and $v_j \in \mathcal{V}^{(t)}$. For an attributed dynamic graph with node and edge features, we can also let $\mathcal{G}^{(t)}=(\mathbf{A}^{(t)}, \mathbf{X}_{\mathbf{node}}^{(t)}, \mathbf{X}_{\mathbf{edge}}^{(t)})$, where the adjacency matrix $\mathbf{A}^{(t)} \in \mathbb{R}^{n \times n}$, node feature matrix $\mathbf{X}_{\mathbf{node}}^{(t)} \in \mathbb{R}^{n \times d_{node}}$ and edge feature matrix $\mathbf{X}_{\mathbf{edge}}^{(t)} \in \mathbb{R}^{n \times d_{node}}$ are evolving with respect to time. 
\end{definition}

\noindent\textbf{Special type}: \textit{Spatial-temporal graph} can be regarded as an special type of attributed dynamic graph with fixed graph adjacency and dynamic features at different time steps. Concretely, at each time step $t$, the dynamic feature matrix is denoted as $\mathbf{X}^{(t)} \in \mathbb{R}^{n \times d}$.

\vspace{1mm}

\begin{table}[t]
\centering
\caption{Commonly used notations with explanations.} 
\begin{tabular}{ l|l}  
\toprule[1.0pt]
Notation & Explanation  \\
\cmidrule{1-2}
$|\cdot|$ & The length of a set. \\
$\circ$ & Hadamard product. \\
\cmidrule{1-2}
$\mathcal{G}$ & A graph. \\
$\mathcal{V}$ & The set of nodes in a graph. \\
$\mathcal{E}$ & The set of edges in a graph. \\
$v_i$ & A node in the node set $\mathcal{V}$. \\
$e_{i,j}$ & An edge in the edge set $\mathcal{E}$. \\
$n$ & The number of nodes in a graph. \\
$m$ & The number of edges in a graph. \\
$\mathbf{A} \in \mathbb{R}^{n \times n}$ & The adjacency matrix of a graph. \\
$d$, $d_{node}$ & The dimension of node features. \\ 
$d_{edge}$ & The dimension of edge features. \\ 
$\mathbf{X}$, $\mathbf{X_{node}} \in \mathbb{R}^{n \times d }$ & The node feature matrix of a graph. \\
$\mathbf{X_{edge}} \in \mathbb{R}^{m \times d_{edge} }$ & The edge feature matrix of a graph. \\
$\mathbf{x_{i}}=[\mathbf{X}]_{v_i}$ & The node feature of $v_i$. \\
$d_r$ & The dimension of node representation. \\
$\mathbf{H}\in \mathbb{R}^{n \times d_r}$ & The node representation matrix of a graph. \\
$\mathbf{h_i}=[\mathbf{H}]_{v_i}$ & The node representation vector of $v_i$. \\
$\mathbf{h}_{\mathcal{G}}$ & The graph representation vector of a graph. \\
$\mathcal{Y}$ & Manual label set of downstream task. \\
$\mathcal{C}$ & Pseudo label set of pretext task. \\
$y$ & Manual label of downstream task. \\
$c$ & Pseudo label of pretext task. \\
$\tilde{\mathcal{G}}$ & Augmented/Perturbed graph. \\
\cmidrule{1-2}
$f_{\theta}(\cdot)$ & Encoder parameterized by $\theta$. \\
$p_{\phi}(\cdot)$ & Pretext decoder parameterized by $\phi$. \\
$q_{\psi}(\cdot)$ & Downstream decoder parameterized by $\psi$. \\
$\mathcal{L}(\cdot)$ & Loss function. \\
$\mathcal{R}(\cdot)$ & Readout function. \\
$\Omega(\cdot)$ & Auxiliary property mapping function. \\
$\mathcal{MI}(\cdot, \cdot)$ & Mutual information function. \\
\bottomrule[1.0pt]
\end{tabular}
\label{table:notation}
\end{table}

\begin{definition}[Heterogeneous Graph]
	A heterogeneous graph denotes a graph consisting of different types of nodes and/ or edges. For a heterogeneous graph $\mathcal{G}=(\mathcal{V}, \mathcal{E})$, each node $v_i \in \mathcal{V}$ and each edge $e_{i,j} \in \mathcal{E}$ are associated with a corresponding mapping function, i.e., $\phi_v (v_i): \mathcal{V} \to \mathcal{S}_v$ and $\phi_e (e_{i,j}): \mathcal{E} \to \mathcal{S}_e$. $\mathcal{S}_v$ and $\mathcal{S}_e$ are the node types and link types, respectively, and they satisfy $\lvert \mathcal{S}_v \rvert + \lvert \mathcal{S}_e \rvert > 2$.
\end{definition}

\noindent\textbf{Special types}: \textit{Bipartite graph} and \textit{multiplex graph} can be viewed as two special types of heterogeneous graphs. Specifically, a bipartite graph is a heterogeneous graph with two types of nodes ($\lvert \mathcal{S}_v \rvert = 2$) and a single type of edge ($\lvert \mathcal{S}_e \rvert = 1$); a multiplex graph has one type of node ($\lvert \mathcal{S}_v \rvert = 1$) and multiple types of edges ($\lvert \mathcal{S}_e \rvert > 1$).

\vspace{1mm}

\subsection{Graph Neural Networks and Readout Layers} \label{appendix:gnns}

A general definition of Graph Neural Networks (GNNs) is represented as:

\begin{definition}[Graph Neural Networks]
    Given an attributed graph $\mathcal{G}$ with its feature matrix $\mathbf{X}$ where $\mathbf{x_{i}}=\mathbf{X}[i,:]^{T}$ is a $d$-dimensional feature vector of the node $v_i$, a GNN learns a node representation $\mathbf{h_i}$ for each node $v_i \in \mathcal{V}$ with two core functions: the aggregation function and combination function. Considering a $K$-layer GNN, the $k$-th layer performs:
    
    \begin{equation}
    \begin{aligned}
    \mathbf{a_{i}^{(k)}}&=\operatorname { aggregate }^{(k)}\left(\left\{\mathbf{h_{j}^{(k-1)}}: v_j \in \mathcal{N}(v_i)\right\}\right), \\
    \mathbf{h_{i}^{(k)}}&=\operatorname{combine}^{(k)}\left(\mathbf{h_{i}^{(k-1)}}, \mathbf{a_{i}^{(k)}}\right), 
    \end{aligned}
    \end{equation}
    
    where $\mathbf{h_{i}^{(k)}}$ is the latent vector of node $v_i$ at the $k$-th iteration/layer with $\mathbf{h_{i}^{(0)}} = \mathbf{x_{i}}$ and $\mathbf{h_{i}^{(K)}} = \mathbf{h_{i}}$,  
    $\operatorname { aggregate }^{(k)} (\cdot)$ and $\operatorname { combine }^{(k)} (\cdot)$ are the aggregation and combination function of the $k-th$ layer, respectively. The design of the core functions is crucial to different GNNs, and we refer the reader to the recent survey \cite{wu2020comprehensive} for a thorough review.
\end{definition}

To learn the property of each node, the node representation $\mathbf{h_{i}}$ is used for downstream tasks directly. Nevertheless, when learning the representation of the entire graph, an extra readout layer is needed, which is defined as:

\begin{definition}[Readout layer]
    Given a graph $\mathcal{G}=(\mathcal{V}, \mathcal{E})$ with its node representation $\mathbf{h_{0}^{(K)}} , \cdots , \mathbf{h_{n}^{(K)}}$, a readout layer generates the graph representation $\mathbf{h_{\mathcal{G}}}$ by:

    \begin{equation}
    \mathbf{h_{\mathcal G}}=\mathcal{R}\left(\left\{\mathbf{h_{i}}^{(K)} \mid v_i \in \mathcal{V}\right\}\right)
    \end{equation}  
    
    where $\mathcal { R }(\cdot)$ is the readout function that fuses the node-level representations into a graph-level representation. It can be a simple permutation invariant function like summation or complex pooling algorithms like DiffPool \cite{ying2018hierarchical} and HaarPool \cite{wang2020haar}.
\end{definition}

% \vspace{-3mm}
\subsection{{Commonly Used Loss Functions}} \label{appendix:loss}

{We formulate two commonly used loss functions, i.e., mean squared error (MSE) loss and cross-entropy (CE) loss. }

\begin{definition}[Mean squared error (MSE) loss]

{Given a predicted vector $\hat{\mathbf{y}} \in \mathbb{R}^{d_y}$ and a target vector $\mathbf{y}\in \mathbb{R}^{d_y}$, the formulation of MSE loss is defined as follows:}

\begin{equation}
\mathcal{L}_{mse}(\hat{\mathbf{y}},\mathbf{y}) = \frac{1}{d_y} \sum_{i=1}^{d_y}({\hat{y}}_i - y_i)^2,
\label{eq: mse_loss}
\end{equation}

{where $\hat{y}_i$ and $y_i$ are the $i$-th element of $\hat{\mathbf{y}}$ and ${\mathbf{y}}$, respectively. Note that $\mathcal{L}_{mse}$ can also be applied to matrices and scalars.}

\end{definition}

\begin{definition}[Cross-entropy (CE) loss]

{Given a predicted vector $\hat{\mathbf{y}} \in \mathbb{R}^{d_y}$ after Softmax activation and a one-hot target vector $\mathbf{y}\in \mathbb{R}^{d_y}$, the formulation of CE loss is defined as follows:}

\begin{equation}
\mathcal{L}_{ce}(\hat{\mathbf{y}},\mathbf{y}) = - \sum_{i=1}^{d_y}y_i \log{{\hat{y}}_i}.
\label{eq: ce_loss}
\end{equation}

{Specially, the binary version of CE loss is denoted as binary cross-entropy (BCE) loss, which is expressed by:}

\begin{equation}
\mathcal{L}_{bce}(\hat{y},y) = - y\log{{\hat{y}}} - (1 - y)\log{{(1-\hat{y})}},
\label{eq: bce_loss}
\end{equation}

{{where $\hat{{y}} \in [0,1]$ and ${{y}} \in \{0,1\}$ are the predicted and target scalars, respectively.}}

\end{definition}

\section{Downstream Tasks} \label{appendix:downstream_tasks}

\subsection{Node-level tasks} 

Node-level tasks includes node regression and node classification. As an representative task, node classification aims to predict the label $y_i \in \mathcal{Y}$ for each node $v_i \in \mathcal{V}$, where $\mathcal{Y}$ is a finite and discrete label set. 
A typical downstream decoder $q_{\psi}$ for node classification is an MLP layer with Softmax activation that takes node representation as its input. The cross-entropy (CE) loss $\mathcal{L}_{ce}$ is typically adopted for  model training. The formulation of node classification is defined as follows based on Equation (\ref{eq: downstrem task}):

\begin{equation}
f_{\theta^{*}}, q_{\psi^{*}} = \mathop{\arg\min}\limits_{f_{\theta^{*}}, q_{\psi}} \frac{1}{\lvert \mathcal{V}_L \rvert}  \sum_{v_i \in \mathcal{V}_L} \mathcal{L}_{ce}\Big( q_{\psi}\big([ f_{\theta}(\mathcal{G})]_{v_i}\big),y_i \Big),
\label{eq: downstream_node}
\end{equation}

\noindent where $\mathcal{V}_L $ is the training node set (where each node $v_i \in \mathcal{V}_L$ has a known label $y_i$ for training), and $[\cdot]_{v_i}$ is a picking function that index the embedding of $v_i$ from the whole embedding matrix.

\subsection{Link-level tasks} 

Link-level tasks includes edge classification and link prediction.
Taking edge classification as an example, given an edge $(v_i,v_j)$, the goal is to predict its label $y_{i,j} \in \mathcal{Y}$ from a given label set.
A downstream decoder $q_{\psi}$ could be a classifier with the embeddings of two nodes as input, while CE $\mathcal{L}_{ce}$ serves as the loss function. We formalize the objective function as follows:

\begin{equation}
f_{\theta^{*}}, q_{\psi^{*}} = \mathop{\arg\min}\limits_{f_{\theta^{*}}, q_{\psi}} \frac{1}{\lvert \mathcal{E}_L \rvert}  \sum_{v_i,v_j \in \mathcal{E}_L}  \mathcal{L}_{ce}\Big( q_{\psi}\big([ f_{\theta}(\mathcal{G})]_{v_i,v_j}\big),y_{i,j} \Big),
\label{eq: downstream_edge}
\end{equation}

\noindent where $\mathcal{E}_L $ is the training edge set, and $[\cdot]_{v_i,v_j}$ is a picking function that index the embedding of $v_i$ and $v_j$ from the whole embedding matrix.

\subsection{Graph-level Tasks}

Graph-level tasks, including graph classification and graph regression, often rely on graph-level representations. For instance, in graph classification task, each graph $\mathcal{G}_i$ has its {target value} $y_i \in \mathcal{Y}$, and the objective is to train a model to predict the labels of input graphs. In general, an encoder $f_\theta$ first learns node embeddings and then aggregates them into a graph embedding via a readout function. After that, the graph embedding is fed into the decoder $q_\psi$ composed by a classifier with Softmax function. The model is trained by CE loss $\mathcal{L}_{ce}$, which can be formalized as:

\begin{equation}
f_{\theta^{*}} = \mathop{\arg\min}\limits_{f_{\theta}, p_{\phi}} \frac{1}{\lvert \mathcal{U}_L \rvert}  \sum_{\mathcal{G}_i \in \mathcal{U}_L}  \mathcal{L}_{ce}\Big( p_{\phi}\big(f_{\theta}(\mathcal{G}_i)\big), y_i \Big),
\label{eq: downstream_graph}
\end{equation}

\noindent where $\mathcal{U}_L $ is the training graph set.

\section{{Performance Comparison of Graph Classification}} \label{appendix:evaluation}

The task of graph classification often adapts an inductive supervised learning setting. We collect the experimental results on nine benchmark datasets, including IMDB-B, IMDB-M, RDT-B, RDT-M, COLLAB, MUTAG, PROTEINS, PTC, and NCI-1 \cite{yanardag2015deep}. We consider the dataset split which is utilized in \cite{xu2018how}. Classification accuracy is leveraged to evaluate the performance. 
Similar to node classification, we divide the approaches into two groups: URL and PF/JL. A powerful GNN for graph classification, GIN \cite{xu2018how}, is employed as our baseline. 

The evaluation results are demonstrated in Table \ref{tab:result_graph_classification}. Based on the results, we have the following observations: 
(1) The overall performance of the methods in URL group is lower than that of the supervised baseline (i.e., GIN). \hlt{The reason is that current SSL methods cannot learn optimal representations as supervised methods do. Considering the URL methods for node classification usually achieve higher results than baselines, future SSL methods for graph-level representation learning are expected to reach higher performance. }
(2) The cross-scale contrast-based methods, in general, have better performance than the same-scale contrast-based methods. \hlt{A possible reason is that cross-scale contrastive learning can improve the quality of global-wise (i.e., graph-level) representations, which benefits graph-level learning tasks. }
(3) Taking GCC as an example, the involvement of graph labels (i.e., PF/JL schemes) does not necessarily lead to better encoders \hlt{on all datasets} compared to pure unsupervised learning (URL) scheme. \hlt{We conclude that pre-trained encoders with SSL sometimes have a negative effect on downstream tasks. In this case, how to circumvent such a negative transfer phenomenon in graph-level SSL deserves further investigation. 
(4) Most of these methods are measured on different sets of graph-level datasets due to the lack of general benchmarks for evaluation. Therefore, unified benchmarks for graph-level SSL evaluation are significantly needed in future works. 
}

\begin{table*}[htbp]
 	\caption{A summary of experimental results for graph classification in nine benchmark datasets.}
	\label{tab:result_graph_classification}
	\centering
	\begin{tabular}{l l l c c c c c c c c c}
		\toprule
		
		\multicolumn{1}{l}{Group} & Approach & Category & IMDB-B & IMDB-M & RDT-B & RDT-M & COLLAB & MUTAG & PROTEINS & PTC & NCI-1\\ 
		\bottomrule
% 		\hline
		\multicolumn{1}{l|}{\multirow{1}{*}
		{Baselines}} & GIN \cite{xu2018how} & - & 75.1 & 52.3 & 92.4 & 57.5 & 80.2 & 89.4 & 76.2 & 64.6 & 82.7 \\  \hline 
		
		\multicolumn{1}{l|}{\multirow{8}{*}
		{URL}} & GCC \cite{qiu2020gcc} & NSC & 72.0 & 49.4 & 89.8 & 53.7 & 78.9 & - & - & - & -\\  \cline{2-12} % 

		\multicolumn{1}{l|}{} & GraphCL(G) \cite{you2020graph} & GSC & 71.1 & - & 89.5 & - & 71.4 & 86.8 & 74.4 & - & 77.9\\  \cline{2-12} 

        \multicolumn{1}{l|}{} & AD-GCL \cite{suresh2021adversarial} & GSC & 72.3 & 49.9 & 85.5 & 54.9 & 73.3 & 89.7 & 73.8 & - & - \\  \cline{2-12} 

        \multicolumn{1}{l|}{} & JOAO \cite{you2021graph} & GSC & 70.8 & - & 86.4 & 56.0 & 69.5 & 87.7 & 74.6 & - & 78.3\\  \cline{2-12} 
        
        \multicolumn{1}{l|}{} & IGSD \cite{zhang2020iterative} & GSC & 74.7 & 51.5 & - & - & 70.4 & 90.2 & - & 61.4 & 75.4\\   \cline{2-12}  

		\multicolumn{1}{l|}{} & MVGRL \cite{hassani2020contrastive} & PGCC & 74.2 & 51.2 & 84.5 & - & - & 89.7 & - & - & -\\  \cline{2-12} 
        
        \multicolumn{1}{l|}{} & InfoGraph \cite{sun2019infograph} & PGCC & 73.0 & 49.7 & 82.5 & 53.5 & - & 89.0 & - & - & \\ \cline{2-12}  
		
		\multicolumn{1}{l|}{} & HTC \cite{wang2021learning} & CGCC & 73.3 & 50.6 & 91.3 & 55.2 & - & 91.8 & - & - & -\\  \hline
		
		\multicolumn{1}{l|}{\multirow{3}{*}
		{PF/JL}} & GCC \cite{qiu2020gcc} & NSC & 73.8 & 50.3 & 87.6 & 53.0 & 81.1 & - & - & - & -\\  \cline{2-12} 
		
		\multicolumn{1}{l|}{} & CSSL \cite{zeng2020contrastive} & GSC & - & - & - & - & - & - & 85.8 & - & 80.1\\  \cline{2-12}

        \multicolumn{1}{l|}{} & LCGNN \cite{ren2021label} & GSC & 76.1 & 52.4 & - & - & 77.5 & 90.5 & 85.2 & 65.9 & 82.9\\  \hline 
        
		\toprule
		
	\end{tabular}
\end{table*}

\section{Datasets} \label{appendix:datasets}

In this section, we conduct an introduction and a summary of commonly used datasets in four categories, including citation networks, co-purchase networks, social networks, and bio-chemical graphs. A statistic of these datasets is given in Table \ref{tab:dataset}.

\begin{table*}[htbp]
 	\caption{A summary of selected benchmark datasets. The markers ``*'' in the ``\# Classes'' column indicate that it is a multi-label classification dataset.}
	\label{tab:dataset}
	\centering
	\begin{tabular}{l l l l l l l l }
		\toprule
		
		\multicolumn{1}{l}{Category} & Dataset &\# Graphs & \# Nodes(Avg.) & \#     Edges (Avg.) & \#Features & \# Classes & Citation \\ 
		\bottomrule
% 		\hline
		\multicolumn{1}{l|}{\multirow{7}{*}
		{\begin{tabular}[c]{@{}l@{}}Citation\\Networks \end{tabular}}} & Cora \cite{sen2008collective} &1 & 2,708 & 5,429 & 1,433 & 7 &  \begin{tabular}[c]{@{}l@{}}
\cite{you2020does,jin2020self,manessi2020graph,sun2020multi,hu2019pre,zhu2020cagnn,kipf2016variational}\\ \cite{peng2020self,kim2021how,zhu2020deep,velivckovic2018deep,hassani2020contrastive,jiao2020sub,mavromatis2020graph}\\ \cite{peng2020graph,zhang2020graph,wan2020contrastive,wang2017mgae,park2019symmetric,pan2018adversarially,hasanzadeh2019semi}\\ \cite{zhu2020self,jovanovic2021towards,kefato2021self,Jin2021MultiScaleCS,kou2021self,jin2021node,hafidi2020graphcl}\\ \cite{fan2021maximizing,zheng2021gzoom,wan2021contrastive,roy2021node}\\ 
		\end{tabular}\\ \cline{2-8}  
		
		\multicolumn{1}{l|}{} & Citeseer \cite{sen2008collective} &1 & 3,327 & 4,732 & 3,703 & 6 &        \begin{tabular}[c]{@{}l@{}}\cite{you2020does,jin2020self,manessi2020graph,sun2020multi,zhu2020cagnn,kipf2016variational,peng2020self}\\ \cite{kim2021how,zhu2020deep,velivckovic2018deep,hassani2020contrastive,jiao2020sub,mavromatis2020graph,peng2020graph}\\ \cite{zhang2020graph,wan2020contrastive,wang2017mgae,park2019symmetric,pan2018adversarially,hasanzadeh2019semi,zhu2020self}\\ \cite{jovanovic2021towards,kefato2021self,Jin2021MultiScaleCS,kou2021self,jin2021node,hafidi2020graphcl,fan2021maximizing}\\ \cite{zheng2021gzoom,wan2021contrastive,roy2021node}\\
		\end{tabular}\\ \cline{2-8} 
		
		\multicolumn{1}{l|}{}    & Pubmed \cite{sen2008collective} &1 & 19,717 & 44,338 & 500  & 3 & \begin{tabular}[c]{@{}l@{}} \cite{you2020does,jin2020self,manessi2020graph,sun2020multi,hu2019pre,zhu2020cagnn,kipf2016variational}\\ \cite{peng2020self,kim2021how,zhu2020deep,velivckovic2018deep,jiao2020sub,mavromatis2020graph,peng2020graph}\\ \cite{zhang2020graph,wan2020contrastive,park2019symmetric,pan2018adversarially,hasanzadeh2019semi,zhu2020self,jovanovic2021towards}\\ \cite{kefato2021self,Jin2021MultiScaleCS,kou2021self,jin2021node,hafidi2020graphcl,fan2021maximizing,wan2021contrastive}\\\cite{roy2021node} \\
		\end{tabular} \\ \cline{2-8} 
		
		\multicolumn{1}{l|}{} & {\begin{tabular}[c]{@{}l@{}}CoAuthor \\CS \cite{shchur2018pitfalls}\end{tabular}} &1 & 18,333 & 81,894 & 6,805 & 15 &        \begin{tabular}[c]{@{}l@{}}\cite{kim2021how,zhu2020graph,mavromatis2020graph,wan2020contrastive,thakoor2021bootstrapped,kefato2021self,Jin2021MultiScaleCS}\\ \cite{bielak2021graph}\\
		\end{tabular}\\ \cline{2-8} 
		
		\multicolumn{1}{l|}{} & {\begin{tabular}[c]{@{}l@{}}CoAuthor \\Physics \cite{shchur2018pitfalls}\end{tabular}} &1 & 34,493 & 247,962 & 8,415 & 5 &        \begin{tabular}[c]{@{}l@{}}\cite{kim2021how,zhu2020graph,mavromatis2020graph,thakoor2021bootstrapped,kefato2021self,bielak2021graph}\\
		\end{tabular}\\ \cline{2-8} 
		
		\multicolumn{1}{l|}{} & Wiki-CS \cite{mernyei2020wiki} &1 & 11,701 & 216,123 & 300 & 10 &        \begin{tabular}[c]{@{}l@{}}\cite{kim2021how,zhu2020graph,thakoor2021bootstrapped,jovanovic2021towards,bielak2021graph}\\
		\end{tabular}\\ \cline{2-8} 
		
		\multicolumn{1}{l|}{} & ogbn-arxiv \cite{hu2020ogb} &1 & 169,343 & 1,166,243 & 128 & 40 &        \begin{tabular}[c]{@{}l@{}}\cite{kim2021how,thakoor2021bootstrapped,bielak2021graph}\\
		\end{tabular}\\ \hline

		\multicolumn{1}{l|}{\multirow{2}{*}{\begin{tabular}[c]{@{}l@{}}Co-\\purchase\\Networks \end{tabular}}} & {\begin{tabular}[c]{@{}l@{}}Amazon \\Photo \cite{shchur2018pitfalls}\end{tabular}}  &1 & 7,650 & 119,081 & 745 & 8 &        \begin{tabular}[c]{@{}l@{}}\cite{kim2021how,zhu2020graph,mavromatis2020graph,wan2020contrastive,thakoor2021bootstrapped,jovanovic2021towards,wan2021contrastive}\\ \cite{kefato2021self,Jin2021MultiScaleCS,fan2021maximizing,bielak2021graph}\\
		\end{tabular}\\ \cline{2-8} 
		
		\multicolumn{1}{l|}{} & {\begin{tabular}[c]{@{}l@{}}Amazon \\Computers \cite{shchur2018pitfalls}\end{tabular}}  &1 & 13,752 & 245,861 & 767 & 10 &        \begin{tabular}[c]{@{}l@{}}\cite{kim2021how,zhu2020graph,mavromatis2020graph,wan2020contrastive,thakoor2021bootstrapped,kefato2021self,fan2021maximizing}\\ \cite{bielak2021graph}\\
		\end{tabular}\\ \hline

		\multicolumn{1}{l|}{\multirow{6}{*}{\begin{tabular}[c]{@{}l@{}}Social  \\Networks\end{tabular}}} & Reddit \cite{hamilton2017inductive} &1 &    232,965             & 11,606,919 & 602 & 41 &     \begin{tabular}[c]{@{}l@{}}\cite{jin2020self,hamilton2017inductive,peng2020self,zhu2020deep,velivckovic2018deep,jiao2020sub,hu2020gpt}\\ \cite{peng2020graph,hafidi2020graphcl}  \\  \end{tabular}                                                 \\ \cline{2-8} 

        \multicolumn{1}{l|}{} & IMDB-B \cite{yanardag2015deep} &1,000 & 19.77 & 193.06 & - & 2 &        \begin{tabular}[c]{@{}l@{}}\cite{hu2019pre,qiu2020gcc,zhang2020iterative,hassani2020contrastive,sun2019infograph,verma2020towards,ren2021label}\\ \cite{wang2021learning,robinson2020contrastive,suresh2021adversarial}\\
		\end{tabular}\\ \cline{2-8} 
		
		\multicolumn{1}{l|}{} & IMDB-M \cite{yanardag2015deep} &1,500 & 13.00 & 65.93 & - & 3 &        \begin{tabular}[c]{@{}l@{}}\cite{hu2019pre,qiu2020gcc,zhang2020iterative,hassani2020contrastive,sun2019infograph,verma2020towards,ren2021label}\\ \cite{wang2021learning,robinson2020contrastive,suresh2021adversarial}\\
		\end{tabular}\\ \cline{2-8} 
		
		\multicolumn{1}{l|}{} & RDT-B \cite{yanardag2015deep} &2,000 & 429.63 & 497.75 & - & 2 &        \begin{tabular}[c]{@{}l@{}}\cite{qiu2020gcc,you2020graph,hassani2020contrastive,sun2019infograph,verma2020towards,wang2021learning,robinson2020contrastive}\\ \cite{suresh2021adversarial}\\
		\end{tabular}\\ \cline{2-8} 
		
		\multicolumn{1}{l|}{} & RDT-M \cite{yanardag2015deep} &4,999 & 594.87 & 508.52 & - & 5 &        \begin{tabular}[c]{@{}l@{}}\cite{qiu2020gcc,sun2019infograph,verma2020towards,wang2021learning,suresh2021adversarial}\\
		\end{tabular}\\ \cline{2-8} 
		
		\multicolumn{1}{l|}{} & COLLAB \cite{yanardag2015deep} &5,000 & 74.49 & 2,457.78 & - & 3 &        \begin{tabular}[c]{@{}l@{}}\cite{qiu2020gcc,you2020graph,zhang2020iterative,ren2021label,suresh2021adversarial}\\
		\end{tabular}\\ \hline

		\multicolumn{1}{l|}{\multirow{16}{*}{\begin{tabular}[c]{@{}l@{}}Bio-\\chemical \\Graphs\end{tabular}}}   & PPI \cite{zitnik2017predicting}     & 24    & 56,944 & 818,716        & 50         & 121*       &    \begin{tabular}[c]{@{}l@{}}\cite{hamilton2017inductive,peng2020self,kim2021how,zhu2020deep,velivckovic2018deep,jiao2020sub,peng2020graph}\\ \cite{thakoor2021bootstrapped,hafidi2020graphcl,bielak2021graph}\end{tabular} \\ \cline{2-8}  
		
		\multicolumn{1}{l|}{} & MUTAG \cite{debnath1991structure} & 188 & 17.93 & 19.79 & 7  & 2 &        \begin{tabular}[c]{@{}l@{}}\cite{zhang2020iterative,hassani2020contrastive,sun2019infograph,sun2021sugar,verma2020towards,ren2021label,wang2021learning}\\ \cite{robinson2020contrastive,you2021graph,suresh2021adversarial}\\
		\end{tabular}\\ \cline{2-8} 
		
		\multicolumn{1}{l|}{} & PROTEINS \cite{borgwardt2005protein} &1,113 & 39.06 & 72.81 & 4 & 2 &        \begin{tabular}[c]{@{}l@{}}\cite{you2020graph,zeng2020contrastive,sun2021sugar,ren2021label,robinson2020contrastive,you2021graph,suresh2021adversarial}\\
		\end{tabular}\\ \cline{2-8} 
		
		\multicolumn{1}{l|}{} & D\&D \cite{dobson2003distinguishing} & 1,178 & 284.31 & 715.65 & 82 & 2 & \begin{tabular}[c]{@{}l@{}}\cite{zeng2020contrastive,sun2021sugar,ren2021label,robinson2020contrastive,you2021graph,suresh2021adversarial}\\
		\end{tabular}\\ \cline{2-8} 
		
		\multicolumn{1}{l|}{} & PTC \cite{toivonen2003statistical} &349 & 14.10 & 14.50 & 19 & 2 &        \begin{tabular}[c]{@{}l@{}}\cite{zhang2020iterative,sun2021sugar,ren2021label,robinson2020contrastive}\\
		\end{tabular}\\ \cline{2-8} 
		
		\multicolumn{1}{l|}{} & PTC-MR \cite{helma2001predictive} &344 & 14.29 & 14.69 & 18 & 2 &        \begin{tabular}[c]{@{}l@{}}\cite{hassani2020contrastive,sun2019infograph,verma2020towards,wang2021learning}\\
		\end{tabular}\\ \cline{2-8} 		
		
		\multicolumn{1}{l|}{} & NCI-1 \cite{wale2008comparison} &4,110 & 29.87 & 32.30 & 37 & 2 &        \begin{tabular}[c]{@{}l@{}}\cite{you2020graph,zeng2020contrastive,zhang2020iterative,sun2021sugar,ren2021label,you2021graph,suresh2021adversarial}\\
		\end{tabular}\\ \cline{2-8} 
		
		\multicolumn{1}{l|}{} & NCI-109 \cite{wale2008comparison} & 4,127 & 29.68 & 32.13 & 38 & 2 &        \begin{tabular}[c]{@{}l@{}}\cite{zeng2020contrastive,sun2021sugar}\\
		\end{tabular}\\ \cline{2-8} 
		
		\multicolumn{1}{l|}{} & BBBP \cite{martins2012bayesian} &2,039 & 24.06 & 25.95 & - & 2 &        \begin{tabular}[c]{@{}l@{}}\cite{hu2019strategies,rong2020self,zhang2020motif,xu2021self,you2021graph,suresh2021adversarial}\\
		\end{tabular}\\ \cline{2-8} 
		
		\multicolumn{1}{l|}{} & Tox21 \cite{tox21} &7,831 & 18.51 & 25.94 & - & 12* &        \begin{tabular}[c]{@{}l@{}}\cite{hu2019strategies,rong2020self,zhang2020motif,xu2021self,you2021graph,suresh2021adversarial}\\
		\end{tabular}\\ \cline{2-8} 
		
		\multicolumn{1}{l|}{} & ToxCast \cite{richard2016toxcast} & 8,575 & 18.78 & 19.26 & - & 167* &        \begin{tabular}[c]{@{}l@{}}\cite{hu2019strategies,rong2020self,zhang2020motif,xu2021self,you2021graph,suresh2021adversarial}\\
		\end{tabular}\\ \cline{2-8} 
		
		\multicolumn{1}{l|}{} & SIDER \cite{kuhn2016sider} & 1,427 & 33.64 & 35.36 & - & 27* &        \begin{tabular}[c]{@{}l@{}}\cite{hu2019strategies,rong2020self,zhang2020motif,xu2021self,you2021graph,suresh2021adversarial}\\
		\end{tabular}\\ \cline{2-8} 
		
		\multicolumn{1}{l|}{} & ClinTox \cite{novick2013sweetlead} & 1,478 & 26.13 & 27.86 & - & 2 &        \begin{tabular}[c]{@{}l@{}}\cite{hu2019strategies,rong2020self,zhang2020motif,xu2021self,you2021graph,suresh2021adversarial}\\
		\end{tabular}\\ \cline{2-8} 
		
		\multicolumn{1}{l|}{} & MUV \cite{gardiner2011effectiveness} & 93,087 & 24.23 & 26.28 & - & 17* &        \begin{tabular}[c]{@{}l@{}}\cite{hu2019strategies,xu2021self,you2021graph,suresh2021adversarial}\\
		\end{tabular}\\ \cline{2-8} 
		
		\multicolumn{1}{l|}{} & HIV \cite{hiv2017} & 41,127 & 25.53 & 27.48 & - & 2 &        \begin{tabular}[c]{@{}l@{}}\cite{hu2019strategies,zhang2020motif,xu2021self,you2021graph,suresh2021adversarial}\\
		\end{tabular}\\ \cline{2-8} 
		
		\multicolumn{1}{l|}{} & BACE \cite{subramanian2016computational} & 1,513 & 34.12 & 36.89 & - & 2 &        \begin{tabular}[c]{@{}l@{}}\cite{hu2019strategies,rong2020self,zhang2020motif,xu2021self,you2021graph,suresh2021adversarial}\\
		\end{tabular}\\ \hline
		\toprule
	\end{tabular}
\end{table*}

\subsection{Citation Networks}

\begin{table*}[htbp]
	\caption{A summary of open-source implementations.}
	\label{tab:codes}
	\centering
	\begin{adjustbox}{width=1.9\columnwidth,center}
	\begin{tabular}{ l l l }
		\toprule
		 Model & Framework & Github Link \\ \bottomrule
		 You et al. (2020) \cite{you2020does} & torch & \url{https://github.com/Shen-Lab/SS-GCNs} \\ \hline
		 Jin et al. (2021) \cite{jin2020self} & torch & \url{https://github.com/ChandlerBang/SelfTask-GNN} \\ \hline
		 Hu et al. (2020) \cite{hu2019strategies} & torch &\url{https://github.com/snap-stanford/pretrain-gnns} \\ \hline 
		 MGAE (2017) \cite{wang2017mgae} & matlab & \url{https://github.com/GRAND-Lab/MGAE} \\ \hline
		 GAE/VGAE (2016) \cite{kipf2016variational} & tensorflow & \url{https://github.com/tkipf/gae} \\ \hline
		 SIG-VAE (2019) \cite{hasanzadeh2019semi} & tensorflow & \url{https://github.com/sigvae/SIGraphVAE} \\ \hline
		 ARGA/ARVGA (2018) \cite{pan2018adversarially} & tensorflow & \url{https://github.com/GRAND-Lab/ARGA} \\ \hline
		 SuperGAT (2021) \cite{kim2021how} & torch & \url{https://github.com/dongkwan-kim/SuperGAT} \\ \hline
		 M3S (2020) \cite{sun2020multi} & tensorflow & \url{https://github.com/datake/M3S} \\ \hline
		 SimP-GCN (2021) \cite{jin2021node} & torch & \url{https://github.com/ChandlerBang/SimP-GCN} \\ \hline
		 DeepWalk (2014) \cite{perozzi2014deepwalk} & gensim & \url{https://github.com/phanein/deepwalk} \\ \hline
		 node2vec (2016) \cite{grover2016node2vec} & gensim & \url{https://github.com/aditya-grover/node2vec} \\ \hline
		 GraphSAGE (2017) \cite{hamilton2017inductive} & tensorflow & \url{https://github.com/williamleif/GraphSAGE} \\ \hline
		 SELAR (2020) \cite{hwang2020self} & torch & \url{https://github.com/mlvlab/SELAR} \\ \hline
		 LINE (2015) \cite{tang2015line} & c++ & \url{https://github.com/tangjianpku/LINE} \\ \hline
		 GRACE (2020) \cite{zhu2020deep} & torch & \url{https://github.com/CRIPAC-DIG/GRACE} \\ \hline
		 GCA (2021) \cite{zhu2020graph} & torch & \url{https://github.com/CRIPAC-DIG/GCA} \\ \hline
 		 GCC (2020) \cite{qiu2020gcc} & torch & \url{https://github.com/THUDM/GCC} \\ \hline
		 HeCo (2021) \cite{wang2021self} & torch & \url{https://github.com/liun-online/HeCo} \\ \hline
		 BGRL (2021) \cite{thakoor2021bootstrapped} & torch & \url{https://github.com/Namkyeong/BGRL_Pytorch} \\ \hline
		 SelfGNN (2021) \cite{kefato2021self} & torch & \url{https://github.com/zekarias-tilahun/SelfGNN} \\ \hline
 		 G-BT (2021) \cite{bielak2021graph} & torch & \url{https://github.com/pbielak/graph-barlow-twins} \\ \hline
		 MERIT (2021) \cite{Jin2021MultiScaleCS} & torch & \url{https://github.com/GRAND-Lab/MERIT} \\ \hline
		 GraphCL(G) (2020) \cite{you2020graph}& torch & \url{https://github.com/Shen-Lab/GraphCL} \\ \hline
		 AD-GCL (2021) \cite{suresh2021adversarial} & torch & \url{https://github.com/susheels/adgcl} \\ \hline
		 JOAO (2021) \cite{you2021graph} & torch & \url{https://github.com/Shen-Lab/GraphCL_Automated} \\ \hline
 		 LCGNN (2021) \cite{ren2021label} & torch & \tabincell{l}{\url{https://github.com/YuxiangRen/Label-Contrastive-Coding-} \\ \url{based-Graph-Neural-Network-for-Graph-Classification-}}  \\ \hline
		 DGI (2019) \cite{velivckovic2018deep} & torch & \url{https://github.com/PetarV-/DGI} \\ \hline
		 GIC (2021) \cite{mavromatis2020graph} & torch & \url{https://github.com/cmavro/Graph-InfoClust-GIC} \\ \hline
		 HDGI (2020) \cite{ren2020heterogeneous} & torch & \url{https://github.com/YuxiangRen/Heterogeneous-Deep-Graph-Infomax} \\ \hline
		 ConCH (2020) \cite{li2020leveraging} & torch & \url{https://github.com/dingdanhao110/Conch} \\ \hline
		 DMGI (2020) \cite{park2020unsupervised} & torch & \url{https://github.com/pcy1302/DMGI} \\ \hline
		 MVGRL (2020) \cite{hassani2020contrastive} & torch & \url{https://github.com/kavehhassani/mvgrl} \\ \hline
		 SUBG-CON (2020) \cite{jiao2020sub} & torch & \url{https://github.com/yzjiao/Subg-Con} \\ \hline
		 InfoGraph (2020) \cite{sun2019infograph} & torch & \url{https://github.com/fanyun-sun/InfoGraph} \\ \hline
		 SLiCE (2021) \cite{wang2020self} & torch & \url{https://github.com/pnnl/SLiCE} \\ \hline
		 Robinson et al. (2020) \cite{robinson2020contrastive} & torch & \url{https://github.com/joshr17/HCL} \\ \hline
		 EGI (2020) \cite{zhu2020transfer} & torch & \url{https://openreview.net/attachment?id=J_pvI6ap5Mn&name=supplementary_material} \\ \hline
		 BiGI (2021) \cite{cao2020bipartite} & torch & \url{https://github.com/caojiangxia/BiGI} \\ \hline
		 HTC (2021) \cite{wang2021learning} & torch & \url{https://github.com/Wastedzz/LGR} \\ \hline
		 MICRO-Graph (2020) \cite{zhang2020motif} & torch & \url{https://openreview.net/attachment?id=qcKh_Msv1GP&name=supplementary_material} \\ \hline
		 SUGAR (2021) \cite{sun2021sugar} & tensorflow & \url{https://github.com/RingBDStack/SUGAR} \\ \hline
		 GPT-GNN (2020) \cite{hu2020gpt} & torch & \url{https://github.com/acbull/GPT-GNN} \\ \hline
		 Graph-Bert (2020) \cite{zhang2020graph} & torch & \url{https://github.com/jwzhanggy/Graph-Bert} \\ \hline
		 PT-DGNN (2021) \cite{zhang2021pre} & torch & \url{https://github.com/Mobzhang/PT-DGNN} \\ \hline
		 GMI (2020) \cite{peng2020graph} & torch & \url{https://github.com/zpeng27/GMI} \\ \hline
		 MVMI-FT (2021) \cite{fan2021maximizing} & torch & \url{https://github.com/xiaolongo/MaxMIAcrossFT} \\ \hline
		 GraphLoG (2021) \cite{xu2021self} & torch & \url{https://openreview.net/attachment?id=DAaaaqPv9-q&name=supplementary_material} \\ \hline
		 HDMI (2021) \cite{jing2021hdmi} & torch & \url{https://github.com/baoyujing/HDMI} \\ \hline
		 LnL-GNN (2021) \cite{roy2021node} & torch & \url{https://github.com/forkkr/LnL-GNN} \\ \hline
 		 GROVER (2020) \cite{rong2020self} & torch & \url{https://github.com/tencent-ailab/grover} \\ \toprule
	\end{tabular}
	\end{adjustbox}
\vspace{-4mm}
\end{table*}

In citation networks, nodes often represent the published papers and/or authors, while edges denote the relationship between nodes, such as citation, authorship, and co-authorship. 
The features of each node usually contain the context of the papers or authors, and the labels denote the fields of study for each paper or author.
Specifically, in Cora, Citeseer and Pubmed \cite{sen2008collective}, the nodes are papers, the edges are citation relationships, and the features are the bag-of-word representation for papers. 
In Coauthor CS and Coauthor Physics \cite{shchur2018pitfalls}, the nodes are authors, the edges are co-authorship between authors, and the features are the keywords for each author’s papers.
In Wiki-CS \cite{mernyei2020wiki}, nodes represent papers about computer science, edges represent the hyperlinks between papers, and node features are the mean vectors of GloVe word embeddings of articles. The definition in ogbn-arxiv \cite{hu2020ogb} is similar to Wiki-CS.

\subsection{Co-purchase Networks}

In the two co-purchase graphs from Amazon (Amazon Computers and Amazon Photo \cite{shchur2018pitfalls}), the nodes indicate goods, and the edges indicate that two goods are frequently bought together. The features of each node are the bag-of-words encoded product reviews, while the class labels are obtained by the category of goods.

\subsection{Social Networks}

The social network datasets are often formed by users and their interactions on online services or communities. 
For instance, in Reddit \cite{hamilton2017inductive}, the data is collected from a large online discussion forum named Reddit. The nodes are the users post of the forum, the edges are the comments between users, and the class labels is the communities. The features are composed by: (1) the average embedding of the post title, (2) the average embedding of all the post’s comments (3) the post’s score, and (4) the number of comments made on the post. 
IMDB-B and IMDB-M \cite{yanardag2015deep} are two movie-collaboration datasets. Each graph includes the actors/actresses and genre information of a movie. In a graph, nodes indicate actors/actresses, while edges indicate if them if they play the same movie.
In REDDIT-BINARY (RDT-B) and REDDIT-MULTI-5K (RDT-M) \cite{yanardag2015deep}, each graph denotes an online discussion thread. In each graph, the nodes are the users, and the edges are the comments between users.
COLLAB \cite{yanardag2015deep} is a scientific-collaboration dataset, where each graph is a ego-network of an researcher, and the label is the field of this researcher.

\subsection{Bio-chemical Graphs}

Biochemical graphs are related to biology and chemistry domains. 
The Protein-Protein Interaction (PPI) \cite{zitnik2017predicting} dataset contains 24 biological graphs where nodes are proteins and edges are the interactions between proteins. 
In MUTAG \cite{debnath1991structure}, each graph indicates a nitro compounds, and its label denotes whether they are aromatic or heteroaromatic.
PROTEINS \cite{borgwardt2005protein} and D\&D \cite{dobson2003distinguishing} are also protein datasets, where graphs indicate proteins, and labels represent whether they are enzymes or non-enzymes.
The PTC \cite{toivonen2003statistical} and PTC-MR \cite{helma2001predictive} datasets contain a series of chemical compounds, while the labels indicate whether they are carcinogenic for male and female rats.
NCI-1 and NCI-109 \cite{wale2008comparison} also consist of chemical compounds, labeled as to whether they are active to hinder the growth of human cancer cell lines.
The Open Graph Benchmark (OGB) molecule property prediction benchmark \cite{hu2020ogb} contains 8 molecule graph datasets from different sources, including BBBP, Tox21, ToxCast, SIDER, ClinTox, MUV, HIB, and BACE. In these datasets, the graphs indicate different types of molecules, while the labels express their specific properties. For detailed information of the domains please refer to \cite{hu2020ogb}.
 
\section{Open-source Implementations} \label{appendix:implementations}

We collect the implementations of graph SSL approaches reviewed in this survey if there exists an open-source code for this method. The hyperlinks of the source codes are provided in Table \ref{tab:codes}.

\section{Other Graph SSL Applications}
\label{subsec:other_application}

For zero-shot expert linking problem in the field of expert finding, \methodHL{COAD} \cite{chen2020coad} leverages a same-scale contrastive learning framework to pre-train the expert encoder model. 
\methodHL{SLAPS} \cite{fatemi2021slaps} integrates a denoising node feature generation task into a classification model for graph structure learning. 
\methodHL{SCRL} \cite{kang2021self} applies a prototype-based pretext tasks for few-label graph learning. 
\methodHL{SDGE} \cite{xu2021self_sdge} is a community detection approach where the model is trained by a same-scale contrastive learning objective.
Aiming to repair program from diagnostic feedback, \methodHL{DrRepair} \cite{yasunaga2020graph} pre-trains the model by a repaired line prediction task which is learned with the automatically corrupted programs from online dataset.
\methodHL{C-SWM} \cite{Kipf2020Contrastive} introduces a same-scale node-level contrastive learning strategy to train structured world models that simulate multi-object systems.
\methodHL{Sehanobish et al.} \cite{sehanobish2020self} use a clustering-based auxiliary property classification task to train a GAT \cite{gat_ve2018graph} model and regard the learned edge weights as the edge feature for downstream tasks learned on biological datasets including SARS-CoV-2 and COVID-19.
To learn representations for medical images, \methodHL{Sun et al.} \cite{sun2021context} apply a context-based hybrid contrastive learning model that maximizes patch- and graph- level agreement on anatomy-aware graphs extracted from medical images. 
For federated learning \cite{tan2021fedproto,mcmahan2017communication}, \methodHL{FedGL} \cite{chen2021fedgl} introduces an auxiliary property classification task to provide a global view for the local training in clients. Specifically, the pseudo labels for each node is acquired by a weighted average fusion of clients' prediction results in a server.

%% file: bio.tex
\vspace{-1cm}
\begin{IEEEbiography}[{\includegraphics[width=1in,height=1.25in,clip,keepaspectratio]{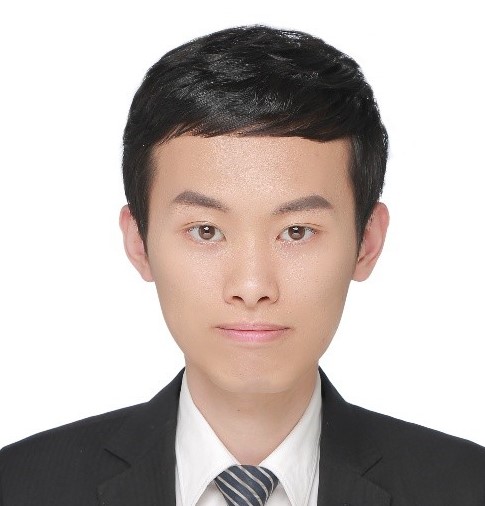}}]{Yixin Liu} received the B.S. degree and M.S. degree from Beihang University, Beijing, China, in 2017 and 2020, respectively. He is currently pursuing his Ph.D. degree in computer science at Monash University, Australia. His research concentrates on data mining, machine learning, and deep learning on graphs. 
\end{IEEEbiography}

\vspace{-1cm}

\begin{IEEEbiography}[{\includegraphics[width=1in,height=1.25in,clip,keepaspectratio]{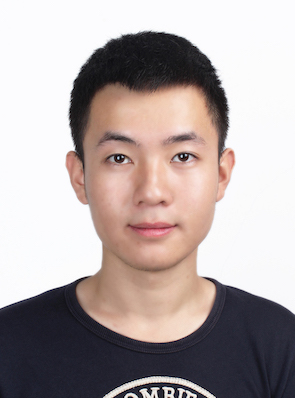}}]{Ming Jin} 
received the B.Eng. degree from the Hebei University of Technology, Tianjin, China, in 2017, and M.Inf.Tech. degree from the University of Melbourne, Melbourne, Australia, in 2019. 
He is currently pursuing his Ph.D. degree in computer science at Monash University, Melbourne, Australia.
His research focuses on graph neural networks (GNNs), time series analyse, data mining, and machine learning. 
\end{IEEEbiography}

\vspace{-1cm}

\begin{IEEEbiography}[{\includegraphics[width=1in,height=1.25in,clip,keepaspectratio]{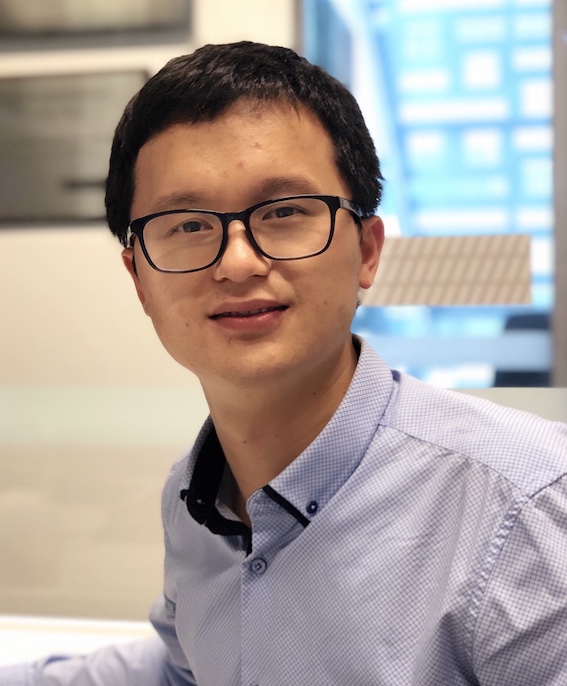}}]{Shirui Pan} received a Ph.D. in computer science from the University of Technology Sydney (UTS), Ultimo, NSW, Australia. He is currently an ARC Future Fellow and Senior Lecturer with the Faculty of Information Technology, Monash University, Australia. His research interests include data mining and machine learning. To date, Dr. Pan has published over 150 research papers in top-tier journals and conferences, including the TPAMI, TKDE,  TNNLS, NeurIPS, ICML, and KDD. He is recognised as one of the \textit{AI 2000 AAAI/IJCAI Most Influential Scholars} in Australia (2021).
%He is currently organizing a special issue, i.e., “\emph{Deep Neural Networks for Graphs:
%Theory, Models, Algorithms and Applications}”, in IEEE TNNLS.
\end{IEEEbiography}

\vspace{-1cm}

\begin{IEEEbiography}[{\includegraphics[width=1in,height=1.25in,clip,keepaspectratio]{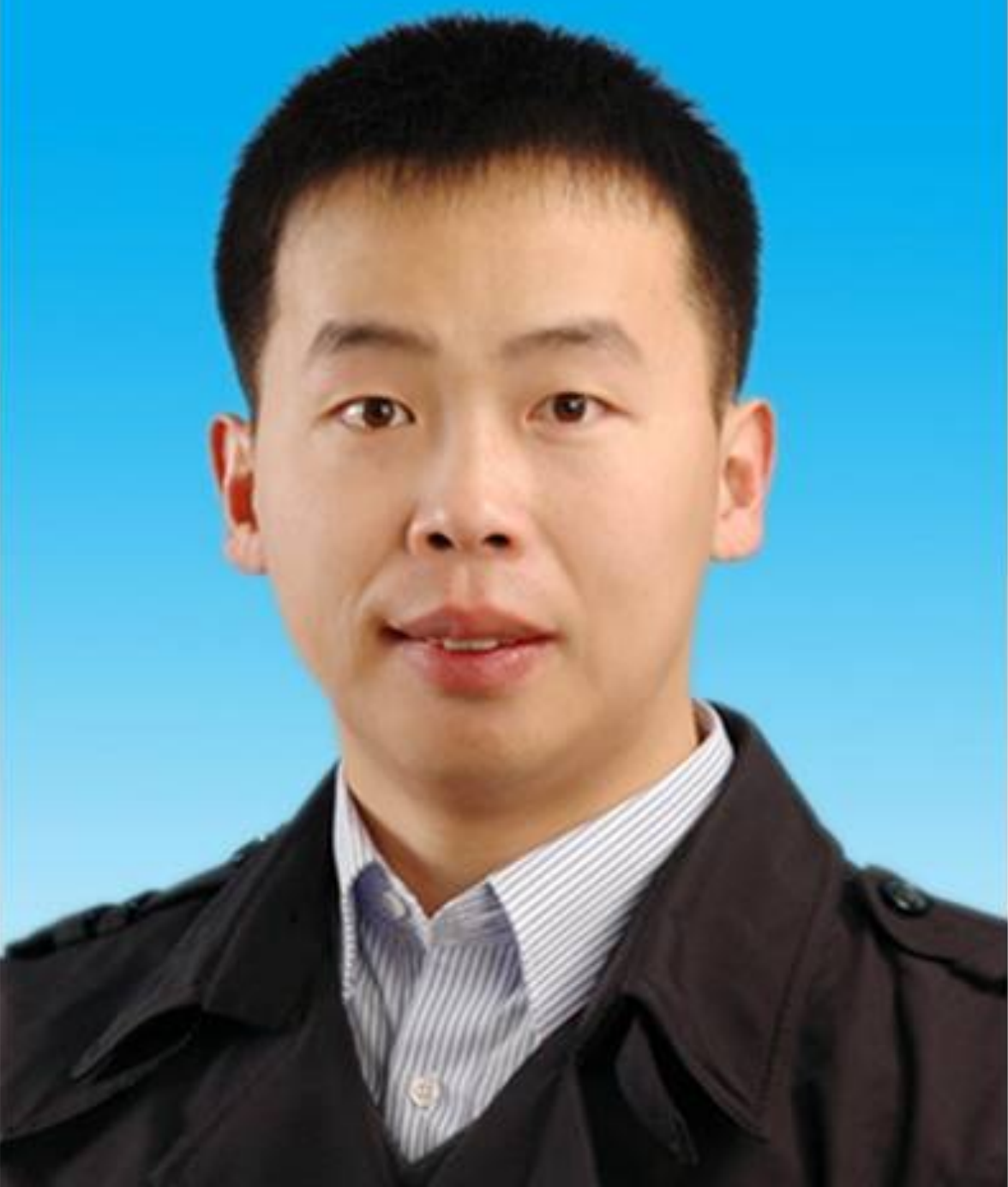}}]{Chuan Zhou} obtained Ph.D. degree from Chinese Academy of Sciences in 2013. He won the outstanding doctoral dissertation of Chinese Academy of Sciences in 2014, the best paper award of ICCS-14, and the best student paper award of IJCNN-17. Currently, he is an Associate Professor at the Academy of Mathematics and Systems Science, Chinese Academy of Sciences. His research interests include social network analysis and graph mining. To date, he has published more than 70 papers, including IEEE TKDE, ICDM, AAAI, CIKM, IJCAI and WWW.
\end{IEEEbiography}

\vspace{-1cm}

\begin{IEEEbiography}[{\includegraphics[width=1in,height=1.25in,clip,keepaspectratio]{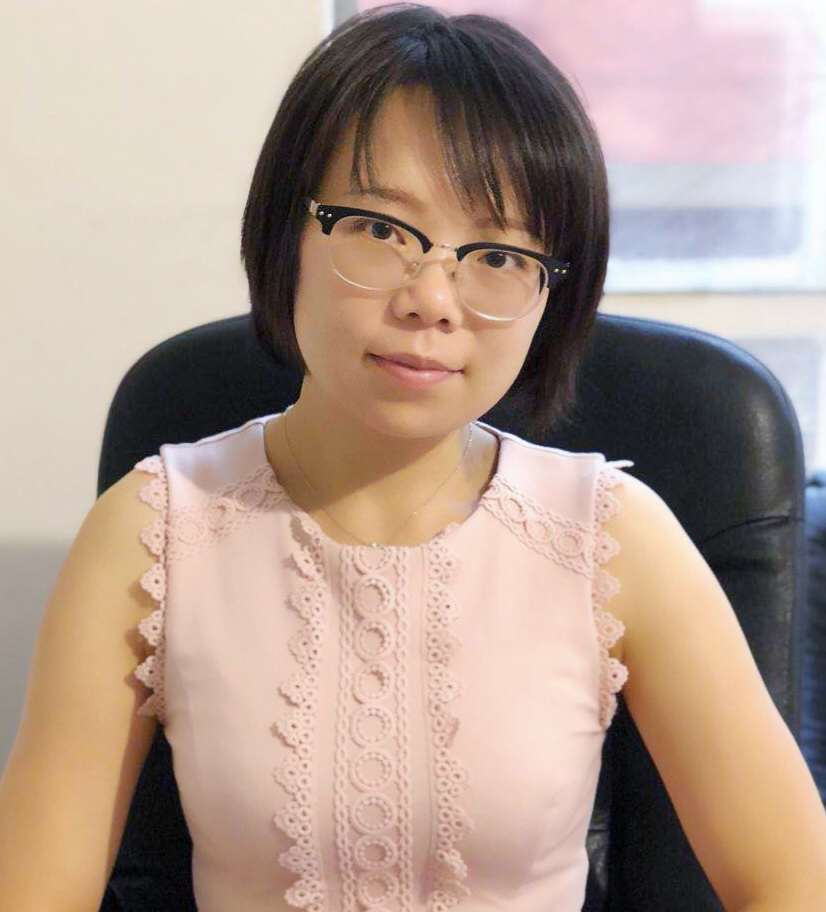}}]{Yu Zheng} received the B.S. and M.S. degrees in computer science from Northwest A\&F University, China, in 2008 and 2011, respectively. She is currently pursuing her Ph.D. degree in computer science at La Trobe University, Melbourne, Australia. Her research interests include image classification, data mining, and machine learning.
\end{IEEEbiography}

\vspace{-1cm}

\begin{IEEEbiography}[{\includegraphics[width=1in,height=1.25in,clip,keepaspectratio]{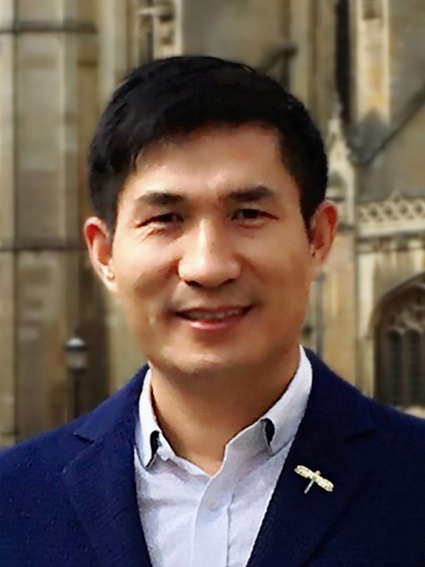}}]{Feng Xia (M’07-SM’12)} received the BSc and PhD degrees from Zhejiang University, Hangzhou, China. He was Full Professor and Associate Dean (Research) in School of Software, Dalian University of Technology, China. He is Associate Professor and former Discipline Leader (IT) in School of Engineering, IT and Physical Sciences, Federation University Australia. Dr. Xia has published 2 books and over 300 scientific papers in international journals and conferences. His research interests include data science, artificial intelligence, graph learning, and systems engineering. He is a Senior Member of IEEE and ACM. 
\end{IEEEbiography}

\vspace{-1cm}

\begin{IEEEbiography}[{\includegraphics[width=1in,height=1.25in,clip,keepaspectratio]{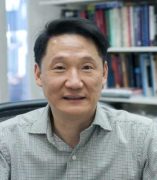}}]{Philip S. Yu} received the Ph.D. degree in electrical engineering from Stanford University, Stanford, CA, USA. He is currently a Distinguished Professor of computer science with the University of Illinois at Chicago, Chicago, IL, USA, where he is also the Wexler Chair in Information Technology. He has published more than 830 articles in refereed journals and conferences. He holds or has applied for more than 300 U.S. patents. His research interests include big data, data mining, data streams, databases, and privacy. Dr. Yu is a fellow of the ACM and IEEE. %He received the ACM SIGKDD 2016 Innovation Award, the Research Contributions Award from the IEEE International Conference on Data Mining in 2003, and the Technical Achievement Award from the IEEE Computer Society in 2013. 
\end{IEEEbiography}

%% file: main.bbl
\begin{thebibliography}{100}
\providecommand{\url}[1]{#1}
\csname url@samestyle\endcsname
\providecommand{\newblock}{\relax}
\providecommand{\bibinfo}[2]{#2}
\providecommand{\BIBentrySTDinterwordspacing}{\spaceskip=0pt\relax}
\providecommand{\BIBentryALTinterwordstretchfactor}{4}
\providecommand{\BIBentryALTinterwordspacing}{\spaceskip=\fontdimen2\font plus
\BIBentryALTinterwordstretchfactor\fontdimen3\font minus
  \fontdimen4\font\relax}
\providecommand{\BIBforeignlanguage}[2]{{%
\expandafter\ifx\csname l@#1\endcsname\relax
\typeout{** WARNING: IEEEtran.bst: No hyphenation pattern has been}%
\typeout{** loaded for the language `#1'. Using the pattern for}%
\typeout{** the default language instead.}%
\else
\language=\csname l@#1\endcsname
\fi
#2}}
\providecommand{\BIBdecl}{\relax}
\BIBdecl

\bibitem{gcn_kipf2017semi}
T.~N. Kipf and M.~Welling, ``Semi-supervised classification with graph
  convolutional networks,'' in \emph{ICLR}, 2017, pp. 1--14.

\bibitem{gat_ve2018graph}
P.~Veličković, G.~Cucurull, A.~Casanova, A.~Romero, P.~Liò, and Y.~Bengio,
  ``Graph attention networks,'' in \emph{ICLR}, 2018, pp. 1--12.

\bibitem{xu2018how}
K.~Xu, W.~Hu, J.~Leskovec, and S.~Jegelka, ``How powerful are graph neural
  networks?'' in \emph{ICLR}, 2019, pp. 1--17.

\bibitem{wu2020comprehensive}
Z.~Wu, S.~Pan, F.~Chen, G.~Long, C.~Zhang, and S.~Y. Philip, ``A comprehensive
  survey on graph neural networks,'' \emph{IEEE TNNLS}, vol.~32, no.~1, pp.
  4--24, 2020.

\bibitem{li2020hierarchical}
Z.~Li, X.~Shen, Y.~Jiao, X.~Pan, P.~Zou, X.~Meng, C.~Yao, and J.~Bu,
  ``Hierarchical bipartite graph neural networks: Towards large-scale
  e-commerce applications,'' in \emph{ICDE}.\hskip 1em plus 0.5em minus
  0.4em\relax IEEE, 2020.

\bibitem{wu2019graphwavenet}
Z.~Wu, S.~Pan, G.~Long, J.~Jiang, and C.~Zhang, ``Graph wavenet for deep
  spatial-temporal graph modeling,'' in \emph{IJCAI}, 2019.

\bibitem{liu2018constrained}
Q.~Liu, M.~Allamanis, M.~Brockschmidt, and A.~L. Gaunt, ``Constrained graph
  variational autoencoders for molecule design,'' in \emph{NeurIPS}, vol.~31,
  2018, pp. 7806--7815.

\bibitem{ji2021survey}
S.~Ji, S.~Pan, E.~Cambria, P.~Marttinen, and P.~S. Yu, ``A survey on knowledge
  graphs: Representation, acquisition, and applications,'' \emph{IEEE TNNLS},
  2021.

\bibitem{hu2020gpt}
Z.~Hu, Y.~Dong, K.~Wang, K.-W. Chang, and Y.~Sun, ``{GPT-GNN}: Generative
  pre-training of graph neural networks,'' in \emph{SIGKDD}, 2020, pp.
  1857--1867.

\bibitem{rong2020self}
Y.~Rong, Y.~Bian, T.~Xu, W.~Xie, Y.~WEI, W.~Huang, and J.~Huang,
  ``Self-supervised graph transformer on large-scale molecular data,'' in
  \emph{NeurIPS}, vol.~33, 2020, pp. 12\,559--12\,571.

\bibitem{Rong2020DropEdge}
Y.~Rong, W.~Huang, T.~Xu, and J.~Huang, ``{DropEdge}: Towards deep graph
  convolutional networks on node classification,'' in \emph{ICLR}, 2020, pp.
  1--17.

\bibitem{zhang2020adversarial}
M.~Zhang, L.~Hu, C.~Shi, and X.~Wang, ``Adversarial label-flipping attack and
  defense for graph neural networks,'' in \emph{ICDM}.\hskip 1em plus 0.5em
  minus 0.4em\relax IEEE, 2020, pp. 791--800.

\bibitem{velivckovic2018deep}
P.~Veli{\v{c}}kovi{\'c}, W.~Fedus, W.~L. Hamilton, P.~Li{\`o}, Y.~Bengio, and
  R.~D. Hjelm, ``Deep graph infomax,'' in \emph{ICLR}, 2019, pp. 1--17.

\bibitem{hassani2020contrastive}
K.~Hassani and A.~H. Khasahmadi, ``Contrastive multi-view representation
  learning on graphs,'' in \emph{ICML}.\hskip 1em plus 0.5em minus 0.4em\relax
  PMLR, 2020.

\bibitem{qiu2020gcc}
J.~Qiu, Q.~Chen, Y.~Dong, J.~Zhang, H.~Yang, M.~Ding, K.~Wang, and J.~Tang,
  ``{GCC}: Graph contrastive coding for graph neural network pre-training,'' in
  \emph{SIGKDD}, 2020, pp. 1150--1160.

\bibitem{hu2019strategies}
W.~Hu, B.~Liu, J.~Gomes, M.~Zitnik, P.~Liang, V.~Pande, and J.~Leskovec,
  ``Strategies for pre-training graph neural networks,'' in \emph{ICLR}, 2020,
  pp. 1--22.

\bibitem{you2020does}
Y.~You, T.~Chen, Z.~Wang, and Y.~Shen, ``When does self-supervision help graph
  convolutional networks?'' in \emph{ICML}.\hskip 1em plus 0.5em minus
  0.4em\relax PMLR, 2020, pp. 10\,871--10\,880.

\bibitem{jovanovic2021towards}
N.~Jovanovi{\'c}, Z.~Meng, L.~Faber, and R.~Wattenhofer, ``Towards robust graph
  contrastive learning,'' in \emph{WWW Workshop}, 2021.

\bibitem{jing2020self}
L.~Jing and Y.~Tian, ``Self-supervised visual feature learning with deep neural
  networks: A survey,'' \emph{IEEE TPAMI}, 2020.

\bibitem{pathak2016context_inpainting}
D.~Pathak, P.~Krahenbuhl, J.~Donahue, T.~Darrell, and A.~A. Efros, ``Context
  encoders: Feature learning by inpainting,'' in \emph{CVPR}, 2016, pp.
  2536--2544.

\bibitem{zhang2016colorful}
R.~Zhang, P.~Isola, and A.~A. Efros, ``Colorful image colorization,'' in
  \emph{ECCV}.\hskip 1em plus 0.5em minus 0.4em\relax Springer, 2016, pp.
  649--666.

\bibitem{noroozi2016unsupervised_jigsaw}
M.~Noroozi and P.~Favaro, ``Unsupervised learning of visual representations by
  solving jigsaw puzzles,'' in \emph{ECCV}.\hskip 1em plus 0.5em minus
  0.4em\relax Springer, 2016, pp. 69--84.

\bibitem{he2020momentum}
K.~He, H.~Fan, Y.~Wu, S.~Xie, and R.~Girshick, ``Momentum contrast for
  unsupervised visual representation learning,'' in \emph{CVPR}, 2020, pp.
  9729--9738.

\bibitem{chen2020simple}
T.~Chen, S.~Kornblith, M.~Norouzi, and G.~Hinton, ``A simple framework for
  contrastive learning of visual representations,'' in \emph{ICML}.\hskip 1em
  plus 0.5em minus 0.4em\relax PMLR, 2020, pp. 1597--1607.

\bibitem{grill2020bootstrap}
J.-B. Grill, F.~Strub, F.~Altch\'{e}, C.~Tallec, P.~Richemond, E.~Buchatskaya,
  C.~Doersch, B.~Avila~Pires, Z.~Guo, M.~Gheshlaghi~Azar, B.~Piot,
  k.~kavukcuoglu, R.~Munos, and M.~Valko, ``Bootstrap your own latent - a new
  approach to self-supervised learning,'' in \emph{NeurIPS}, vol.~33, 2020, pp.
  21\,271--21\,284.

\bibitem{mikolov2013befficient}
T.~Mikolov, K.~Chen, G.~Corrado, and J.~Dean, ``Efficient estimation of word
  representations in vector space,'' in \emph{ICLR}, 2013.

\bibitem{mikolov2013distributed}
T.~Mikolov, I.~Sutskever, K.~Chen, G.~S. Corrado, and J.~Dean, ``Distributed
  representations of words and phrases and their compositionality,'' in
  \emph{NeurIPS}, vol.~26, 2013, pp. 3111--3119.

\bibitem{devlin2018bert}
J.~Devlin, M.-W. Chang, K.~Lee, and K.~Toutanova, ``{BERT}: Pre-training of
  deep bidirectional transformers for language understanding,'' in
  \emph{NAACL}, 2019, pp. 4171--4186.

\bibitem{yang2019xlnet}
Z.~Yang, Z.~Dai, Y.~Yang, J.~Carbonell, R.~R. Salakhutdinov, and Q.~V. Le,
  ``{XLNet}: Generalized autoregressive pretraining for language
  understanding,'' in \emph{NeurIPS}, vol.~32, 2019.

\bibitem{perozzi2014deepwalk}
B.~Perozzi, R.~Al-Rfou, and S.~Skiena, ``{DeepWalk}: Online learning of social
  representations,'' in \emph{SIGKDD}, 2014, pp. 701--710.

\bibitem{grover2016node2vec}
A.~Grover and J.~Leskovec, ``node2vec: Scalable feature learning for
  networks,'' in \emph{SIGKDD}, 2016, pp. 855--864.

\bibitem{kipf2016variational}
T.~N. Kipf and M.~Welling, ``Variational graph auto-encoders,'' in
  \emph{NeurIPS Workshop}, 2016, pp. 1--3.

\bibitem{zhu2020deep}
Y.~Zhu, Y.~Xu, F.~Yu, Q.~Liu, S.~Wu, and L.~Wang, ``{Deep Graph Contrastive
  Representation Learning},'' in \emph{ICML Workshop}, 2020.

\bibitem{liu2020self}
X.~Liu, F.~Zhang, Z.~Hou, L.~Mian, Z.~Wang, J.~Zhang, and J.~Tang,
  ``Self-supervised learning: Generative or contrastive,'' \emph{IEEE TKDE},
  2021.

\bibitem{jaiswal2021survey}
A.~Jaiswal, A.~R. Babu, M.~Z. Zadeh, D.~Banerjee, and F.~Makedon, ``A survey on
  contrastive self-supervised learning,'' \emph{Technologies}, vol.~9, no.~1,
  p.~2, 2021.

\bibitem{xie2021self}
Y.~Xie, Z.~Xu, J.~Zhang, Z.~Wang, and S.~Ji, ``Self-supervised learning of
  graph neural networks: A unified review,'' \emph{arXiv:2102.10757}, 2021.

\bibitem{wu2021self}
L.~Wu, H.~Lin, Z.~Gao, C.~Tan, S.~Li \emph{et~al.}, ``Self-supervised on
  graphs: Contrastive, generative, or predictive,'' \emph{arXiv:2105.07342},
  2021.

\bibitem{you2020graph}
Y.~You, T.~Chen, Y.~Sui, T.~Chen, Z.~Wang, and Y.~Shen, ``Graph contrastive
  learning with augmentations,'' in \emph{NeurIPS}, 2020.

\bibitem{zhang2020graph}
J.~Zhang, H.~Zhang, C.~Xia, and L.~Sun, ``{Graph-Bert}: Only attention is
  needed for learning graph representations,'' \emph{arXiv:2001.05140}, 2020.

\bibitem{sun2020multi}
K.~Sun, Z.~Lin, and Z.~Zhu, ``Multi-stage self-supervised learning for graph
  convolutional networks on graphs with few labeled nodes,'' in \emph{AAAI},
  vol.~34, no.~04, 2020, pp. 5892--5899.

\bibitem{peng2020graph}
Z.~Peng, W.~Huang, M.~Luo, Q.~Zheng, Y.~Rong, T.~Xu, and J.~Huang, ``Graph
  representation learning via graphical mutual information maximization,'' in
  \emph{WWW}, 2020, pp. 259--270.

\bibitem{hinton2006reducing}
G.~E. Hinton and R.~R. Salakhutdinov, ``Reducing the dimensionality of data
  with neural networks,'' \emph{science}, vol. 313, no. 5786, pp. 504--507,
  2006.

\bibitem{jin2020self}
W.~Jin, T.~Derr, H.~Liu, Y.~Wang, S.~Wang, Z.~Liu, and J.~Tang,
  ``Self-supervised learning on graphs: Deep insights and new direction,'' in
  \emph{WWW Workshop}, 2021.

\bibitem{wang2017mgae}
C.~Wang, S.~Pan, G.~Long, X.~Zhu, and J.~Jiang, ``{MGAE}: Marginalized graph
  autoencoder for graph clustering,'' in \emph{CIKM}, 2017.

\bibitem{manessi2020graph}
F.~Manessi and A.~Rozza, ``Graph-based neural network models with multiple
  self-supervised auxiliary tasks,'' \emph{Pattern Recognition Letters}, vol.
  148, pp. 15--21, 2021.

\bibitem{park2019symmetric}
J.~Park, M.~Lee, H.~J. Chang, K.~Lee, and J.~Y. Choi, ``Symmetric graph
  convolutional autoencoder for unsupervised graph representation learning,''
  in \emph{ICCV}, 2019, pp. 6519--6528.

\bibitem{hasanzadeh2019semi}
E.~Hajiramezanali, A.~Hasanzadeh, N.~Duffield, K.~Narayanan, M.~Zhou, and
  X.~Qian, ``Semi-implicit graph variational auto-encoders,'' in
  \emph{NeurIPS}, vol.~32, 2019, pp. 10\,712--10\,723.

\bibitem{pan2018adversarially}
S.~Pan, R.~Hu, G.~Long, J.~Jiang, L.~Yao, and C.~Zhang, ``Adversarially
  regularized graph autoencoder for graph embedding,'' in \emph{IJCAI}, 2018,
  pp. 2609--2615.

\bibitem{kim2021how}
D.~Kim and A.~Oh, ``How to find your friendly neighborhood: Graph attention
  design with self-supervision,'' in \emph{ICLR}, 2021.

\bibitem{hu2019pre}
Z.~Hu, C.~Fan, T.~Chen, K.-W. Chang, and Y.~Sun, ``Pre-training graph neural
  networks for generic structural feature extraction,''
  \emph{arXiv:1905.13728}, 2019.

\bibitem{zhu2020self}
Q.~Zhu, B.~Du, and P.~Yan, ``Self-supervised training of graph convolutional
  networks,'' \emph{arXiv:2006.02380}, 2020.

\bibitem{wold1987principal}
S.~Wold, K.~Esbensen, and P.~Geladi, ``Principal component analysis,''
  \emph{Chemometrics and intelligent laboratory systems}, 1987.

\bibitem{vincent2010stacked}
P.~Vincent, H.~Larochelle, I.~Lajoie, Y.~Bengio, P.-A. Manzagol, and L.~Bottou,
  ``Stacked denoising autoencoders: Learning useful representations in a deep
  network with a local denoising criterion.'' \emph{Journal of machine learning
  research}, vol.~11, no.~12, 2010.

\bibitem{kingma2013auto}
D.~P. Kingma and M.~Welling, ``Auto-encoding variational bayes,'' in
  \emph{ICLR}, 2014, pp. 1--14.

\bibitem{goodfellow2014generative}
I.~J. Goodfellow, J.~Pouget-Abadie, M.~Mirza, B.~Xu, D.~Warde-Farley, S.~Ozair,
  A.~Courville, and Y.~Bengio, ``Generative adversarial nets,'' in
  \emph{NeurIPS}, vol.~2.\hskip 1em plus 0.5em minus 0.4em\relax MIT Press,
  2014, p. 2672–2680.

\bibitem{zhu2020cagnn}
Y.~Zhu, Y.~Xu, F.~Yu, S.~Wu, and L.~Wang, ``{CAGNN}: Cluster-aware graph neural
  networks for unsupervised graph representation learning,''
  \emph{arXiv:2009.01674}, 2020.

\bibitem{peng2020self}
Z.~Peng, Y.~Dong, M.~Luo, X.-M. Wu, and Q.~Zheng, ``Self-supervised graph
  representation learning via global context prediction,''
  \emph{arXiv:2003.01604}, 2020.

\bibitem{jin2021node}
W.~Jin, T.~Derr, Y.~Wang, Y.~Ma, Z.~Liu, and J.~Tang, ``Node similarity
  preserving graph convolutional networks,'' in \emph{WSDM}, 2021, pp.
  148--156.

\bibitem{karypis1995multilevel}
G.~Karypis and V.~Kumar, ``Multilevel graph partitioning schemes,'' in
  \emph{ICPP}, 1995, pp. 113--122.

\bibitem{kang2021multi}
Z.~Lin, Z.~Kang, L.~Zhang, and L.~Tian, ``Multi-view attributed graph
  clustering,'' \emph{IEEE TKDE}, 2021.

\bibitem{caron2018deep}
M.~Caron, P.~Bojanowski, A.~Joulin, and M.~Douze, ``Deep clustering for
  unsupervised learning of visual features,'' in \emph{ECCV}, 2018, pp.
  132--149.

\bibitem{kang2021structured}
Z.~Kang, Z.~Lin, X.~Zhu, and W.~Xu, ``Structured graph learning for scalable
  subspace clustering: From single view to multiview,'' \emph{IEEE TCYB}, 2021.

\bibitem{schaeffer2007graph}
S.~E. Schaeffer, ``Graph clustering,'' \emph{Computer science review}, vol.~1,
  no.~1, pp. 27--64, 2007.

\bibitem{karypis1998fast}
G.~Karypis and V.~Kumar, ``A fast and high quality multilevel scheme for
  partitioning irregular graphs,'' \emph{SIAM Journal on scientific Computing},
  vol.~20, no.~1, pp. 359--392, 1998.

\bibitem{hjelm2018learning}
R.~D. Hjelm, A.~Fedorov, S.~Lavoie-Marchildon, K.~Grewal, P.~Bachman,
  A.~Trischler, and Y.~Bengio, ``Learning deep representations by mutual
  information estimation and maximization,'' in \emph{ICLR}, 2019.

\bibitem{Tian2020ContrastiveMC}
Y.~Tian, D.~Krishnan, and P.~Isola, ``Contrastive multiview coding,'' in
  \emph{ECCV}.\hskip 1em plus 0.5em minus 0.4em\relax Springer, 2020, pp.
  776--794.

\bibitem{wang2021contrastive}
X.~Wang and G.-J. Qi, ``Contrastive learning with stronger augmentations,''
  \emph{arXiv:2104.07713}, 2021.

\bibitem{Jin2021MultiScaleCS}
M.~Jin, Y.~Zheng, Y.-F. Li, C.~Gong, C.~Zhou, and S.~Pan, ``Multi-scale
  contrastive siamese networks for self-supervised graph representation
  learning,'' in \emph{IJCAI}, 2021.

\bibitem{zhu2020graph}
Y.~Zhu, Y.~Xu, F.~Yu, Q.~Liu, S.~Wu, and L.~Wang, ``Graph contrastive learning
  with adaptive augmentation,'' in \emph{WWW}, 2021.

\bibitem{you2021graph}
Y.~You, T.~Chen, Y.~Shen, and Z.~Wang, ``Graph contrastive learning
  automated,'' in \emph{ICML}.\hskip 1em plus 0.5em minus 0.4em\relax PMLR,
  2021.

\bibitem{opolka2019spatio}
F.~L. Opolka, A.~Solomon, C.~Cangea, P.~Veli{\v{c}}kovi{\'c}, P.~Li{\`o}, and
  R.~D. Hjelm, ``Spatio-temporal deep graph infomax,'' in \emph{ICLR Workshop},
  2019, pp. 1--6.

\bibitem{ren2019heterogeneous}
Y.~Ren, B.~Liu, C.~Huang, P.~Dai, L.~Bo, and J.~Zhang, ``{HDGI}: An
  unsupervised graph neural network for representation learning in
  heterogeneous graph,'' in \emph{AAAI Workshop}, 2020.

\bibitem{zhang2020iterative}
H.~Zhang, S.~Lin, W.~Liu, P.~Zhou, J.~Tang, X.~Liang, and E.~P. Xing,
  ``Iterative graph self-distillation,'' in \emph{WWW Workshop}, 2021.

\bibitem{zeng2020contrastive}
J.~Zeng and P.~Xie, ``Contrastive self-supervised learning for graph
  classification,'' in \emph{AAAI}, vol.~35, no.~12, 2021.

\bibitem{suresh2021adversarial}
S.~Suresh, P.~Li, C.~Hao, and J.~Neville, ``Adversarial graph augmentation to
  improve graph contrastive learning,'' in \emph{NeurIPS}, 2021.

\bibitem{klicpera2019diffusion}
J.~Klicpera, S.~Wei{\ss}enberger, and S.~G{\"u}nnemann, ``Diffusion improves
  graph learning,'' in \emph{NeurIPS}, vol.~32, 2019.

\bibitem{jiao2020sub}
Y.~Jiao, Y.~Xiong, J.~Zhang, Y.~Zhang, T.~Zhang, and Y.~Zhu, ``Sub-graph
  contrast for scalable self-supervised graph representation learning,'' in
  \emph{ICDM}.\hskip 1em plus 0.5em minus 0.4em\relax IEEE, 2020, pp. 222--231.

\bibitem{hamilton2017inductive}
W.~L. Hamilton, R.~Ying, and J.~Leskovec, ``Inductive representation learning
  on large graphs,'' in \emph{NeurIPS}, 2017, p. 1025–1035.

\bibitem{tang2015line}
J.~Tang, M.~Qu, M.~Wang, M.~Zhang, J.~Yan, and Q.~Mei, ``{LINE}: Large-scale
  information network embedding,'' in \emph{WWW}, 2015.

\bibitem{hwang2020self}
D.~Hwang, J.~Park, S.~Kwon, K.~Kim, J.-W. Ha, and H.~J. Kim, ``Self-supervised
  auxiliary learning with meta-paths for heterogeneous graphs,'' in
  \emph{NeurIPS}, vol.~33, 2020, pp. 10\,294--10\,305.

\bibitem{hafidi2020graphcl}
H.~Hafidi, M.~Ghogho, P.~Ciblat, and A.~Swami, ``{GraphCL}: Contrastive
  self-supervised learning of graph representations,'' \emph{arXiv:2007.08025},
  2020.

\bibitem{wan2021contrastive}
S.~Wan, Y.~Zhan, L.~Liu, B.~Yu, S.~Pan, and C.~Gong, ``Contrastive graph
  poisson networks: Semi-supervised learning with extremely limited labels,''
  in \emph{NeurIPS}, 2021.

\bibitem{wang2021self}
X.~Wang, N.~Liu, H.~Han, and C.~Shi, ``Self-supervised heterogeneous graph
  neural network with co-contrastive learning,'' in \emph{SIGKDD}, 2021, pp.
  1726--1736.

\bibitem{thakoor2021bootstrapped}
S.~Thakoor, C.~Tallec, M.~G. Azar, R.~Munos, P.~Veli{\v{c}}kovi{\'c}, and
  M.~Valko, ``Bootstrapped representation learning on graphs,'' in \emph{ICLR
  Workshop}, 2021.

\bibitem{kefato2021self}
Z.~T. Kefato and S.~Girdzijauskas, ``Self-supervised graph neural networks
  without explicit negative sampling,'' in \emph{WWW Workshop}, 2021.

\bibitem{zbontar2021barlow}
J.~Zbontar, L.~Jing, I.~Misra, Y.~LeCun, and S.~Deny, ``Barlow twins:
  Self-supervised learning via redundancy reduction,'' in \emph{ICML}, 2021.

\bibitem{bielak2021graph}
P.~Bielak, T.~Kajdanowicz, and N.~V. Chawla, ``{Graph Barlow Twins}: A
  self-supervised representation learning framework for graphs,''
  \emph{arXiv:2106.02466}, 2021.

\bibitem{chen2021exploring}
X.~Chen and K.~He, ``Exploring simple siamese representation learning,'' in
  \emph{CVPR}, 2021, pp. 15\,750--15\,758.

\bibitem{verma2020towards}
V.~Verma, T.~Luong, K.~Kawaguchi, H.~Pham, and Q.~Le, ``Towards domain-agnostic
  contrastive learning,'' in \emph{ICML}.\hskip 1em plus 0.5em minus
  0.4em\relax PMLR, 2021, pp. 10\,530--10\,541.

\bibitem{ren2021label}
Y.~Ren, J.~Bai, and J.~Zhang, ``Label contrastive coding based graph neural
  network for graph classification,'' in \emph{Database Systems for Advanced
  Applications}, 2021, pp. 123--140.

\bibitem{mavromatis2020graph}
C.~Mavromatis and G.~Karypis, ``Graph infoclust: Maximizing coarse-grain mutual
  information in graphs,'' in \emph{PAKDD}.\hskip 1em plus 0.5em minus
  0.4em\relax Springer, 2021, pp. 541--553.

\bibitem{ren2020heterogeneous}
Y.~Ren and B.~Liu, ``Heterogeneous deep graph infomax,'' in \emph{AAAI
  Workshop}, 2020, pp. 1--6.

\bibitem{li2020leveraging}
X.~Li, D.~Ding, B.~Kao, Y.~Sun, and N.~Mamoulis, ``Leveraging meta-path
  contexts for classification in heterogeneous information networks,'' in
  \emph{ICDE}.\hskip 1em plus 0.5em minus 0.4em\relax IEEE, 2021, pp. 912--923.

\bibitem{park2020unsupervised}
C.~Park, D.~Kim, J.~Han, and H.~Yu, ``Unsupervised attributed multiplex network
  embedding,'' in \emph{AAAI}, vol.~34, no.~04, 2020.

\bibitem{zhu2020transfer}
Q.~Zhu, Y.~Xu, H.~Wang, C.~Zhang, J.~Han, and C.~Yang, ``Transfer learning of
  graph neural networks with ego-graph information maximization,'' in \emph{WWW
  Workshop}, 2021.

\bibitem{wang2020self}
P.~Wang, K.~Agarwal, C.~Ham, S.~Choudhury, and C.~K. Reddy, ``Self-supervised
  learning of contextual embeddings for link prediction in heterogeneous
  networks,'' in \emph{WWW}, 2021.

\bibitem{sun2019infograph}
F.-Y. Sun, J.~Hoffman, V.~Verma, and J.~Tang, ``{InfoGraph}: Unsupervised and
  semi-supervised graph-level representation learning via mutual information
  maximization,'' in \emph{ICLR}, 2020.

\bibitem{robinson2020contrastive}
J.~D. Robinson, C.-Y. Chuang, S.~Sra, and S.~Jegelka, ``Contrastive learning
  with hard negative samples,'' in \emph{ICLR}, 2021, pp. 1--29.

\bibitem{cao2020bipartite}
J.~Cao, X.~Lin, S.~Guo, L.~Liu, T.~Liu, and B.~Wang, ``Bipartite graph
  embedding via mutual information maximization,'' in \emph{WSDM}, 2021, pp.
  635--643.

\bibitem{wang2021learning}
C.~Wang and Z.~Liu, ``Learning graph representation by aggregating subgraphs
  via mutual information maximization,'' \emph{arXiv:2103.13125}, 2021.

\bibitem{zhang2020motif}
A.~Subramonian, ``Motif-driven contrastive learning of graph representations,''
  in \emph{AAAI}, vol.~35, no.~18, 2021, pp. 15\,980--15\,981.

\bibitem{sun2021sugar}
Q.~Sun, J.~Li, H.~Peng, J.~Wu, Y.~Ning, P.~S. Yu, and L.~He, ``{SUGAR}:
  Subgraph neural network with reinforcement pooling and self-supervised mutual
  information mechanism,'' in \emph{WWW}, 2021.

\bibitem{tschannen2019mutual}
M.~Tschannen, J.~Djolonga, P.~K. Rubenstein, S.~Gelly, and M.~Lucic, ``On
  mutual information maximization for representation learning,'' in
  \emph{ICLR}, 2020, pp. 1--16.

\bibitem{oord2018representation}
A.~v.~d. Oord, Y.~Li, and O.~Vinyals, ``Representation learning with
  contrastive predictive coding,'' \emph{arXiv:1807.03748}, 2018.

\bibitem{zhang2021pre}
J.~Zhang, K.~Chen, and Y.~Wang, ``Pre-training on dynamic graph neural
  networks,'' \emph{arXiv:2102.12380}, 2021.

\bibitem{wan2020contrastive}
S.~Wan, S.~Pan, J.~Yang, and C.~Gong, ``Contrastive and generative graph
  convolutional networks for graph-based semi-supervised learning,'' in
  \emph{AAAI}, vol.~35, no.~11, 2021, pp. 10\,049--10\,057.

\bibitem{fan2021maximizing}
X.~Fan, M.~Gong, Y.~Wu, and H.~Li, ``Maximizing mutual information across
  feature and topology views for learning graph representations,''
  \emph{arXiv:2105.06715}, 2021.

\bibitem{xu2021self}
M.~Xu, H.~Wang, B.~Ni, H.~Guo, and J.~Tang, ``Self-supervised graph-level
  representation learning with local and global structure,'' in
  \emph{ICML}.\hskip 1em plus 0.5em minus 0.4em\relax PMLR, 2021.

\bibitem{jing2021hdmi}
B.~Jing, C.~Park, and H.~Tong, ``{HDMI}: High-order deep multiplex infomax,''
  in \emph{WWW}, 2021, pp. 2414--2424.

\bibitem{zheng2021gzoom}
Y.~Zheng, M.~Jin, S.~Pan, Y.-F. Li, H.~Peng, M.~Li, and Z.~Li, ``Towards graph
  self-supervised learning with contrastive adjusted zooming,''
  \emph{arXiv:2111.10698}, 2021.

\bibitem{roy2021node}
K.~K. Roy, A.~Roy, A.~Rahman, M.~A. Amin, and A.~A. Ali, ``Node embedding using
  mutual information and self-supervision based bi-level aggregation,'' in
  \emph{IJCNN}, 2021.

\bibitem{kou2021self}
S.~Kou, W.~Xia, X.~Zhang, Q.~Gao, and X.~Gao, ``Self-supervised graph
  convolutional clustering by preserving latent distribution,''
  \emph{Neurocomputing}, vol. 437, pp. 218--226, 2021.

\bibitem{sen2008collective}
P.~Sen, G.~Namata, M.~Bilgic, L.~Getoor, B.~Galligher, and T.~Eliassi-Rad,
  ``Collective classification in network data,'' \emph{AI magazine}, vol.~29,
  no.~3, pp. 93--93, 2008.

\bibitem{hao2021pre}
B.~Hao, J.~Zhang, H.~Yin, C.~Li, and H.~Chen, ``Pre-training graph neural
  networks for cold-start users and items representation,'' in \emph{WSDM},
  2021, pp. 265--273.

\bibitem{yu2021self}
J.~Yu, H.~Yin, J.~Li, Q.~Wang, N.~Q.~V. Hung, and X.~Zhang, ``Self-supervised
  multi-channel hypergraph convolutional network for social recommendation,''
  in \emph{WWW}, 2021, pp. 413--424.

\bibitem{xia2021self}
X.~Xia, H.~Yin, J.~Yu, Q.~Wang, L.~Cui, and X.~Zhang, ``Self-supervised
  hypergraph convolutional networks for session-based recommendation,'' in
  \emph{AAAI}, vol.~35, no.~5, 2021.

\bibitem{liu2021contrastive}
Z.~Liu, Y.~Ma, Y.~Ouyang, and Z.~Xiong, ``Contrastive learning for recommender
  system,'' \emph{arXiv:2101.01317}, 2021.

\bibitem{liu2020pre}
Y.~Liu, S.~Yang, C.~Lei, G.~Wang, H.~Tang, J.~Zhang, A.~Sun, and C.~Miao,
  ``Pre-training graph transformer with multimodal side information for
  recommendation,'' in \emph{ACM Multimedia}, 2021.

\bibitem{liu2021anomaly}
Y.~Liu, Z.~Li, S.~Pan, C.~Gong, C.~Zhou, and G.~Karypis, ``Anomaly detection on
  attributed networks via contrastive self-supervised learning,'' \emph{IEEE
  TNNLS}, 2021.

\bibitem{ding2019deep}
K.~Ding, J.~Li, R.~Bhanushali, and H.~Liu, ``Deep anomaly detection on
  attributed networks,'' in \emph{SDM}.\hskip 1em plus 0.5em minus 0.4em\relax
  SIAM, 2019, pp. 594--602.

\bibitem{li2019specae}
Y.~Li, X.~Huang, J.~Li, M.~Du, and N.~Zou, ``{SpecAE}: Spectral autoencoder for
  anomaly detection in attributed networks,'' in \emph{CIKM}, 2019, pp.
  2233--2236.

\bibitem{ding2020inductive}
K.~Ding, J.~Li, N.~Agarwal, and H.~Liu, ``Inductive anomaly detection on
  attributed networks,'' in \emph{IJCAI}, 2020, pp. 1288--1294.

\bibitem{jin2021anemone}
M.~Jin, Y.~Liu, Y.~Zheng, L.~Chi, Y.-F. Li, and S.~Pan, ``Anemone: Graph
  anomaly detection with multi-scale contrastive learning,'' in \emph{CIKM},
  2021, pp. 3122--3126.

\bibitem{zheng2021generative}
Y.~Zheng, M.~Jin, Y.~Liu, L.~Chi, K.~T. Phan, and Y.-P.~P. Chen, ``Generative
  and contrastive self-supervised learning for graph anomaly detection,''
  \emph{IEEE TKDE}, 2021.

\bibitem{huang2021hop}
T.~Huang, Y.~Pei, V.~Menkovski, and M.~Pechenizkiy, ``Hop-count based
  self-supervised anomaly detection on attributed networks,''
  \emph{arXiv:2104.07917}, 2021.

\bibitem{wang2021molclr}
Y.~Wang, J.~Wang, Z.~Cao, and A.~B. Farimani, ``{MolCLR}: Molecular contrastive
  learning of representations via graph neural networks,''
  \emph{arXiv:2102.10056}, 2021.

\bibitem{fang2021knowledge}
Y.~Fang, H.~Yang, X.~Zhuang, X.~Shao, X.~Fan, and H.~Chen, ``Knowledge-aware
  contrastive molecular graph learning,'' \emph{arXiv:2103.13047}, 2021.

\bibitem{cheng2021graphms}
S.~Cheng, L.~Zhang, B.~Jin, Q.~Zhang, X.~Lu, M.~You, and X.~Tian, ``{GraphMS}:
  Drug target prediction using graph representation learning with
  substructures,'' \emph{Applied Sciences}, 2021.

\bibitem{wang2021multi}
Y.~Wang, Y.~Min, X.~Chen, and J.~Wu, ``Multi-view graph contrastive
  representation learning for drug-drug interaction prediction,'' in
  \emph{WWW}, 2021, pp. 2921--2933.

\bibitem{jin2021survey}
D.~Jin, Z.~Yu, P.~Jiao, S.~Pan, P.~S. Yu, and W.~Zhang, ``A survey of community
  detection approaches: From statistical modeling to deep learning,''
  \emph{IEEE TKDE}, 2021.

\bibitem{ying2018hierarchical}
R.~Ying, J.~You, C.~Morris, X.~Ren, W.~L. Hamilton, and J.~Leskovec,
  ``Hierarchical graph representation learning with differentiable pooling,''
  in \emph{NeurIPS}, 2018, p. 4805–4815.

\bibitem{wang2020haar}
Y.~G. Wang, M.~Li, Z.~Ma, G.~Montufar, X.~Zhuang, and Y.~Fan, ``Haar graph
  pooling,'' in \emph{ICML}.\hskip 1em plus 0.5em minus 0.4em\relax PMLR, 2020,
  pp. 9952--9962.

\bibitem{yanardag2015deep}
P.~Yanardag and S.~Vishwanathan, ``Deep graph kernels,'' in \emph{SIGKDD},
  2015, pp. 1365--1374.

\bibitem{shchur2018pitfalls}
O.~Shchur, M.~Mumme, A.~Bojchevski, and S.~G{\"u}nnemann, ``Pitfalls of graph
  neural network evaluation,'' in \emph{NeurIPS}, 2018.

\bibitem{mernyei2020wiki}
P.~Mernyei and C.~Cangea, ``{Wiki-CS}: A wikipedia-based benchmark for graph
  neural networks,'' in \emph{ICML Workshop}, 2020.

\bibitem{hu2020ogb}
W.~Hu, M.~Fey, M.~Zitnik, Y.~Dong, H.~Ren, B.~Liu, M.~Catasta, and J.~Leskovec,
  ``{Open Graph Benchmark}: Datasets for machine learning on graphs,'' in
  \emph{NeurIPS}, vol.~33, 2020, pp. 22\,118--22\,133.

\bibitem{zitnik2017predicting}
M.~Zitnik and J.~Leskovec, ``Predicting multicellular function through
  multi-layer tissue networks,'' \emph{Bioinformatics}, vol.~33, no.~14, pp.
  i190--i198, 2017.

\bibitem{debnath1991structure}
A.~K. Debnath, R.~L. Lopez~de Compadre, G.~Debnath, A.~J. Shusterman, and
  C.~Hansch, ``Structure-activity relationship of mutagenic aromatic and
  heteroaromatic nitro compounds. correlation with molecular orbital energies
  and hydrophobicity,'' \emph{Journal of medicinal chemistry}, vol.~34, no.~2,
  pp. 786--797, 1991.

\bibitem{borgwardt2005protein}
K.~M. Borgwardt, C.~S. Ong, S.~Sch{\"o}nauer, S.~Vishwanathan, A.~J. Smola, and
  H.-P. Kriegel, ``Protein function prediction via graph kernels,''
  \emph{Bioinformatics}, vol.~21, no. suppl\_1, pp. i47--i56, 2005.

\bibitem{dobson2003distinguishing}
P.~D. Dobson and A.~J. Doig, ``Distinguishing enzyme structures from
  non-enzymes without alignments,'' \emph{Journal of molecular biology}, vol.
  330, no.~4, pp. 771--783, 2003.

\bibitem{toivonen2003statistical}
H.~Toivonen, A.~Srinivasan, R.~D. King, S.~Kramer, and C.~Helma, ``Statistical
  evaluation of the predictive toxicology challenge 2000--2001,''
  \emph{Bioinformatics}, vol.~19, no.~10, pp. 1183--1193, 2003.

\bibitem{helma2001predictive}
C.~Helma, R.~D. King, S.~Kramer, and A.~Srinivasan, ``The predictive toxicology
  challenge 2000--2001,'' \emph{Bioinformatics}, 2001.

\bibitem{wale2008comparison}
N.~Wale, I.~A. Watson, and G.~Karypis, ``Comparison of descriptor spaces for
  chemical compound retrieval and classification,'' \emph{Knowledge and
  Information Systems}, vol.~14, no.~3, 2008.

\bibitem{martins2012bayesian}
I.~F. Martins, A.~L. Teixeira, L.~Pinheiro, and A.~Falcao, ``A {Bayesian}
  approach to in silico blood-brain barrier penetration modeling,''
  \emph{JCIM}, vol.~52, no.~6, pp. 1686--1697, 2012.

\bibitem{tox21}
\BIBentryALTinterwordspacing
Tox21, ``Tox21 data challenge 2014,'' 2014. [Online]. Available:
  \url{https://tripod.nih.gov/tox21/ challenge/}
\BIBentrySTDinterwordspacing

\bibitem{richard2016toxcast}
A.~M. Richard, R.~S. Judson, K.~A. Houck, C.~M. Grulke, P.~Volarath,
  I.~Thillainadarajah, C.~Yang, J.~Rathman, M.~T. Martin, J.~F. Wambaugh
  \emph{et~al.}, ``{ToxCast} chemical landscape: paving the road to 21st
  century toxicology,'' \emph{Chemical research in toxicology}, vol.~29, no.~8,
  pp. 1225--1251, 2016.

\bibitem{kuhn2016sider}
M.~Kuhn, I.~Letunic, L.~J. Jensen, and P.~Bork, ``The {SIDER} database of drugs
  and side effects,'' \emph{Nucleic acids research}, vol.~44, no.~D1, pp.
  D1075--D1079, 2016.

\bibitem{novick2013sweetlead}
P.~A. Novick, O.~F. Ortiz, J.~Poelman, A.~Y. Abdulhay, and V.~S. Pande,
  ``{SWEETLEAD}: an in silico database of approved drugs, regulated chemicals,
  and herbal isolates for computer-aided drug discovery,'' \emph{PloS one},
  vol.~8, no.~11, p. e79568, 2013.

\bibitem{gardiner2011effectiveness}
E.~J. Gardiner, J.~D. Holliday, C.~O’Dowd, and P.~Willett, ``Effectiveness of
  {2D} fingerprints for scaffold hopping,'' \emph{Future medicinal chemistry},
  vol.~3, no.~4, pp. 405--414, 2011.

\bibitem{hiv2017}
\BIBentryALTinterwordspacing
``{AIDS} antiviral screen data,'' 2017. [Online]. Available:
  \url{http://wiki.nci.nih.gov/display/NCIDTPdata/AIDS}
\BIBentrySTDinterwordspacing

\bibitem{subramanian2016computational}
G.~Subramanian, B.~Ramsundar, V.~Pande, and R.~A. Denny, ``Computational
  modeling of $\beta$-secretase 1 ({BACE-1}) inhibitors using ligand based
  approaches,'' \emph{JCIM}, vol.~56, no.~10, pp. 1936--1949, 2016.

\bibitem{chen2020coad}
B.~Chen, J.~Zhang, X.~Zhang, X.~Tang, L.~Cai, H.~Chen, C.~Li, P.~Zhang, and
  J.~Tang, ``{COAD}: Contrastive pre-training with adversarial fine-tuning for
  zero-shot expert linking,'' \emph{arXiv:2012.11336}, 2020.

\bibitem{fatemi2021slaps}
B.~Fatemi, L.~E. Asri, and S.~M. Kazemi, ``{SLAPS}: Self-supervision improves
  structure learning for graph neural networks,'' in \emph{NeurIPS}, 2021.

\bibitem{kang2021self}
C.~Liu, L.~Wen, Z.~Kang, G.~Luo, and L.~Tian, ``Self-supervised consensus
  representation learning for attributed graph,'' in \emph{ACM Multimedia},
  2021, pp. 2654--2662.

\bibitem{xu2021self_sdge}
S.~Xu, S.~Liu, and L.~Feng, ``Self-supervised deep graph embedding with
  high-order information fusion for community discovery,''
  \emph{arXiv:2102.03302}, 2021.

\bibitem{yasunaga2020graph}
M.~Yasunaga and P.~Liang, ``Graph-based, self-supervised program repair from
  diagnostic feedback,'' in \emph{ICML}.\hskip 1em plus 0.5em minus 0.4em\relax
  PMLR, 2020.

\bibitem{Kipf2020Contrastive}
T.~Kipf, E.~van~der Pol, and M.~Welling, ``Contrastive learning of structured
  world models,'' in \emph{ICLR}, 2020, pp. 1--21.

\bibitem{sehanobish2020self}
A.~Sehanobish, N.~G. Ravindra, and D.~van Dijk, ``Self-supervised edge features
  for improved graph neural network training,'' \emph{arXiv:2007.04777}, 2020.

\bibitem{sun2021context}
L.~Sun, K.~Yu, and K.~Batmanghelich, ``Context matters: Graph-based
  self-supervised representation learning for medical images,'' in \emph{AAAI},
  vol.~35, no.~6, May 2021, pp. 4874--4882.

\bibitem{tan2021fedproto}
Y.~Tan, G.~Long, L.~Liu, T.~Zhou, Q.~Lu, J.~Jiang, and C.~Zhang, ``Fedproto:
  Federated prototype learning over heterogeneous devices,'' in \emph{AAAI},
  2022.

\bibitem{mcmahan2017communication}
B.~McMahan, E.~Moore, D.~Ramage, S.~Hampson, and B.~A. y~Arcas,
  ``Communication-efficient learning of deep networks from decentralized
  data,'' in \emph{AISTATS}, 2017, pp. 1273--1282.

\bibitem{chen2021fedgl}
C.~Chen, W.~Hu, Z.~Xu, and Z.~Zheng, ``{FedGL}: Federated graph learning
  framework with global self-supervision,'' \emph{arXiv:2105.03170}, 2021.

\end{thebibliography}
